\pdfoutput=1

\documentclass[11pt]{article}
\usepackage{booktabs} 
\usepackage{longtable} 
\usepackage[table]{xcolor} 

\usepackage[final]{acl}

\usepackage{times}
\usepackage{latexsym}

\usepackage[T1]{fontenc}

\usepackage[utf8]{inputenc}

\usepackage{microtype}

\usepackage{inconsolata}

\usepackage{graphicx}
\usepackage{subcaption}

\usepackage{array,multirow}
\usepackage{arydshln} 
\usepackage{amssymb}
\usepackage{pifont} 
\usepackage{float}
\usepackage{amsfonts} 
\usepackage{amsmath}

\usepackage{hyperref}
\usepackage{cleveref} 
\crefformat{section}{\S#2#1#3} 
\crefformat{subsection}{\S#2#1#3}
\crefformat{subsubsection}{\S#2#1#3}
\usepackage{url}
\usepackage{tikz-dependency}
\usepackage{mathabx}
\usepackage{tabularray}
\usepackage{tcolorbox}
\usepackage{lipsum}
\usepackage{etoolbox}
\AtBeginEnvironment{tcolorbox}{\small}
\makeatletter
\newcommand*\bigcdot{\mathpalette\bigcdot@{.5}}
\newcommand*\bigcdot@[2]{\mathbin{\vcenter{\hbox{\scalebox{#2}{$\m@th#1\bullet$}}}}}
\makeatother

\usepackage{placeins}
\usepackage{tabularx}
\usepackage{enumitem}

\title{BeDiscovER: The Benchmark of Discourse Understanding in the Era of Reasoning Language Models}

\author{Chuyuan Li \and Giuseppe Carenini \\
  Department of Computer Science \\
  University of British Columbia \\
  Vancouver, BC, Canada, V6T 1Z4 \\
  \texttt{chuyuan.li@ubc.ca}, \ \texttt{carenini@cs.ubc.ca}
  }

\begin{document}
\maketitle

\begin{abstract}
We introduce BeDiscovER (Benchmark of Discourse Understanding in the Era of Reasoning Language Models), an up-to-date, comprehensive suite for evaluating the discourse-level knowledge of modern LLMs.
BeDiscovER compiles 5 publicly available discourse tasks across discourse lexicon, (multi-)sentential, and documental levels, with in total 52 individual datasets. 
It covers both extensively studied tasks such as discourse parsing 
and temporal relation extraction, as well as some novel challenges such as discourse particle disambiguation (e.g., ``\textit{just}''), and also aggregates a shared-task on Discourse Relation Parsing and Treebanking
for multilingual and multi-framework discourse relation classification.
We evaluate open-source LLMs: Qwen3 series, DeepSeek-R1, and frontier reasoning model GPT-5-mini on BeDiscovER, and find that state-of-the-art models exhibits strong performance in arithmetic aspect of temporal reasoning, but they struggle with long-dependency reasoning and some subtle semantic and discourse phenomena, such as rhetorical relation classification.
\end{abstract}

\section{Introduction}


Large Language Models (LLMs) have been tasked with solving increasingly difficult problems that require strong reasoning abilities across text understanding, common-sense, math, and coding \citep{shao2024deepseekmath, jaech2024openai, ke2025survey}. Still, it is an open question how the discourse knowledge---that is, the ability to process and understand the interplay of multi-sentence and long-context information---of state-of-the-art reasoning models varies across various tasks and phenomena.

Discourse analysis 
has demonstrated impact across many downstream tasks.
Its attention to both local and global levels of text understanding has also spurred advances in long-document language modeling \citep{prasad2023meetingqa, feng2023kalm, ivgi2023efficient, buchmann2024document, li2025topic}. To achieve deeper understanding of natural language and induce stronger future reasoning models, it is essential for LLMs to capture and reason over discourse knowledge.



Recent studies have explored this question by evaluating LLMs' 
performance on a range of sentence-level tasks, such as temporal \citep{yuan2023zero, wei2024llms}, causal \citep{chan2024Exploring}, and discourse relations \citep{yung2024prompting, fan2024uncovering}. 
However, each of these studies uses a different set of models, metrics (probing, prompting, acceptability judgment), and tasks, limiting any possible bigger-picture conclusions.
\textit{Discourse understanding} is also approached from differing perspectives within the field:
some work focuses on generalization ability \citep{braud2024disrpt, eichin2025probing},
while other work on the answer faithfulness \citep{miao2024discursive}.
We, instead, use discourse understanding to refer broadly to the full range of discourse knowledge.
To this end, we introduce BeDiscovER -- the Benchmark of Discourse Understanding in the Era of Reasoning Language Models:
the first comprehensive, practical benchmark that covers five broad areas of discourse knowledge, namely: (1) Discourse Marker Understanding, (2) Temporal Reasoning, (3) Discourse Relation Recognition, (4) Sentence Ordering, and (5) Dialogue Discourse Parsing. 
In total, BeDiscovER comprises around 30,000 test instances across 52 publicly available datasets (most were annotated by human experts), 16 languages, and 6 frameworks, providing a practical testbed for evaluating LLMs’ reasoning skills across lexical, semantic, rhetorical, temporal, and commonsense knowledge.


We use BeDiscovER to study several reasoning-oriented LLMs: Qwen3 series \citep{yang2025qwen3}, DeepSeek-R1 \citep{guo2025deepseek}, DeepSeek-R1-Distll-Qwen-32B \cite{guo2025deepseek}, and GPT-5-mini \citep{gpt5}, 
since they have been shown to be more capable than models not explicitly optimized for reasoning, like GPT-4o-mini \citep{hurst2024gpt4o}.
Empirically, we approximate \textit{discourse understanding} ability using automated scoring based on task-specific answer correctness.
This gives us indirect evidence about each model's discourse knowledge and allows us to compare models in a consistent way. 
Through extensive experiments,
we conclude that whereas models like GPT-5-mini appear to have significant reasoning ability across many tasks, these strengths are concentrated in discourse phenomena well represented in training data, such as frequent particles (e.g., \textit{Just}\footnote{
\textit{Just} is 51st most frequent word and 2nd most frequent adverb (COCA \citep{davies2009coca}).})
or arithmetic-style temporal reasoning.
We hope that BeDiscovER unifies research efforts across different levels of discourse knowledge by revealing phenomena where even state-of-the-art language models still exhibit notable reasoning gaps, and by drawing attention to areas that future studies (both in evaluation and in pre-/post-training) should explore in greater depth.
We release the unified 
evaluation pipeline to support future assessment across the community\footnote{\url{https://github.com/chuyuanli/BeDiscovER}.}.

\begin{table*}[t]
    \centering
    \resizebox{\textwidth}{!}{
    \begin{tabular}{llllcc}
        \toprule
        Level & Task & Nature & Dataset & \# Examples$^\mathparagraph$ & Knowledge$^\ddagger$ \\
        \midrule
        \multirow{3}{*}{Lexical} & (1) Disc. Marker & Classification & \textit{Just-Manual} \shortcite{sheffield2025just} & 90 & L\&S, CS\\
        & Understanding & -- & \textit{Just-Subtitle} \shortcite{sheffield2025just} & 149 & L\&S, CS \\
        & & -- & \textit{Otherwise} \shortcite{liu2025otherwise} & 294 & L\&S, CS\\
        \midrule
        & (2) Temporal Reasoning & Classification & TimeBank-Dense \shortcite{cassidy2014tbdense} & 1515 & T, Logi, CS \\
        (multi-) & & -- & TDDiscourse \shortcite{naik2019tddiscourse} & 1,500 & T, Logi, CS\\
        Sentence & & QA & ToT-Arithmetic \shortcite{fatemi2024testoftime} & 1,850 & T, A, CS\\
        & (3) Relation Recognition & Classification & DISRPT Shared Task$^*$ \shortcite{braud2025disrpt} & 8,757 (43,716) & R, CS \\
        \midrule
        \multirow{10}{*}{Document} & (4) Sentence Ordering & Seq-generation &  AAN abstract \shortcite{wang2018aan} & 806 (2,687) & Logi, R, CS \\
        & & --  & ArXiv abstract \shortcite{chen2016arxiv} & 898 (179,691) & Logi, R, CS\\
        & & -- & Neurips abstract \shortcite{logeswaran2018neurips-nsf} & 377 (377) & Logi, R, CS\\
        & & -- & NSF abstract\shortcite{logeswaran2018neurips-nsf} & 814 (20,366) & Logi, R, CS\\
        & & -- & ROC stories \shortcite{mostafazadeh2016roc} & 883 (9,816) & Logi, R, CS \\
        & & -- & SIND \shortcite{huang2016sind} & 808 (5,055) & Logi, R, CS \\
        & & -- & Wikipedia movie plots \shortcite{chowdhury2021everything} & 836 (3,345) & Logi, R, CS \\
        & (5) Dialogue Disc. Parsing & Generation &  STAC \shortcite{asher2016discourse} & 1,045 & R, T, L\&S, CS \\
        & & -- & Molweni \shortcite{li2020molweni} & 3,930 & R, T, L\&S, CS\\
        & & -- & MSDC \shortcite{thompson2024discourse} & 1,474 & R, T, L\&S, CS\\ 
        \bottomrule
    \end{tabular}}
    \caption{Overview of level, task, nature of task, dataset, number of test examples, and discourse knowledge covered by BeDiscovER. 
    The first three tasks are multi-class classification (except ToT-arithmetic), tasks (4) and (5) are generation tasks. `--' indicates same as above. 
    \# examples$^\mathparagraph$: we evaluate subsets of tasks (3) and (4) due to their extremely large size (original sizes in parentheses). Representative text examples are provided in \cref{append:tasks}.
    Discourse Knowledge$^\ddagger$ abbreviations: L\&S, R, T, Logi, A, CS denote \textit{Lexicon \& Semantics}, \textit{Rhetorical}, \textit{Temporal}, \textit{Logic}, \textit{Arithmetic}, and \textit{Commonsense}, respectively.
    DISRPT$^*$: the latest 2025 version consists of 38 datasets.
    }
    \label{tab:overview}
    \vspace{-1ex}
\end{table*}

\section{Related Work}
\label{sec:related}

\textbf{Large Language Models.}
Recent advances in computational infrastructure and the availability of large-scale corpora have driven the development of LLMs capable of remarkable generalization \citep{gpt5, achiam2023gpt, team2023gemini, touvron2023llama}. 
These capabilities have been further enhanced by instruction tuning \citep{wei2021finetuned} and reinforcement learning from human feedback \citep{ouyang2022training}, enabling models to perform complex text reasoning.

This scaling paradigm has fueled a wave of increasingly capable LLMs 
and pushing the boundaries across a wide spectrum of benchmarks, from graduate-level exams \citep{rein2024gpqa} and mathematical problem solving \citep{phan2025humanity} to competitive programming and multi-agent systems \citep{kapoor2024ai, liang2024learning}. Despite these breakthroughs, LLMs continue to exhibit persistent limitations, particularly in modeling long-range dependencies, maintaining discourse coherence, and avoiding factual inconsistencies \citep{ivgi2023efficient, liang-etal-2023-open, wu2024rethinking}.
Discourse analysis, which focuses on understanding document-level structure, coherence, and information flow, offers a principled lens through which these limitations can be exposed and addressed. We pursue this line of research in this paper.

\textbf{Discourse Knowledge of LLMs. }
Earlier studies evaluate LM's discourse knowledge using probing tasks \citep{kim2019probing, li2021implicit, tao2024probing} 
or extracting sentence representation from PLM's attention networks \citep{chen2019evaluation, huber2022towards, li2023discourse}.
QA-based evaluation has emerged as a popular alternative, offering greater granularity and flexibility: QADiscourse \citep{pyatkin2020qadiscourse}, QA for reference/ellipsis resolution \citep{aralikatte2021ellipsis}, QA for discourse faithfulness \citep{miao2024discursive}, and Question Under Discussion (QUD) \citep{wu2023qudeval, wu2024questions}. 
We follow a similar approach to prompt LLMs to answer questions that designed to test different levels of discourse knowledge.

The scope of discourse knowledge evaluation varies widely. Some studies focus on specific theoretical frameworks, such as PDTB-style \citep{webber2019penn} discourse relation recognition \citep{miao2024discursive, yung2024prompting, mehri2025discourse} or SDRT-style \citep{asher2003logics} dialogue parsing \citep{fan2024uncovering}, while others examine how models generalize across frameworks \cite{eichin2025probing}.
Some studies target particular domains, 
such as stylistic discourse in literature \citep{wang2023disco}, or specific relational phenomena, including event relations \citep{wei2024llms} and temporal relations \citep{yuan2023zero, fatemi2024testoftime}.
Another 
line of research focuses on models' generation 
capabilities,
such as DiscoScore \citep{zhao2023discoscore} for coherence assessment, DiSQ \cite{miao2024discursive} for answer faithfulness, and LongBench \citep{bailongwriter2024} and LongGenBench \citep{wulonggenbench2025} for 
long-text generation quality.

Closely related to our work,
DiscoEval \cite{chen2019evaluation} proposes a sentence-level test suite for learning discourse-aware representations, but is limited to local discourse phenomena and binary classification.
Disco-Bench \citep{wang2023disco} evaluates various aspects of discourse understanding 
and generation, but focuses on specific phenomena in Chinese and English.
DiscoTrack \citep{bu2025discotrack} introduces new tasks that probe LLMs both explicitly (e.g., salience recognition) and implicitly (e.g., bridging inference),
but primarily targets local phenomena such as entities.
Unlike prior studies that concentrate on specific frameworks or narrowly defined discourse tasks, our goal is to develop a comprehensive evaluation suite that spans multiple levels of discourse understanding.





\section{BeDiscovER: Motivation and Method}
\label{sec:motivation}

\subsection{Motivation}
\label{subsec:motivation}

Our motivation stems from the convergence of two active research areas: discourse analysis and large language models.
Traditionally, discourse analysis has centered on theoretical frameworks and phenomena such as anaphora resolution 
and textual coherence. More recently, it has expanded toward computational goals: modeling long-distance dependencies, generating long narratives, 
and supporting pragmatic reasoning \citep{parmar2024towards, wu2024rethinking, cao2025pragmatic}.
Meanwhile, LLMs have evolved from sentence-level learners to systems capable of long-document understanding, coherent generation, and advanced reasoning \citep{besta2025reasoning}. This evolution underscores the growing need for discourse-level evaluation to diagnose weaknesses and guide further model development.
Despite this convergence, research on LLMs' discourse understanding remains fragmented across tasks and frameworks (see \cref{sec:related}), offering only a partial view of their true capabilities. 

To address this gap, we introduce a unified benchmark to assess discourse knowledge across multiple dimensions,  
aiming to provide a big picture of the current state of discourse understanding in reasoning-oriented LLMs and their limitations.

\vspace{-1ex}
\subsection{Task Coverage}
\label{subsec:tasks}

BeDiscovER tasks represent well-established discourse evaluation settings in three levels: 
\textbf{I.~Lexicon}: Task (1) Discourse Marker Understanding. 
\textbf{II.~Multi-sentence}: Task (2) Temporal Reasoning and (3) Discourse Relation Recognition. 
\textbf{III.~Document-level structure}: Task (4) Sentence Ordering and (5) Dialogue Discourse Parsing.
Examples for each task are provided in \cref{append-tasks-example}.

Although not exhaustive, our task selection is representative, covering diverse 
knowledge types. 
Each task is supported by multiple datasets, summarized in Table~\ref{tab:overview}.
When different splits exist, we only include test subset.
We carefully curate datasets to capture complementary aspects of discourse phenomena. For example, within Task (2), TimeBank-Dense \citep{cassidy2014tbdense} targets events in adjacent sentences, while TDDiscourse \citep{naik2019tddiscourse} emphasizes long-distance relations. To further enrich this category, we include a subset of Test-of-Time \citep{fatemi2024testoftime} on arithmetic temporal reasoning.

To ensure diversity across languages, genres, and theoretical frameworks, we adopt the discourse relation classification subtask in DISRPT 2025\footnote{We do not incorporate other two tasks in DISRPT: segmentation is widely considered nearly solved (~90\% F1); connective identification is effectively a subset of relation classification (restricted to explicit cases) and thus less challenging.} rather than limiting to a single framework.
The latest DISRPT version introduces a unified relation taxonomy that consolidates over 300 relation types into 17 unified relation types, 
facilitating cross-framework comparison \citep{braud2025disrpt}.
We further incorporate the most recent datasets: 
\textit{Just} \citep{sheffield2025just} and \textit{Otherwise} \citep{liu2025otherwise} for Task~(1). 
These corpora introduce an interesting perspective on polyfunctional discourse particles and their subtle effects on textual meaning, enabling assessment of LLMs' sensitivity to fine-grained pragmatic distinctions.


\subsection{Model and Evaluation}

We test the following reasoning-oriented LLMs: 
    \textbf{(a) Qwen3 series} \citep{yang2025qwen3}: 
    We evaluate both the ``thinking'' mode (high reasoning effort) and ``non-thinking'' (low reasoning effort) mode.
    \textbf{(b) DeepSeek-R1} \cite{guo2025deepseek}: DeepSeek reasoning model. 
    \textbf{(c) DeepSeek-R1-distill-Qwen32B} \citep{guo2025deepseek}: Qwen2.5-32B \citep{qwen2.5} as the base model with direct distillation from DeepSeek-R1. 
    \textbf{(d) GPT-5-mini} \citep{gpt5}: 
    We test ``low'' and ``high'' reasoning effort. 
In addition, we evaluate three widely used LLMs without dedicated reasoning-oriented training: GPT-4o-mini \citep{hurst2024gpt4o}, LLaMA-4-Scout \citep{llama4}, and Qwen2.5-72B-Instruct \citep{qwen2.5}.
Model configurations are in \cref{append:model}.

We adopt QA-based prompting for its flexibility and its ability to support fine-grained interpretability. Precisely, we cast all tasks in an open-ended QA format to enable a unified evaluation pipeline. 
For classification tasks with a fixed label space, the label set is explicitly provided in the system prompt (\cref{append:prompt}).
To mitigate model variation, each experiment is run three times. 
For clarity, we report average accuracies in the main text when applicable, while details (e.g., additional metrics and fine-grained scores per dataset) are provided in \cref{append-results-tables}, along with error analysis in \cref{append:error}.

For open-source reasoning models such as Qwen3 and DeepSeek-R1, their reasoning process is 
observable, whereas in proprietary models like GPT-5, it remains implicit. Nevertheless, we can infer reasoning effort indirectly through the usage of thinking tokens. 
In \cref{append-plot}, we visualize model performance under low (i.e., ``non-thinking'') versus high (i.e., ``thinking'') reasoning effort and present interesting observations.





\begin{table*}[t]
    \centering
    \resizebox{\textwidth}{!}{
    \begin{tabular}{lllccccccccc}
        \toprule
        & & & \multicolumn{3}{c}{\textit{Just-Manual} (acc.)} & \multicolumn{3}{c}{\textit{Just-Subtitle} (acc.)} & \multicolumn{3}{c}{\textit{Otherwise} (acc.)} \\
        \cmidrule(lr){4-6} \cmidrule(lr){7-9} \cmidrule(lr){10-12}
        & Model & Size & Basic & Def & Def+Exp & Basic & Def & Def+Exp & Basic & Def & Def+Exp \\
        \midrule
        \multirow{10}{*}{\rotatebox[origin=c]{90}{\small BeDiscovER}} 
        & Qwen3-1.7B & 1.7B & $38.9_{.7}$ & $40.7_{4.2}$ & $40.2_{2.2}$ & $25.9_{0.9}$ & $26.2_{1.5}$ & $25.5_{1.3}$ & $23.9_{0.6}$ & $21.2_{1.3}$ & $16.2_{1.3}$ \\
        & Qwen3-14B & 14B & $60.2_{.6}$ & $54.9_{2.0}$ & $55.6_{1.4}$ & $43.2_{2.1}$ & $32.9_{1.7}$ & $33.2_{1.3}$ & $32.1_{.7}$ & $\underline{45.2}_{2.5}$ & $51.4_{1.6}$ \\
        & Qwen3-32B & 32B & $59.8_{1.8}$ & $65.6_{3.5}$ & $64.4_{2.0}$ & $48.5_{1.4}$ & $48.9_{2.0}$ & $47.1_{1.2}$ & $34.3_{.4}$ & $\textcolor{red}{\mathbf{49.3}}_{2.6}$ & $\underline{66.1}_{.6}$ \\
        & DS-r1-distill-Qwen & 32B & $56.7_{1.1}$ & $63.7_{1.5}$ & $59.6_{4.0}$ & $31.8_{4.3}$ & $40.5_{2.9}$ & $35.8_{.2}$ & $33.1_{1.5}$ & $32.4_{1.8}$ & $44.3_{3.5}$ \\
        & DeepSeek-r1-0528 & 37/671B & $53.7_{.4}$ & $\underline{66.3}_{.7}$ & $67.0_{1.4}$ & $48.1_{3.1}$ & $\underline{60.2}_{1.8}$ & $59.5_{4.5}$ & $31.2_{.9}$ & $38.9_{1.8}$ & $58.8_{1.7}$ \\
        & GPT-5-mini (low) & Unknown & $\textcolor{red}{\mathbf{63.3}}_{2.2}$ & $65.9_{.3}$ & $65.6_{.0}$ & $\underline{56.4}_{1.4}$  & $58.4_{1.4}$ & $\underline{62.9}_{.5}$ & $\underline{34.6}_{.6}$ & $36.8_{1.1}$ & $51.9_{3.3}$ \\
        & GPT-5-mini (high) & Unknown & $\underline{61.7}_{.6}$ & $\textcolor{red}{\mathbf{66.1}}_{1.7}$ & $65.6_{1.1}$ & $\textcolor{red}{\mathbf{57.4}}_{1.7}$  & $\textcolor{red}{\mathbf{63.1}}_{.0}$ & $\textcolor{red}{\mathbf{63.4}}_{.3}$  & $\textcolor{red}{\mathbf{35.5}}_{.5}$ & $43.9_{1.0}$ & $\textcolor{red}{\mathbf{71.8}}_{1.7}$ \\
        \cdashline{2-12}

        & GPT-4o-mini & Unknown & $51.5_{.4}$ & $60.7_{1.8}$ & $66.8_{3.4}$ & $24.6_{1.8}$  & $29.1_{.2}$ & $28.0_{1.8}$ & $22.0_{.6}$ & $21.4_{1.0}$ & $30.6_{.7}$ \\
        & Llama-4-Scout & 17/109B & $60.6_{.6}$ & $61.7_{.6}$ & $65.0_{.6}$ & $34.6_{.3}$  & $38.9_{.7}$ & $38.9_{.0}$ & $31.6_{.3}$ & $19.2_{.2}$ & $37.6_{1.2}$ \\
        & Qwen2.5-72B & 72B & $61.1_{1.1}$ & $66.7_{1.1}$ & $66.1_{1.7}$ & $32.6_{1.0}$  & $31.2_{1.0}$ & $38.9_{.7}$ & $34.7_{.0}$ & $31.3_{1.0}$ & $32.0_{1.0}$ \\
        \midrule
        
        \multirow{2}{*}{\rotatebox[origin=c]{90}{\small Contin.}}
        & GPT-Neo-1.3B & 1.3B &  & - &  &  & - & & \multicolumn{3}{c}{$56.0^\dagger$ (surprisal score)}\\[0.06cm]
        & Mistral-7B-v0.1 & 7B &  & - &  &  & - &  & \multicolumn{3}{c}{$59.0^\dagger$ (surprisal score)}\\[0.04cm]
        \midrule
        
        \multirow{3}{*}{\rotatebox[origin=c]{90}{\small Prompting}}
        & Mistral-0.3-7B \shortcite{sheffield2025just} & 7B & & & $\underline{69}^*$ & & &  $27^*$ &  & - & \\
        & OLMo2-13B \shortcite{sheffield2025just} & 13B & & & $46^*$ & & & $18^*$ &  & - & \\
        & Llama-3.3-70B \shortcite{sheffield2025just} & 70B & & & $\textcolor{red}{\mathbf{75}}^*$ & & & $35^*$ &  & - & \\
        \bottomrule
    \end{tabular}}
    \caption{
    \textbf{Task (1) Discourse Marker Understanding} performance with BeDiscovER and baselines on \textit{Just-Manual}, \textit{Just-Subtitle}, and \textit{Otherwise} datasets. 
    Top: BeDiscovER scores with reasoning and non-reasoning LLMs. 
    Middle: \textit{Otherwise} dataset baselines \shortcite{liu2025otherwise}, $^\dagger$ denotes surprisal scores which calculates the continuation acceptability in language models. 
    Bottom: prompting baselines in \textit{Just} \shortcite{sheffield2025just}.
    $^*$are \textit{approximated} numbers from the Figure~2 in the original paper. 
    Best score per column is in \textcolor{red}{\textbf{red}} and second best \underline{underlined}.
    }
    \label{tab:just-otherwise}
    \vspace{-1ex}
\end{table*}

\section{Understanding on Discourse Lexicons}
\label{sec:lexical}
\textbf{Task (1) Discourse Marker Understanding} contains two newly curated datasets \textit{Just} (with two subsets: Just-Manual and Just-Subtitle) \citep{sheffield2025just} and \textit{Otherwise} \citep{liu2025otherwise}.
Owing to their polyfunctional nature, 
discourse markers produce subtle yet diverse semantic and discourse effects \citep{lee1987semantics, rohde2016filling, warstadt2020just, beltrama2022just}.
While 
human can reliably distinguish these discourse functions in context \citep{rohdediscourse2018}, the extent to which LLMs exhibit similar sensitivity and proficiency remains unclear.

\subsection{Settings and Baselines}

We design three QA-style prompting formats that vary in the amount of background information provided about the particle’s functions:
(i) \textsc{Basic}: offers no explanation and directly asks the model to select a function label;
(ii) \textsc{Def}: adds a one-sentence definition for each label; and
(iii) \textsc{Def+Exp}: further includes a concrete example, resembling one-shot in-context learning.
Prompt examples 
are provided in \cref{append-prompt-dm}.

We compare with direct prompting results for \textit{Just} and continuation acceptability scores for \textit{Otherwise} in their original papers. \textit{Just} paper adopts a prompting setup similar to our \textsc{Def+Exp}, whereas \textit{Otherwise} employs a continuation acceptability approach: given a left-hand side (LHS) and right-hand side (RHS) clause, the language model evaluates the most plausible connective linking them. These connectives---explicit markers such as \textit{because if not}---capture the distinctive semantics associated with the adverbial \textit{otherwise}.
For evaluation, we report accuracy; 
other metrics (precision, recall, F1 scores) are provided in \cref{append-dm}.

\subsection{Results and Analysis}
\label{subsec:lexical-result}
Table~\ref{tab:just-otherwise} presents main results of \textit{Just-Manual}, \textit{Just-Subtitle}, and \textit{Otherwise} with BeDiscovER (top)
and baselines (middle and bottom parts). 
Regarding prompting strategy, providing richer background information (\textsc{Def} and \textsc{Def+Exp}) generally improves performance, especially for large reasoning models (>32B),
but yields only modest gains for non-reasoning models. This suggests that reasoning LLMs benefit more from in-context learning, whereas non-reasoning models rely more heavily on internal knowledge.
This trend holds across both datasets. However, for smaller and medium-sized models (small: <4B; medium: 4B–32B), additional context does not always help. On the more challenging \textit{Just-Subtitle} and \textit{Otherwise} datasets, excessive contextual information can even degrade performance.
A relevant phenomenon is noted by \citet{huauxiliary2024}, who show that increased task complexity can widen the demand gap for smaller models with limited parameters and training data, leading to reduced accuracy. We also observe a critical model size threshold (>2B parameters), beyond which performance improves sharply—consistent with findings by \citet{sheffield2025just}.
Compared to 
baselines, 
GPT-5-mini with high reasoning effort achieves the best performance, setting new state-of-the-art results on \textit{Just–Subtitle} and \textit{Otherwise}.

\section{Understanding on (multi-)Sentences}
\label{sec:sentencial}

\begin{table}[t]
    \centering
    \resizebox{\columnwidth}{!}{
    \begin{tabular}{lllccc}
        \toprule
         & \multirow{2}{*}{Model} & \multirow{2}{*}{Size} & TBD & TDD-Man & ToT-Ari \\
         & & & (f1) & (f1) & (acc.) \\
        \midrule
        \multirow{10}{*}{\rotatebox[origin=c]{90}{\small BeDiscovER}} 
        & Qwen3-1.7B & 1.7B & $26.2_{0.8}$ & $21.9_{0.6}$ & $46.5_{0.5}$ \\
        & Qwen3-14B & 14B & $39.8_{0.1}$ & $29.5_{0.6}$ & $80.4_{1.2}$ \\
        & Qwen3-32B & 32B & $40.8_{0.5}$ & $31.6_{0.3}$ & $81.6_{0.5}$ \\
        & DS-r1-distill-Qwen & 32B & $23.8_{2.5}$ & $18.6_{3.3}$ & $68.0_{1.2}$ \\
        & DeepSeek-r1-0528 & 671B & $33.1_{0.0}$ & $22.2_{0.0}$ & $63.3_{0.5}$ \\
        & GPT-5-mini (low) & Unk. & $\underline{38.7}_{0.2}$ & $\underline{34.4}_{0.2}$ & $\underline{87.2}_{0.1}$ \\
        & GPT-5-mini (high) & Unk. & $\textcolor{red}{\mathbf{43.4}}_{0.3}$ & $\textcolor{red}{\mathbf{36.7}}_{0.3}$ & $\textcolor{red}{\mathbf{88.2}}_{0.0}$ \\
        \cdashline{2-6}
        
        & GPT-4o-mini & Unk. & $27.8_{.3}$ & $25.0_{.2}$ & $31.4_{.1}$  \\
        & Llama-4-Scout & 109B & $29.3_{.1}$ & $27.0_{.2}$ & $30.4_{.2}$  \\
        & Qwen2.5-72B & 72B & $33.2_{.7}$ & $28.4_{.3}$ & $30.4_{.2}$  \\
        
        \midrule
        \multirow{4}{*}{\rotatebox[origin=c]{90}{\small Prompting}} 
        & ChatGPT \shortcite{yuan2023zero} & Unk. & $37.0$ & $24.3$ & -  \\
        & ChatGPT \shortcite{chan2024Exploring} & Unk. & $27.0$ & $16.8$ & -\\
        & GPT-4 \shortcite{fatemi2024testoftime} & Unk. & - & - & $54.2$ \\
        & Llama3.3-70B \shortcite{fan2025consistent} & 70B & - & $26.8$ & - \\
        \midrule
        \multirow{5}{*}{\rotatebox[origin=c]{90}{\small Supervised}} 
        & UCGraph \shortcite{liu2021discourse} & 125B & $59.1$ & $43.4$ & - \\
        & RSGT \shortcite{zhou2022rsgt} & 355M & $\underline{68.7}$ & - & - \\
        & DTRE \shortcite{wang2022dct} & 125M & $\textcolor{blue}{\mathbf{70.2}}$ & $56.3$ & - \\
        & CPTRE \shortcite{yuan2024temporal} & 125M & $61.1$ & $\underline{56.5}$ & -\\
        & FT Llama3.3 \shortcite{fan2025consistent} & 70B & - & $\textcolor{blue}{\mathbf{57.9}}$ & -\\
        \bottomrule
    \end{tabular}}
    \caption{
    \textbf{Task (2) Temporal Reasoning} performance: BeDiscovER (top), LLM-prompting baselines (middle), and supervised baselines (bottom) on TBD, TDD-Man, and ToT-arithmetic datasets. Scores are micro-F1.
    In the first four supervised baselines, ``size'' refers to the parameter count of the PLM encoders (BERT and RoBERTa).
    The last supervised baseline uses a fine-tuned Llama3.3-70B model.
    Best unsupervised scores are in \textcolor{red}{\textbf{red}} and best supervised scores in \textcolor{blue}{\textbf{blue}}.
    }
    \label{tab:temporal}
    \vspace{-1ex}
\end{table}

\begin{table}[t]
    \centering
    \resizebox{\columnwidth}{!}{
    \begin{tabular}{llcccccc}
        \toprule
        & Model
        & DEP & eRST & ISO & PDTB & RST & SDRT \\
        \midrule
        \multirow{10}{*}{\rotatebox[origin=c]{90}{\small BeDiscovER}} 
        & Qwen3-1.7B
        & $21.2$ & $19.7$ & $25.0$ & $22.1$ & $18.0$ & $18.7$ \\
        & Qwen3-14B 
        & $37.9$ & $32.7$ & $41.4$ & $36.4$ & $33.6$ & $26.2$ \\
        & Qwen3-32B 
        & $39.5$ & $34.7$ & $40.3$ & $36.8$ & $35.5$ & $29.9$ \\
        & DS-r1-distill-Qwen 
        & $32.6$ & $29.9$ & $38.7$ & $29.8$ & $32.9$ & $24.6$ \\
        & DeepSeek-r1-0528
        & $44.3$ & $\textcolor{red}{\mathbf{36.6}}$ & $\underline{51.4}$ & $\underline{43.7}$ & $\underline{43.6}$ & $26.6$ \\
        & GPT-5-mini (low)
        & $\underline{46.1}$ & $33.9$ & $47.3$ & $42.9$ & $42.3$ & $\underline{34.2}$ \\
        & GPT-5-mini (high) 
        & $\textcolor{red}{\mathbf{51.2}}$ & $\underline{36.0}$ & $\textcolor{red}{\mathbf{56.8}}$ & $\textcolor{red}{\mathbf{47.4}}$ & $\textcolor{red}{\mathbf{46.6}}$ & $\textcolor{red}{\mathbf{37.9}}$ \\
        
        \cdashline{2-8}
        & GPT-4o-mini & $28.3$ & $25.9$ & $28.6$ & $31.2$ & $24.2$ & $25.3$ \\
        & Llama-4-Scout & $43.2$ & $34.3$ & $44.8$ & $36.0$ & $36.6$ & $25.3$ \\
        & Qwen2.5-72B & $44.9$ & $34.0$ & $37.6$ & $31.9$ & $34.7$ & $26.4$ \\
        
        \midrule        
        \multirow{5}{*}{\rotatebox[origin=c]{90}{\small Supervised}} 
        & DeDisCo \shortcite{ju2025dedisco} & $\textcolor{blue}{\mathbf{77.2}}$ & $\textcolor{blue}{\mathbf{71.8}}$ & $\textcolor{blue}{\mathbf{72.0}}$ & $\textcolor{blue}{\mathbf{79.0}}$ & $\textcolor{blue}{\mathbf{64.9}}$ & $\textcolor{blue}{\mathbf{83.0}}$ \\
        & HITS \shortcite{banerjee2025hits} & $74.1$ & $\underline{64.5}$ & $\underline{72.0}$ & $\underline{76.3}$ & $\underline{61.9}$ & $\underline{82.1}$ \\
        & DisCreT \shortcite{pujol2025discut} & $72.3$ & $58.1$ & $60.0$ & $75.3$ & $56.4$ & $77.5$ \\
        & CLAC \shortcite{turk2025clac} & $\underline{74.9}$ & $57.5$ & $54.8$ & $74.7$ & $56.2$ & $77.4$ \\
        & SeCoRel \shortcite{lalitha2025secorel} & $69.7$ & $53.8$ & $52.8$ & $70.6$ & $52.5$ & $76.4$ \\
        \bottomrule
    \end{tabular}}
    \caption{
    \textbf{Task (3): Discourse Relation Recognition [framework-split]}. BeDiscovER results (top) vs. supervised systems (bottom; all <4B parameters).
    Framework abbreviations: DEP -- Dependency Structure \shortcite{yang2018scidtb}; eRST -- Enhanced RST \shortcite{zeldes2025erst}; ISO -- ISO Framework \shortcite{bunt:iso:2016}; PDTB -- Penn Discourse TreeBank \shortcite{prasad2005penn}; RST -- Rhetorical Structure Theory \shortcite{mann1988rhetorical}; 
    SDRT -- Segmented Discourse Representation Theory \shortcite{asher2003logics}.
    }
    \label{tab:dr-2}
    \vspace{-1ex}
\end{table}

\subsection{Temporal Relation}
\label{subsec:temporal}

\textbf{Task~(2) Temporal Reasoning} requires EVENT-EVENT pair reasoning across sentences, both adjacent pairs as in TBD \citep{cassidy2014tbdense}) and distant as in TDD-Man \citep{naik2019tddiscourse}).
We also include Test-of-Time (ToT) \citep{fatemi2024testoftime} to test arithmetic skills in temporal reasoning.

\subsubsection{Settings and Baselines}
In TBD and TDD-Man, we use full news documents with target events annotated by <EVENT></EVENT> tags. 
Our prompt, adapted from the best-performing template in \citet{yuan2023zero}, presents 
relation labels as a multiple-choice QA task (see \cref{append-prompt-tr}).

For baselines, we consider both prompting-based baselines using LLMs, such as ChatGPT \citep{yuan2023zero, chan2024Exploring}, GPT-4 \citep{fatemi2024testoftime}, and Llama \citep{fan2025consistent}, as well as supervised approaches, including graph-based neural networks UCGraph, RSGT, and DTRE \citep{liu2021discourse, zhou2022rsgt, wang2022dct}, prototypical network CPTRE \citep{yuan2024temporal}, 
and a fine-tuned Llama 
\citep{fan2025consistent}.
Most supervised baselines use much smaller pretrained language models such as BERT \citep{devlin2019bert} and RoBERTa \citep{liu2019roberta} for node/label representation learning. 





\subsubsection{Results and Analysis}
\label{subsec:tr-results}

Table~\ref{tab:temporal} summarizes the main results across all three datasets, with details provided in \cref{append-tr}.
Overall, recent reasoning LLMs substantially outperform earlier models such as ChatGPT and non-reasoning counterparts, achieving gains of 10\textendash20\% on TBD and TDD-Man, and a notable 30-50\% improvement on ToT-Arithmetic. These results indicate enhanced temporal reasoning capability, particularly in arithmetic-related temporal inference.
However, compared with supervised baselines, zero-shot LLMs still underperform on TBD and TDD-Man. 
Both datasets involve full documents containing multiple events, requiring LLMs to reconstruct the underlying narrative to determine the correct event order (e.g., an \textit{investigation} typically follows a \textit{crime}): a task that remains challenging even for advanced reasoning models.

In an ablation study, we simplify inputs by limiting context to sentences containing the target events. This pre-selection reduces reasoning load and yields modest gains on TBD (2\textendash3\%; \cref{tab:app-2-1-1,tab:app-2-1-2}). Conversely, for TDD-Man, where events are farther apart, removing intermediate context leads to a 2\textendash9\% drop in performance (\cref{tab:app-2-2-1,tab:app-2-2-2}). These findings suggest that LLM reasoning is sensitive to both context complexity and the availability of relevant information.

\begin{table*}[t]
    \centering
    \resizebox{\textwidth}{!}{%
    \begin{tabular}{llcccccccccccccc}
    \toprule
        & \multirow{2}{*}{Model} &
        \multicolumn{2}{c}{AAN abstract} & \multicolumn{2}{c}{ArXiv abstract} & \multicolumn{2}{c}{Neurips abstract} & \multicolumn{2}{c}{NSF abstract} & \multicolumn{2}{c}{ROC stories} & \multicolumn{2}{c}{SIND} & \multicolumn{2}{c}{Wiki. mv. plts.} \\
        \cmidrule(lr){3-4} \cmidrule(lr){5-6} \cmidrule(lr){7-8} \cmidrule(lr){9-10} \cmidrule(lr){11-12} \cmidrule(lr){13-14} \cmidrule(lr){15-16}
        & & PMR & Acc & PMR & Acc & PMR & Acc & PMR & Acc & PMR & Acc & PMR & Acc & PMR & Acc \\
        \midrule
        \multirow{10}{*}{\rotatebox[origin=c]{90}{\small BeDiscovER}} 
        & Qwen3-1.7B &
        $21.1_{.5}$ & $38.0_{.5}$ & $19.0_{.5}$ & $31.4_{.3}$ & $11.0_{2.2}$ & $30.8_{.7}$ & $6.6_{0.5}$ & $14.7_{.1}$ & $17.1_{.7}$ & $44.7_{.9}$ & $14.6_{1.0}$ & $35.6_{1.0}$ & $8.9_{.3}$  & $12.7_{.1}$ \\
        & Qwen3-14B &
        $55.5_{.9}$ & $70.6_{.3}$ & $43.4_{.3}$ & $59.1_{.6}$ & $43.0_{.8}$ & $66.1_{.1}$ & $14.3_{.0}$ & $29.2_{.6}$ & $67.3_{.4}$ & $82.5_{.4}$ & $31.2_{.9}$ & $55.1_{.4}$ & $18.2_{.1}$ & $27.1_{.2}$ \\
        & Qwen3-32B &
        $59.6_{.4}$ & $74.0_{.3}$ & $46.1_{.4}$ & $61.9_{.4}$ & $44.2_{.4}$ & $68.6_{.2}$ & $14.3_{.3}$ & $30.2_{.4}$ & $70.6_{.7}$ & $84.4_{.3}$ & $31.2_{.5}$ & $55.3_{.3}$ & $19.1_{.6}$ & $29.1_{.5}$ \\
        & DS-r1-distill-Qwen &
        $51.4_{.9}$ & $67.7_{.5}$ & $38.8_{.4}$ & $55.4_{.2}$ & $38.9_{.4}$ & $63.5_{.7}$ & $11.4_{.2}$ & $26.0_{.2}$ & $63.2_{2.8}$ & $79.8_{1.6}$ & $29.0_{1.8}$ & $53.3_{1.2}$ & $17.6_{.6}$ & $24.3_{.5}$ \\
        & DeepSeek-r1-0528 &
        $\underline{64.8}_{.4}$ & $\underline{75.5}_{.2}$ & $46.6_{.5}$ & $61.9_{.3}$ & $\underline{56.8}_{.5}$ & $\underline{74.9}_{.6}$ &  $15.7_{.2}$ & $28.5_{.2}$ & $\textcolor{red}{\mathbf{77.1}}_{1.2}$ & $\textcolor{red}{\mathbf{88.2}}_{.6}$ & $\textcolor{red}{\mathbf{36.4}}_{1.3}$ & $\textcolor{red}{\mathbf{58.4}}_{.9}$ & $\textcolor{red}{\mathbf{23.8}}_{.4}$ & $28.0_{.4}$ \\
        & GPT-5-mini (low) &
        $61.1_{.6}$ & $75.4_{.0}$ & $\underline{49.3}_{.1}$ & $\underline{65.6}_{.4}$ & $50.5_{.4}$ & $73.1_{.1}$ & $\underline{15.8}_{.1}$ & $\underline{33.2}_{.4}$ & $68.3_{1.1}$ & $83.6_{.6}$ & $31.5_{1.1}$ & $54.6_{.8}$ & $20.3_{.1}$ & $\underline{32.0}_{.2}$ \\
        & GPT-5-mini (high) &
        $\textcolor{red}{\mathbf{66.3}}_{.7}$ & $\textcolor{red}{\mathbf{78.7}}_{.1}$ & $\textcolor{red}{\mathbf{55.0}}_{.1}$ & $\textcolor{red}{\mathbf{69.8}}_{.1}$ & $\textcolor{red}{\mathbf{58.6}}_{.3}$ & $\textcolor{red}{\mathbf{78.6}}_{.2}$ & $\textcolor{red}{\mathbf{17.8}}_{.2}$ & $\textcolor{red}{\mathbf{35.5}}_{.1}$ & $\underline{75.7}_{1.0}$ & $\underline{87.3}_{.04}$ & $\underline{35.0}_{1.5}$ & $\underline{58.2}_{1.0}$ & $\underline{23.0_{.5}}$ & $\textcolor{red}{\mathbf{40.1}}_{.4}$ \\

        \cdashline{2-16}
        & GPT-4o-mini &
        $37.2_{.7}$ & $54.7_{.1}$ & $30.2_{.4}$ & $43.9_{.2}$ & $22.8_{.0}$ & $49.1_{.8}$ & $10.5_{0.1}$ & $21.1_{.1}$ & $47.3_{.1}$ & $69.8_{.4}$ & $25.6_{.2}$ & $50.2_{.1}$ & $14.3_{.1}$  & $19.4_{.2}$ \\
        & Llama-4-Scout &
        $42.9_{.0}$ & $61.9_{.1}$ & $33.6_{.0}$ & $51.2_{.1}$ & $31.4_{.7}$ & $58.1_{.1}$ & $10.5_{0.3}$ & $26.2_{.1}$ & $49.1_{.3}$ & $71.8_{.2}$ & $27.5_{.3}$ & $50.9_{.2}$ & $16.1_{.3}$  & $22.3_{.3}$ \\
        & Qwen2.5-72B &
        $55.2_{.6}$ & $71.1_{.2}$ & $41.3_{.1}$ & $58.6_{.2}$ & $48.3_{.8}$ & $70.7_{.1}$ & $15.0_{0.5}$ & $30.7_{.2}$ & $66.4_{.5}$ & $81.9_{.1}$ & $31.9_{.2}$ & $55.8_{.2}$ & $19.7_{.0}$  & $28.8_{.3}$ \\

        \midrule
        \multirow{4}{*}{\rotatebox[origin=c]{90}{\small Supervised}} 
        & BERSON \shortcite{cui2020bert} &
        $59.8$ & $78.0$ & $56.1$ & $75.1$ & $48.0$ & $73.9$ & $23.1$ & $50.2$ & $68.2$ & $82.9$ & $31.7$ & $58.9$ & - & - \\
        & Re-BART \shortcite{chowdhury2021everything} &
        $\underline{73.5}$ & $\underline{84.3}$ & $\underline{62.4}$ & $\underline{74.3}$ & $\underline{57.0}$ & $\underline{77.4}$ & $\underline{29.7}$ & $\underline{50.2}$ & $\underline{81.9}$ & $\underline{90.8}$ & $\underline{43.2}$ & $\underline{64.5}$ & $\textcolor{blue}{\mathbf{25.8}}$ & $\textcolor{blue}{\mathbf{42.0}}$ \\
        & \textsc{CoVer} \shortcite{jia2023sentence} &
        $59.2$ & $78.1$ & - & - & $50.9$ & $74.9$ & - & - & $72.3$ & $84.8$ & $33.0$ & $60.3$ & - & - \\
        & NAON-BART \shortcite{bin2023non} & 
        $\textcolor{blue}{\mathbf{73.9}}$ & $\textcolor{blue}{\mathbf{87.2}}$ & $\textcolor{blue}{\mathbf{62.6}}$ & $\textcolor{blue}{\mathbf{79.2}}$ & $\textcolor{blue}{\mathbf{61.2}}$ & $\textcolor{blue}{\mathbf{84.2}}$ & $\textcolor{blue}{\mathbf{30.6}}$ & $\textcolor{blue}{\mathbf{54.8}}$ & $\textcolor{blue}{\mathbf{89.1}}$ & $\textcolor{blue}{\mathbf{95.1}}$ & $\textcolor{blue}{\mathbf{55.6}}$ & $\textcolor{blue}{\mathbf{79.6}}$ & - & -  \\
        \bottomrule
    \end{tabular}}
    \caption{
    \textbf{Task (4) Sentence Ordering} main results. 
    We present perfect match rate (PMR) and accuracy, with additional metrics in \cref{append-so}. 
    Best unsupervised scores are in \textcolor{red}{\textbf{red}} and 
    best supervised scores in \textcolor{blue}{\textbf{blue}}.
    }
    \label{tab:so}
    \vspace{-2ex}
\end{table*}

\begin{figure}[t]
    \centering
    \includegraphics[width=\linewidth]{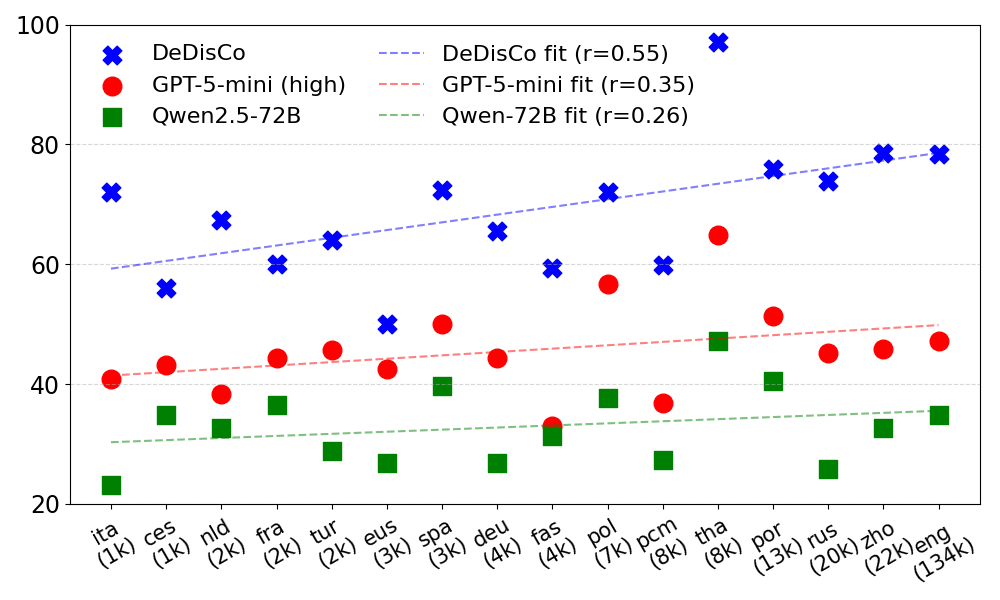}
    \vspace{-20pt}
    \caption{\textbf{Task (3): Discourse Relation Recognition [language-split]}. Best supervised (DeDisCo), reasoning and non-reasoning models (GPT-5-mini, Qwen2.5-72B), with languages ordered by increasing training size (Italian, Czech, Dutch, French, Turkish, Basque, Spanish, German, Persian, Polish, Nigerian Pidgin, Thai, Portuguese, Russian, Chinese, English).
    }
    \label{fig:dr-1}
    \vspace{-1ex}
\end{figure}

\subsection{Discourse Relation}
\label{subsec:dis-relation}


\textbf{Task (3): Discourse Relation Recognition} adopts data from the latest version of DISRPT Shared Task, which provides a unified label set of 17 relations across 38 datasets \citep{braud2025disrpt}. The task requires predicting a rhetorical relation (e.g., \textit{contrast}) between two segments, Arg1 and Arg2.

\subsubsection{Settings and Baselines}
We enrich the input prompt with additional features, including language, framework, corpus name, argument direction (Arg1$\rightarrow$Arg2 or inverse), and context in which the arguments appear, motivated by prior studies \citep{gessler-etal-2021-discodisco, metheniti2024feature}.
Our prompting setup closely follows DeDisCo \citep{ju2025dedisco} (see \cref{append-prompt-dr}), the top-performing system in the DISRPT25 competition. Despite the multilingual data, we use English for all prompt instructions.
All baseline systems are supervised and trained with a model size cap of 4B parameters.

\subsubsection{Results and Analysis}
\label{subsec:dr-results}

Tables~\ref{tab:dr-2} and Figure~\ref{fig:dr-1} give the framework-view and language-view results, respectively.
We show detailed scores for language-view and dataset-view in
\cref{tab:app-3-0,tab:app-3-1-1,tab:app-3-1-2,tab:app-3-1-3,tab:app-3-1-4} in \cref{append-dr}.

All reasoning LLMs perform worse than supervised baselines, with an average gap of about 20 points, while non-reasoning models consistently trail reasoning models.
Among the frameworks, eRST yields the lowest scores---likely because the framework is relatively new and its theoretical foundations have seen limited exposure in modern LLM---with even GPT-5-mini achieving only the low 30s. The ISO framework attains the highest LLM performance at $57\%$, while remains well below supervised results ($72\%$). 
In contrast, all fine-tuned systems perform markedly better, with DeDisCo reaching $70$\textendash$80\%$ across frameworks.

From a language perspective (Figure~\ref{fig:dr-1}), BeDiscovER models exhibit nearly flat (and low) performance across languages ($r=0.2$\textendash$0.3$, not significant), whereas supervised models show clear disparities between high-resource (e.g., Chinese (zho), English (eng)) and low-resource languages (e.g., Czech (ces), Basque (eus)). 
This pattern suggests that LLMs possess a certain degree of cross-lingual generalizability, but still lack robust internal representations of discourse relations, as evidenced by the low overall accuracy.
Interestingly, two mid-resource languages ($\approx$8k): Thai (tha) and Nigerian Pidgin (pcm), show completely contrasting outcomes both in all models. This difference likely reflects divergent discourse properties: the Thai corpus is annotated heavily with explicit discourse connectives \citep{prasertsom2024thai}, making the task easier, 
whereas Nigerian Pidgin has 
highly flexible syntax and fewer explicit connectives \citep{scholman2025disconaija}, increasing task difficulty.


\begin{table*}[t]
    \centering
    \resizebox{.87\textwidth}{!}{%
    \begin{tabular}{llccccccc}
        \toprule
        & \multirow{2}{*}{Model} & \multirow{2}{*}{Size} & \multicolumn{2}{c}{STAC} & \multicolumn{2}{c}{Molweni} & \multicolumn{2}{c}{MSDC} \\
        \cmidrule(lr){4-5} \cmidrule(lr){6-7} \cmidrule(lr){8-9}
        & & & Link & Full & Link & Full & Link & Full \\
        
        \midrule
        \multirow{10}{*}{\rotatebox[origin=c]{90}{\small BeDiscovER}} 
        & Qwen3-1.7B & 1.7B & $35.9_{1.8}$ & $5.2_{0.0}$ & $43.1_{0.3}$ & $5.3_{0.3}$ & $36.7_{2.8}$ & $3.6_{0.2}$ \\
        & Qwen3-14B & 14B & $61.3_{0.8}$ & $29.8_{1.4}$ & $\underline{60.1}_{0.6}$ & $25.7_{0.7}$ & $\textcolor{red}{\mathbf{72.7}}_{0.2}$ & $22.6_{0.1}$ \\
        & Qwen3-32B & 32B & $62.4_{1.1}$ & $31.3_{0.3}$ & $58.3_{0.6}$ & $22.2_{0.9}$ & $\underline{70.5}_{0.1}$ & $23.0_{0.3}$ \\
        & DS-r1-distill-Qwen & 32B & $61.5_{0.7}$ & $22.3_{2.3}$ & $58.6_{0.7}$ & $16.1_{0.3}$ & $69.6_{0.4}$ & $25.7_{1.1}$ \\
        & DeepSeek-r1-0528 & 37/671B & $\textcolor{red}{\mathbf{66.3}}_{0.1}$ & $\underline{38.7}_{0.5}$ & $56.7_{0.7}$ & $22.3_{1.3}$ & $69.6_{0.7}$ & $\textcolor{red}{\mathbf{34.2}}_{0.7}$ \\
        & GPT-5-mini (low) & Unk. & $61.1_{0.3}$ & $32.1_{0.8}$ & $55.6_{0.1}$ & $\underline{27.7}_{0.3}$ & $66.1_{0.1}$ & $26.8_{0.4}$ \\
        & GPT-5-mini (high) & Unk. & $\underline{66.0}_{0.2}$ & $\textcolor{red}{\mathbf{38.8}}_{0.7}$ & $58.4_{0.6}$ & $\textcolor{red}{\mathbf{30.2}}_{0.3}$ & $66.4_{0.7}$ & $\underline{31.9}_{0.5}$ \\
        
        \cdashline{2-9}
        & GPT-4o-mini & Unk. & $53.9_{.1}$ & $10.3_{.1}$ & $49.7_{.3}$ & $13.9_{.2}$ & $56.3_{.2}$ & $23.6_{.0}$ \\
        & Llama-4-Scout & 17/109B & $57.7_{.2}$ & $13.7_{.2}$ & $58.4_{.1}$ & $11.0_{.1}$ & $70.2_{.1}$ & $17.3_{.1}$ \\
        & Qwen2.5-72B & 72B & $58.8_{.6}$ & $15.4_{.0}$ & $55.7_{.1}$ & $12.9_{.4}$ & $64.6_{.1}$ & $18.1_{.2}$ \\
        
        \midrule
        \multirow{2}{*}{\rotatebox[origin=c]{90}{\small Prompt}} 
        & ChatGPT zero-shot \shortcite{chan2024Exploring} & Unk. & $20.0$ & $4.4$ & $28.3$ & $5.4$ & - & - \\
        & ChatGPT prompt engineering \shortcite{fan2024uncovering} & Unk. & $59.9$ & $25.3$ & $\textcolor{red}{\mathbf{63.7}}$ & $23.8$ & - & - \\
        
        \midrule
        \multirow{4}{*}{\rotatebox[origin=c]{90}{\small Supervised}}
        & RoBERTa+pointer \shortcite{liu2021improving} & 125M & $72.9$ & $57.0$ & $\underline{79.0}$ & $55.4$ & - & - \\
        & RoBERTa+CLE \shortcite{chi2022structured} & 125M & $\underline{73.0}$ & $\underline{58.1}$ & $81.0$ & $\underline{58.6}$ & - & -\\
        & T0+transition \shortcite{li2024dialogue} & 3B & $72.3$ & $56.6$ & $\textcolor{blue}{\mathbf{83.4}}$ & $\textcolor{blue}{\mathbf{60.0}}$ & - & - \\
        & Llama3 \shortcite{thompson2024llamipa} & 8B & $\textcolor{blue}{\mathbf{77.5}}$ & $\textcolor{blue}{\mathbf{60.7}}$ & - & - & $\textcolor{blue}{\mathbf{88.3}}$ & $\textcolor{blue}{\mathbf{79.5}}$ \\
        \bottomrule
    \end{tabular}}
    \caption{
    \textbf{Task (5) Dialogue Discourse Parsing} F1 scores on STAC, Molweni, and MDSC: BeDiscovER (top), LLM-based prompting (middle) and supervised baselines (bottom).
    Best unsupervised scores are in \textcolor{red}{\textbf{red}} and best supervised scores in \textcolor{blue}{\textbf{blue}}.
    }
    \label{tab:ddp}
    \vspace{-1ex}
\end{table*}

\section{Understanding on Document Structure}
\label{sec:documental}

\subsection{Sentence Ordering}

\textbf{Task (4) Sentence Ordering} reorders a set of shuffled sentences into a coherent text. It has been used to assess a model's understanding of causal and temporal relations \citep{barzilay2008modeling} and shown to aid discourse structure extraction \citep{li2023discourse}.
We include, in this task, 7 publicly available datasets from two domains: scientific paper abstracts (AAN, ArXiv, Neurips, NSF \citep{wang2018aan, chen2016arxiv, logeswaran2018neurips-nsf}) and narratives (ROC stories, SIND, Wiki movie plots \citep{mostafazadeh2016roc, huang2016sind}), with examples in \cref{append-tasks-example}.

\subsubsection{Settings and Baselines}
In our setup, we adopt a text-to-marker format that prompts the model to generate a sequence of reordered sentence labels (e.g., <s3>, <s1>) as output, following ReBART \citep{chowdhury2021everything} (prompts in \cref{append-prompt-so}).

Supervised baselines use PLMs to encode sentences, followed by either a Pointer Network to predict sentence positions (e.g., BERSON and \textsc{CoVer} \citep{cui2020bert, jia2023sentence}) or a sequence-to-sequence model that generates the position sequence autoregressively, such as Re-BART \citep{chowdhury2021everything}, or non-autoregressively: 
NAON-BART \citep{bin2023non}.

\subsubsection{Results and Analysis}
\label{subsec:so-results}

Table~\ref{tab:so} presents the performance of BeDiscovER and supervised models across seven datasets.
A clear scaling trend emerges where larger LLMs consistently outperform smaller ones. 
In scientific abstract domain, GPT-5-mini (high) achieves competitive results, potentially benefiting from exposure to similar text during pretraining, though there remains room for improvement on long-input dataset such as NSF.
In narrative datasets, GPT-5-mini and DeepSeek-R1 perform comparably, matching supervised performance on Wiki Movie, a challenging dataset with the longest document length.
Overall, these results suggest that reasoning LLMs, especially larger ones, possess a good grasp of textual coherence, possibly stemming from their extensive pretraining on narrative-rich corpora.

\subsection{Dialogue Discourse Parsing}
\label{sec:ddp}

\textbf{Task (5): Dialogue Discourse Parsing} focuses on 
structure construction
across the entire document. We additionally target dialogue rather than monologue to introduce greater genre diversity.
This task includes three widely used dialogue discourse datasets: STAC \citep{asher2016discourse}, Molweni \citep{li2020molweni}, and the recently annotated MSDC based on Minecraft Dialogue Corpus \citep{thompson2024discourse}, all datasets in the SDRT framework \citep{asher2003logics}. 

\subsubsection{Settings and Baselines}

Directly prompting LLMs to generate SDRT-style discourse graphs from raw text remains challenging, as shown in prior studies with ChatGPT \citep{chan2024Exploring, fan2024uncovering}.

Following the parsing-as-generation paradigm \citep{li2024dialogue}, we instead prompt the model to produce relation triples incrementally, where model predicts relations for each new utterance and appends the generated structure as context for subsequent predictions (prompt given in \cref{append-prompt-ddp}).
This incremental approach has achieved state-of-the-art performance with both sequence-to-sequence models (e.g., T0+Transition) and decoder-only architectures such as Llamipa \citep{thompson2024llamipa}.
For comparison, we also include earlier baselines employing smaller PLMs with task-specific decoding strategies \citep{liu2021improving, chi2022structured}.
For evaluation, we employ 
the micro-averaged F1 scores for link attachment (Link) and full structure (Full).


\subsubsection{Results and Analysis}
\label{subsec:ddp-results}

Table~\ref{tab:ddp} reports BeDiscovER performance using incremental prompting. This approach establishes a new state of the art using zero-shot prompting, achieving a $7$\textendash$13\%$ improvement in full structure prediction over prior ChatGPT results \citep{chan2024Exploring, fan2024uncovering}.
Despite these gains, all reasoning and non-reasoning LLMs still trail supervised baselines by a wide margin of $20$\textendash$30\%$, particularly on full structure prediction, which requires accurate identification of both discourse links and their relations. The relation component appears to be the primary bottleneck, with relation recognition in the SDRT scores below $40$ (discussed in \cref{subsec:dr-results}). 

In an oracle setting, where gold triples are provided at each step (see ``single-turn'' setup in \cref{append-ddp}), we observe only modest improvements over auto-regressive parsing, suggesting limited impact from error propagation.
Notably, models with higher reasoning effort outperform their lower-effort counterparts, especially on full-structure prediction, indicating that explicit reasoning enhances discourse parsing performance.

\section{Inter-task Correlation Analysis}


\begin{figure}
    \centering
    \includegraphics[width=\linewidth]{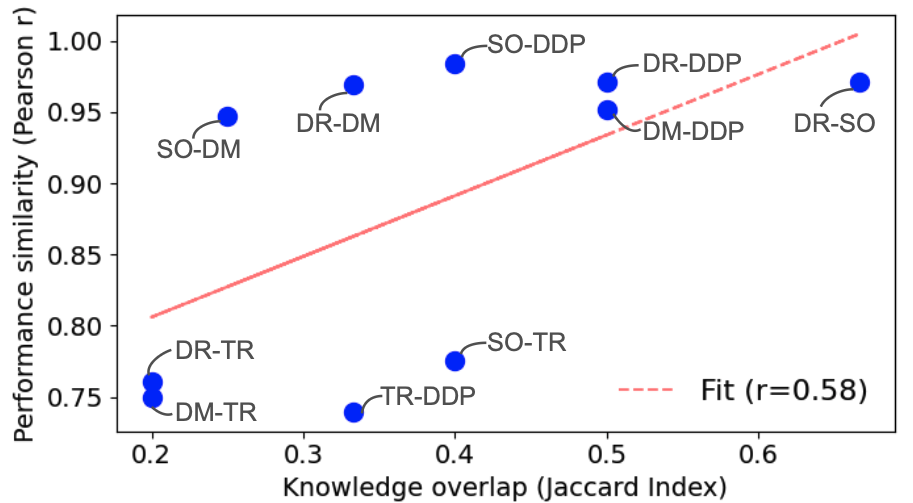}
    \caption{The correlation between sub-task performance similarity (Pearson r)  vs. sub-task knowledge overlap, calculated using 7 reasoning LLMs scores). Task abbreviations: DM: discourse marker, TR: temporal reasoning, DR: discourse relation, SO: sentence ordering, DDP: dialogue discourse parsing.}
    \label{fig:correlation}
    \vspace{-2ex}
\end{figure}

We analyze the correlation among the five subtasks in BeDiscovER: for each task, we aggregate results (accuracies) across all datasets. This yields one task score per model. We take data points from the seven reasoning-oriented LLMs.
The pairwise correlations across all subtasks exhibit strong positive relationships ($r\in [0.73, 0.98]$, $p<0.05$).

To understand what drives the strong performance correlations among tasks, we further examine the discourse knowledge associated with each subtask (e.g., lexical, temporal, rhetorical), as shown in the last column of Table~\ref{tab:overview}. 
Although this mapping from tasks to required knowledge is not exhaustive, it provides a useful basis for our analysis. To quantify how much knowledge two subtasks share, we compute their Jaccard overlap:
$\text{Jaccard(A,B)}=\frac{|A\cup B|}{|A\cap B|}$,
where the intersection indicates number of shared knowledge types, and union the number of knowledge types required by at least one of the tasks. Naturally, a higher overlap score means the two tasks rely on more of the same underlying knowledge.

Figure~\ref{fig:correlation} shows the relationship between task performance similarity ($r$) and knowledge overlap (Jaccard index). We find a moderate positive correlation between knowledge overlap and performance similarity ($r' = 0.58$), suggesting that shared underlying knowledge types explain why some subtasks exhibit similar performance. For example, dialogue discourse parsing (DDP) task shares substantial knowledge with discourse relation classification (DR) and discourse marker understanding (DM), leading to high performance similarity, whereas temporal reasoning (TR) appears to rely on largely distinct knowledge and shows lower similarity.
This corroborates some findings in the recent work on discursive circuits \citep{miao2025discursive}, where different frameworks show a consistent use of linguistic sub-skills in discourse relation.
Additionally, we find that performance similarity is not determined solely by the knowledge taxonomy: the sentence ordering (SO) and DDP tasks exhibit the highest correlation (0.98) despite only moderate overlap (0.4). This likely reflects that both are document-level tasks requiring modeling of global structure in long context---a factor that is not captured by the knowledge types.
Overall, these findings are intriguing and motivate future work to extend BeDiscovER with tasks that cover a broader range of knowledge types.
\section{Conclusion and Discussion}


We introduce BeDiscovER, a comprehensive benchmark designed to evaluate the discourse knowledge of reasoning-oriented LLMs. Covering 3 levels, 5 tasks, and 52 datasets, 
BeDiscovER offers a linguistically grounded and multifaceted assessment of how current LLMs understand discourse structure and coherence.

BeDiscovER also serves as a baseline and reference point for future research on discourse-aware LLMs. Our findings highlight persistent challenges, particularly in temporal reasoning (logical not arithmetic) and rhetorical relation recognition where LLMs still lag behind humans and supervised systems. 
Arguably, these shortcomings may stem from limited exposure to (explicit) structural phenomena during pre-/post-training and from the difficulty of transferring such abstract reasoning skills.
From an interpretability perspective, the diverse tasks in BeDiscovER also offer a rich testbed for probing the internal mechanisms of LLMs, complementing recent work such as CausalGym \citep{arora2024causalgym} and discursive circuits \cite{miao2025discursive}.

As an initial step toward deeper insight, we visualize the relationship between thinking effort and performance (\cref{append-plot}).
On ToT arithmetic temporal reasoning, LLMs' performances did consistently improve with longer reasoning traces.
However, across many other tasks, longer reasoning traces \textbf{do not} necessarily yield better outcomes. Models often become more verbose without producing more meaningful reasoning. 
Improving the quality of reasoning thus remains a promising direction for advancing discourse-aware LLMs.

\section*{Limitations}


BeDiscovER is designed to provide a broad, high-level overview of recent LLM performance across diverse discourse tasks.
Although we made considerable efforts to include a broad range of datasets, it is impractical to cover all available benchmarks within each task. 
Also, as many underlying datasets are publicly available, it is difficult to determine what data proprietary models were trained on, and thus whether their performance reflects genuine reasoning ability or parametric memorization. 
To address this, we intentionally include a few recent datasets (e.g., Just, Otherwise, DISRPT 2025 version, MSDC) to reduce the chance of their usage as pretraining data. 
Furthermore, we release our evaluation pipeline to support evaluation of future new datasets and more advanced LLMs, aimed at making BeDiscovER a dynamic benchmark and allowing the community to continuously evaluate LLMs over time.


For evaluation, we adopt a unified open-ended QA prompting strategy to ensure simplicity and consistency across tasks.
While model outputs can be sensitive to prompt phrasing, we intentionally employ minimal and standardized wording and retain default or suggested hyperparameter settings. We make all prompt templates and model hyperparameter publicly available (\cref{append:prompt,append:model}) to facilitate replication and future comparison.
We acknowledge that alternative methods exist, such as sentence log-probabilities \citep{hu2023prompting} and surprisal-based predictability scores \citep{giulianelli2023information, tsipidi2024surprise} have been used to probe linguistic sensitivity. 
However, these approaches are less suited for structure prediction in the benchmark.

Regarding the benchmark's ``ceiling performance'', we do not have human evaluators providing directly comparable scores, and therefore use inter-annotation agreement (IAA) as a proxy (\cref{append-tasks-iaa}). 
We acknowledge that IAA is an imperfect substitute for direct human evaluation, given differences in task formulation (e.g., text prompting vs. IAA task descriptions, original label sets vs. the unified DISRPT labels). 
Nonetheless, IAA offers a useful indicator of how consistently human annotators perform each task, providing insights into task difficulty.

Finally, we note that, in its current form, BeDiscovER is intended as a benchmark for assessing discourse knowledge (by measuring answer correctness); it is not yet positioned as a benchmark for evaluating reasoning quality. 
Our initial attempt in this regard is to track the ``reasoning trace'' or chain-of-thought tokens understand the model's process (\cref{append-plot}). 
To move toward assessing reasoning quality more directly, there may involve different strategies. For instance, future work can conduct human evaluation on part of the benchmark. 
Such evaluation could provide direct feedback on the strengths and weaknesses of model reasoning traces and, importantly, inform how high-quality reasoning traces might be leveraged in future pre-/post-training strategies. 

\section*{Acknowledgments}
The authors thank the anonymous reviewers and the Area Chair for their valuable feedback and suggestions.
The authors acknowledge the support of the Natural Sciences and Engineering Research Council of Canada (NSERC).
Nous remercions le Conseil de recherches en sciences naturelles et en génie du Canada (CRSNG) de son soutien.

\bibliography{custom}

\appendix



\section{Appendix A. Task Description}
\label{append:tasks}

\FloatBarrier
\subsection{Task Examples}
\label{append-tasks-example}

\Cref{tab:app-example}.

\onecolumn

{

\small

\begin{longtable}{l l p{8cm} p{3cm}}
\toprule
Task & Dataset & Example & Answer \\
\midrule
\endfirsthead

\toprule
Task & Dataset & Example & Answer \\
\midrule
\endhead

\midrule
\multicolumn{4}{r}{\textit{Continued on next page}} \\
\endfoot

\endlastfoot
    (1) DM & Just-Manual & (i) Betsy's picky, she \textit{just} eats chicken nuggets. & Exclusionary \\
    & & (ii) You are \textit{just} in time for a demonstration. & Temporal \\
    \\
    & Just-Subtitle & (iii) I can \textit{just} go? & Emphatic \\
    & & (iv) S-same way they \textit{just} decided. & Unexplanatory \\
    
    \\
    & Otherwise & (v) A close-up mode of this camera will give sharp results from a subject only two feet away, \textit{otherwise}, I thought the specifications somewhat limited for a camera of this price. & Exception \\
    & & (vi) A respectful adaptation must consider details such as local speech and culture, translated into a more universal dimension for a global audience. \textit{Otherwise}, the series will fall flat. & Argumentation\\
    \midrule
    
    (2) TR & TBD & (i) ADDIS ABABA, Ethiopia (AP). The Organization of African Unity said Friday it would investigate the Hutu-organized \textbf{<EVENT e4>genocide</EVENT>} of more than 500,000 minority Tutsis in Rwanda nearly four years ago. Foreign ministers of member-states \textbf{<EVENT e5>meeting</EVENT>} in the Ethiopian capital agreed to set up a seven-member panel to investigate who shot down Rwandan President Juvenal Habyarimana's plane on April 6, 1994. The assassination touched off a murderous rampage by Hutu security forces and civilians, who slaughtered mainly Tutsis but also Hutus who favored reconciliation with the minority. It also reignited the civil war. The panel also will look at the exodus of about 2 million Rwanda Hutus to neighboring countries where they lived in U.N.-run refugee camps for 2 1/2 years. The investigation will consider the role of ``internal and external forces prior to the genocide and subsequently, and the role of the United Nations and its agencies and the OAU before, during and after the genocide,'' the OAU said. The panel will be based in Addis Ababa, and will finish its investigation within a year, it said. It is to be funded by voluntary contributions from within and outside the continent. (aa-kjd) & BEFORE (e4, e5) \\
    
    \\
    & TDD-Man & (ii) BUDAPEST, Hungary (AP). Tired of being sidelined, Hungarian astronaut Bertalan Farkas is \textbf{<EVENT e2>leaving</EVENT>} for the United States to start a new career, he said Saturday. ``Being 48 is too early to be retired,'' a fit-looking Farkas said on state TV's morning talk show. With American astronaut Jon McBride, Farkas set up an American-Hungarian joint venture called Orion 1980, manufacturing space-travel related technology. Farkas will \textbf{<EVENT e11>move</EVENT>} to the company's U.S. headquarters. Farkas, an air force captain, was sent into space on board the Soyuz 36 on May 26, 1980. He spent six days aboard the Salyut 6 spacecraft with three Soviet astronauts, Valery Kubasov, Leonid Popov and Valery Riumin. McBride, 54, of Lewisburg, West Virginia, was part of a seven-member crew aboard the Orbiter Challenger in October 1984 and later served as assistant administrator for congressional relations for NASA. Farkas expressed the hope he one day follow in the footsteps of fellow astronaut John Glenn, who at 77 is about to go into space again. On May 22, 1995, Farkas was made a brigadier general, and the following year he was appointed military attache at the Hungarian embassy in Washington. However, cited by District of Columbia traffic police in December for driving under the influence of alcohol, Farkas was ordered home and retired. (ab/dc) & SIMULTANEOUS (e2, e11) \\
    \\
    & ToT-arith & (iii) The war started in 360 BC and went on for 8 years. What year did the war end in? Answer with the form of: {answer: <year> <era>}, where year is yyyy and era is one of BC or AD. Eg: 1958 BC & {'answer': '352 BC'} \\
    & & (iv) In a movie, here are some activities Theodore did. He was traveling on 24-09-1912 at 15:42:07 (24hr). He also did some snowboarding and hiking on Jan 21, 1760 and in February 1836 respectively. He was also vacationing on 13 Aug, 1927 at 22:53:42 (24hr) and skateboarding on 21-01-1760 at 18:00 (24hr). Arrange the activities Theodore did in ascending order. If a date/time is at a lower granularity, assume the earliest value for the missing information. Eg: if the date is Feb, 2020, assume the complete time to be 1 Feb 2020 00:00:00. Answer with the form of an ordered list of activities: \{'ordered\_list': [activity1, activity2, ...]\}. & \{'ordered\_list': ['snowboarding', 'skateboarding', 'hiking', 'traveling', 'vacationing']\} \\ 
    \midrule

    (3) DR & deu.rst.pcc & (i) \textbf{[ARG1]} Dagmar Ziegler sitzt in der Schuldenfalle . \textbf{[ARG2]} Auf Grund der dramatischen Kassenlage in Brandenburg hat sie jetzt eine seit mehr als einem Jahr erarbeitete Kabinettsvorlage überraschend auf Eis gelegt & causal \\
    \\
    & eng.pdtb.gum & (ii) \textbf{[ARG1]} How do people look at and experience art? \textbf{[ARG2]} Which elements of specific artworks do they focus on? & conjunction \\
    \\
    & fra.sdrt.annodis & (iii) \textbf{[ARG1]}  Milutinovic devant le TPI . \textbf{[ARG2]} L'ancien président de Serbie Milan Milutinovic , <*> s'est rendu volontairement hier au Tribunal pénal international <*> pour l'ex - Yougoslavie de La Haye . & elaboration \\
    \midrule
    
    (4) SO & ROC & (i) \textbf{<s1>} Tom let his friend borrow his phone. \textbf{<s2>} He kept draining the battery. \textbf{<s3>} The phone died shortly after. \textbf{<s4>} The friend kept using it. \textbf{<s5>} Tom got it back way later. & s1 s4 s2 s5 s3 \\
    \\
    & AAN abstract & (ii) \textbf{<s1>} We represent each medical event as a time duration, with a corresponding start and stop, and learn to rank the starts/stops based on their proximity to the admission date. \textbf{<s2>} We investigate the problem of ordering medical events in unstructured clinical narratives by learning to rank them based on their time of occurrence. \textbf{<s3>} Interestingly, we observe that this methodology performs better than a classification-based approach for this domain, but worse on the relationships found in the Timebank corpus. \textbf{<s4>} Such a representation allows us to learn all of Allen's temporal relations between medical events. \textbf{<s5>} This finding has important implications for styles of data representation and resources used for temporal relation learning: clinical narratives may have different language attributes corresponding to temporal ordering relative to Timebank, implying that the field may need to look at a wider range of domains to fully understand the nature of temporal ordering. & s2 s1 s4 s3 s5 \\
    \\
    & Wiki Movies & (iii) \textbf{<s1>} In 1940, after the fall of France, the fictitious Channel Island of Armorel is occupied by a small garrison of German troops under the benign command of Hauptmann Weiss (George Coulouris). \textbf{<s2>} In a race against the Germans discovering their presence, they spirit the cow onto a beach and via a special craft, onto a Royal Navy Motor Torpedo Boat which takes them to Britain, though they are pursued by German E-boat. \textbf{<s3>} Back in London, the Ministry of Agriculture realise that during the evacuation of the island, Venus, a prize pedigree cow, has been left behind. \textbf{<s4>} He finds that the hereditary ruler, the Suzerain, is away in the British army, leaving the Provost in charge. \textbf{<s5>} They petition the War Office to do something urgently due to the value of the cow's bloodline, and Major Morland (David Niven), is assigned the task of rescuing Venus. \textbf{<s6>} They contact the Provost and discover that the Hauptmann, a cattle breeder in civilian life, is about to have the cow shipped to Germany. \textbf{<s7>} When he realises that the Suzerain's sister, Nicola Fallaize (Glynis Johns) is in Wales, serving as an Auxiliary Territorial Service army cook, she is quickly posted to the War Office and the two, with a radio operator sergeant and a Channel Islander naval officer who knows the local waters, are landed on the island. & s1 s4 s3 s5 s7 s6 s2 \\
    \midrule

    (5) DDP & STAC & (i) \textbf{0.} Gaeilgeoir: anyone have clay? :) & \\
    & & \textbf{1.} Gaeilgeoir: I have wheat and wood & ELABORATION(0,1) \\
    & & \textbf{2.} yiin: no sorry & QA\_PAIR(0,2) \\
    & & \textbf{3.} inca: not any more, & QA\_PAIR(0,3) \\
    & & \textbf{4.} inca: sorry & COMMENT(3,4) \\
    & & \textbf{5.} nareik15: nope, & QA\_PAIR(0,5) \\
    & & \textbf{6.} nareik15: no clay & ELABORATION(5,6) \\
    & &  \textbf{7.} Gaeilgeoir: oh well &  ACK(5,7) ACK(2,7) ACK(3,7) \\
    
    & Molweni & (ii) \textbf{0.} cr1mson: apt-get i doubt my apt thing is bad though , i just installed ubuntu today \\
    && \textbf{1.} APT-GET\_INSTALL\_: now you're inside your sources.list ( the file that apt uses to find servers ) & EXPLANATION(0,1) \\
    && \textbf{2.} APT-GET\_INSTALL\_: wait ! i found a much easier way & COMMENT(0,2) \\
    && \textbf{3.} APT-GET\_INSTALL\_: well , i want you to read all of that & COMMENT(0,3) \\
    && \textbf{4.} APT-GET\_INSTALL\_: before you start mucking around in system files & BACKGROUND(0,4) \\
    && \textbf{5.} cr1mson: most of it was rem'd out & CONTINUATION(4,5) \\
    && \textbf{6.} APT-GET\_INSTALL\_: you are going to learn what all of them all from the url i just pasted & COMMENT(5,6) \\
    && \textbf{7.} cr1mson: i can always use more than one terminal & COMMENT(6,7) \\
    && \textbf{8.} cr1mson: okay , so i have to add or change a 'repository' & RESULT(6,8) \\
    \\
    & MSDC & (iii) \textbf{0} <Buil> Mission has started. \\
    && \textbf{1} <Arch> We'll be making a pair of horseshoes. & CONTINUATION(0,1) \\
    && \textbf{2} <Arch> First, start by making a 3x3 square of red bricks on the ground. & ELABORATION(1,2) \\
    && \textbf{3} <Buil> place red 2 1 -1, place red 1 1 -1, place red 0 1 -1, place red 0 1 0, place red 0 1 1, place red 1 1 0, place red 2 1 0, place red 2 1 2, place red 1 1 1, pick 2 1 2, place red 2 1 1. & RESULT(2,3)\\
    && \textbf{4} <Arch> Nice! & ACK(3,4) \\
    && \textbf{5} <Arch> Now remove the middle brick, & RESULT(4,5) NARRATION(1,5) \\
    && \textbf{6} <Arch> and one brick on the edge. & CONTINUATION(5,6) \\
    && \textbf{7} <Buil> place red 1 2 0, pick 1 2 0, pick 1 1 0, pick 2 1 1. & RESULT(6,7) \\
    && \textbf{8} <Arch> Edge, & CORRECTION(7,8) \\
    && \textbf{9} <Arch> rather than corner. & CONTRAST(8,9) \\
    && \textbf{10} <Buil> place red 2 2 0, pick 2 2 0, place red 2 1 1, pick 1 1 1. & CORRECTION(7,10) RESULT(9,10) \\
    && \textbf{11} <Arch> Awesome, & ACK(10,11) \\
\bottomrule
\caption{Representative task examples from a randomly sampled dataset. Task abbreviations: DM -- discourse markers, TR -- temporal reasoning, DR -- discourse relation, SO -- sentence ordering, DDP -- dialogue discourse parsing. Dataset name in task (3) DR follows the format ``language.framework.dataset''.}
\label{tab:app-example}
\end{longtable}
}

\twocolumn

\FloatBarrier
\subsection{Inter-Annotator Agreement}
\label{append-tasks-iaa}

\begin{table*}[htbp]
\centering
\resizebox{\textwidth}{!}{
\begin{tabular}{llcc}
\toprule
Task & Dataset & IAA & Interpretation$^\wedge$ \\
\midrule
(1) Discourse marker & Just* & -- & strong consensus \\
(1) Discourse marker & Otherwise & $K=0.87$ & near perfect \\

(2) Temporal reasoning & TBD & $K=0.56$--$0.64$ & moderate--substantial \\
(2) Temporal reasoning & TDD-Man & $K=0.69$ & substantial \\

(3) Discourse relation & ces.rst.crdt & $K=0.41$--$0.66$ & moderate--substantial \\
(3) Discourse relation & deu.pdtb.pcc & $K=0.74$ & substantial \\
(3) Discourse relation & deu.rst.pcc & $K=0.74$--$0.91$ & substantial--near perfect \\
(3) Discourse relation & eng.dep.covdtb** & -- & high consistency \\
(3) Discourse relation & eng.dep.scidtb & $K \ge 0.7$ & substantial \\
(3) Discourse relation & eng.erst.gentle & $K > 0.9$ & near perfect \\
(3) Discourse relation & eng.erst.gum & $K=0.9$ & near perfect \\
(3) Discourse relation & eng.pdtb.gentle & $K > 0.9$ & near perfect \\
(3) Discourse relation & eng.pdtb.gum & $K=0.77$--$0.83$ & substantial--near perfect \\
(3) Discourse relation & eng.pdtb.tedm & $K=0.92$ & near perfect \\
(3) Discourse relation & eng.rst.oll & -- & unknown \\
(3) Discourse relation & eng.rst.rstdt & $K=0.62$--$0.82$ & substantial--near perfect \\
(3) Discourse relation & eng.rst.umuc & $K=0.31$ & fair \\
(3) Discourse relation & eng.sdrt.stac & $K=0.58$ & moderate \\
(3) Discourse relation & eus.rst.ert & $K=0.56$ & moderate \\
(3) Discourse relation & fas.rst.prstc & -- & unknown \\
(3) Discourse relation & fra.sdrt.annodis & $K=0.57$ & moderate \\
(3) Discourse relation & ita.pdtb.luna & -- & unknown \\
(3) Discourse relation & nld.rst.nldt & $K=0.7$ & substantial \\
(3) Discourse relation & pcm.pdtb.disconaija & $K=0.60$--$0.94$ & moderate--near perfect \\
(3) Discourse relation & pol.iso.pdc & -- & unknown \\
(3) Discourse relation & por.pdtb.crpc & $K=0.71$--$0.88$ & substantial--near perfect \\
(3) Discourse relation & por.pdtb.tedm & $K=0.76$ & substantial \\
(3) Discourse relation & por.rst.cstn & $K=0.66$ & substantial \\
(3) Discourse relation & rus.rst.rrt & $K=0.8$ & substantial \\
(3) Discourse relation & spa.rst.rststb & $K=0.78$ & substantial \\
(3) Discourse relation & spa.rst.sctb & $K=0.64$--$1.0$ & substantial--near perfect \\
(3) Discourse relation & tha.pdtb.tdtb & $K=0.84$ & near perfect \\
(3) Discourse relation & tur.pdtb.tdb & $K=0.73$--$0.94$ & substantial--near perfect \\
(3) Discourse relation & tur.pdtb.tedm & $K=0.71$ & substantial \\
(3) Discourse relation & zho.dep.scidtb & $K=0.72$ & substantial \\
(3) Discourse relation & zho.pdtb.cdtb & $K=0.94$ & near perfect \\
(3) Discourse relation & zho.pdtb.ted & $K=0.81$--$0.94$ & near perfect \\
(3) Discourse relation & zho.rst.gcdt & $K=0.57$ & moderate \\
(3) Discourse relation & zho.rst.sctb & $K=0.73$--$0.84$ & substantial--near perfect \\

(5) Dialogue discourse parsing & STAC & link $K=0.72$, relation $K=0.58$ & substantial, moderate \\
(5) Dialogue discourse parsing & Molweni & link $K=0.91$, relation $K=0.56$ & near perfect, moderate \\
(5) Dialogue discourse parsing & MSDC & -- & unknown \\
\bottomrule
\end{tabular}}
\caption{Inter-annotation agreement (IAA) statistics across tasks and datasets.
Interpretation$^\wedge$ follows the Standard interpretation of inner-annotator agreement (IAA) score \citep{landis1977measurement}.
Just*: All sentences in both Just subsets have a strong primary reading, either by construction (in the hand-constructed corpus, i.e., Just-manual) or by annotator agreement (in the annotated corpus, i.e., Just-subtitle).
eng.dep.covdtb**: 
The authors had several discussions with each annotator to maintain the annotation consistency at a satisfactory level.
}
\label{tab:app-iaa}
\end{table*}

\Cref{tab:app-iaa}.
For Task (3) discourse relation classification, we adopt the labeling scheme used in the DISRPT shared task, which follows the format ``language.framework.dataset''.
All scores are obtained from corresponding original datasets.
We observe the following:

\begin{itemize}[itemsep=1pt]
    \item Task (1) discourse marker disambiguation, both Just and Otherwise exhibit strong annotation consensus, with Otherwise achieving particularly high agreement (K=0.87). Despite this, current LLMs still fall short of human-level performance.
    \item Task (2) temporal reasoning, event-based datasets (TBD and TDD-man) also share substantial annotation agreement. Comparable results have so far only been achieved by task-specific supervised models. We find that both reasoning-oriented and general-purpose LLMs lag significantly behind.
    \item Task (3) discourse relation prediction: across languages and frameworks, IAA scores are generally substantial to near-perfect, with only a few challenging datasets, e.g. STAC (K=0.58) and ERT (K=0.56), showing moderate agreement. In contrast, LLM performance on BeDiscovER remains uniformly low, indicating considerable room for improvement.
    \item Task (4) sentence ordering, this task does not require human annotation, so we do not have a direct IAA comparison. However, as highlighted by a recent survey \citep{shi2024overview}, supervised language models still underperform relative to humans, especially on multi-paragraph ordering. 
    \item Task (5) dialogue discourse parsing, link and full structure annotation yields substantial and moderate human agreement, respectively. As noted by the STAC creators, ``Bearing in mind that annotating full discourse structures is a very complex task, this is a relatively good score \citep{asher2016discourse}. Although supervised parsers can approach this level, general-purpose LLMs still fall noticeably behind.
\end{itemize}

\FloatBarrier
\section{Appendix B. Model Hyperparameters}
\label{append:model}

\Cref{tab:app-model}.

\begin{table*}[htbp]
    \centering
    \resizebox{0.9\textwidth}{!}{
    \begin{tabular}{llccccccccc}
        \toprule
        & & \multicolumn{4}{c}{{w/o reasoning}} & \multicolumn{5}{c}{w/ reasoning} \\
        \cmidrule(lr){3-6} \cmidrule(lr){7-11}
        Model & Size & Max\_new\_tokens & Temp & Top-p & Seeds & Max\_new\_tokens & Temp & Top-p & Seeds & Verbosity \\
        
        \midrule
        Qwen3-0.6B & 0.6B & 2048 & 0.7 & 0.8 & 0,1,42 & 4096 & 0.6 & 0.95 & 0,1,42 & - \\
        Qwen3-1.7B & 1.7B  & 2048 & 0.7 & 0.8 & 0,1,42 & 4096 & 0.6 & 0.95 & 0,1,42 & - \\
        Qwen3-4B & 4B & 2048 & 0.7 & 0.8 & 0,1,42 & 4096 & 0.6 & 0.95 & 0,1,42 & - \\
        Qwen3-8B & 8B & 2048 & 0.7 & 0.8 & 0,1,42 & 4096 & 0.6 & 0.95 & 0,1,42 & - \\
        Qwen3-14B & 14B & 2048 & 0.7 & 0.8 & 0,1,42 & 4096 & 0.6 & 0.95 & 0,1,42 & - \\
        Qwen3-32B & 32B & 2048 & 0.7 & 0.8 & 0,1,42 & 4096 & 0.6 & 0.95 & 0,1,42 & - \\
        
        \cdashline{1-11}
        DS-r1-distill-Qwen & 32B  & - & - & - & - & 4096 & - & - & 0,1,42 & - \\
        DeepSeek-r1-0528 & 37/671B & - & - & - & - & 4096-16384 & - & - & 0,1,42 & - \\
        GPT-5-mini (low) & Unk. & - & - & - & - & 4096 & - & - & 0,1,42 & low \\
        GPT-5-mini (high) & Unk. & - & - & - & - & 8192-32768 & - & - & 0,1,42 & low \\
        
        \cdashline{1-11}
        GPT-4o-mini & Unknown  & 4096 & - & - & 0,1,42 & - & - & - & - & -\\
        Llama-4-Scout & 17/109B  & 4096 & - & - & 0,1,42 & - & - & - & - & -\\
        Qwen2.5-72B & 72B  & 4096 & - & - & 0,1,42 & - & - & - & - & - \\
        \bottomrule
    \end{tabular}}
    \caption{A summary of hyperparameters used for LLMs.
    For Qwen3 models, we use the suggested temperature and top-p values.
   For the dialogue discourse parsing task, we set a higher max\_new\_tokens value for DeepSeek-r1-0528 and GPT-5-mini with higher reasoning effort.
    }
    \label{tab:app-model}
\end{table*}

\FloatBarrier

\section{Appendix C. Error Analysis}
\label{append:error}

\FloatBarrier
\subsection{Task (1) Discourse Marker Understanding}
\label{append-error-dm}

\begin{figure*}[htbp]
    \centering
    \begin{subfigure}[b]{0.49\textwidth}
        \centering
        \includegraphics[width=\textwidth]{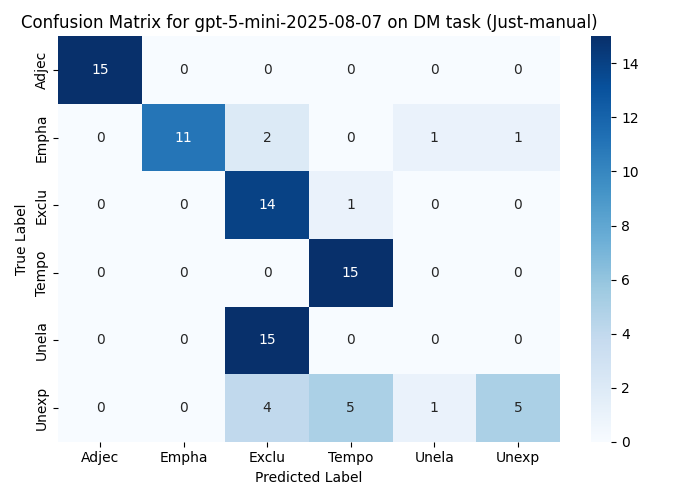}
    \end{subfigure}
    \begin{subfigure}[b]{0.49\textwidth}
        \centering
        \includegraphics[width=\textwidth]{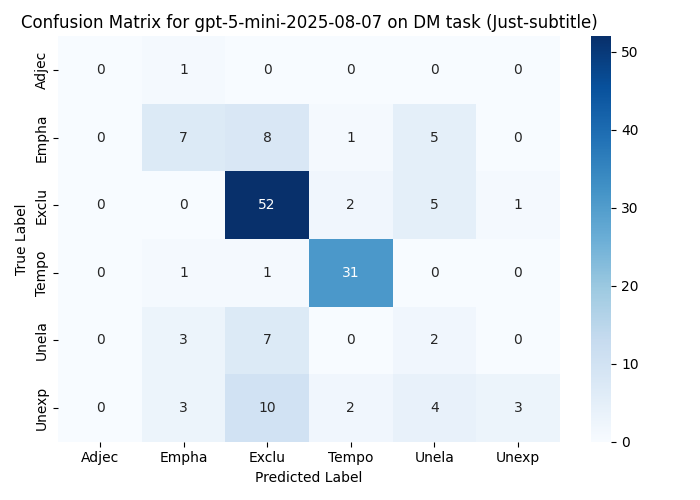}
    \end{subfigure}
    \begin{subfigure}[b]{0.49\textwidth}
        \centering
        \includegraphics[width=\textwidth]{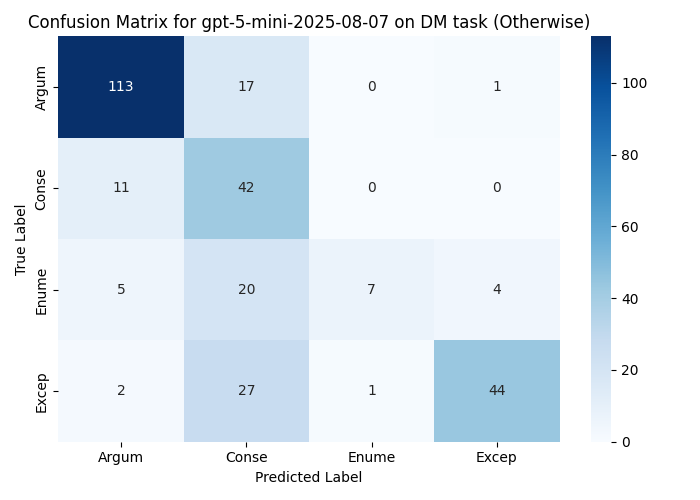}
    \end{subfigure}
    \caption{Confusion Matrices on \textbf{Task (1) Discourse Marker Understanding} on \textbf{\textit{Just-Manual}, \textit{Just-Subtitle}, \textit{Otherwise}} datasets, in the \textsc{Def+Exp} setup.
    Just relation labels are: Adjective, Emphatic, Exclusionary, Temporal, Unelaboratory, Unexplanatory.
    Otherwise relation labels are: Argumentation, Consequence, Exception, Enumeration.
    }
    \label{fig:app-error-dm}
\end{figure*}

\Cref{fig:app-error-dm}.
Confusion matrices using GPT-5-mini (high) for \textit{Just} and \textit{Otherwise} datasets, all in the \textsc{Def+Exp} setting.

We observe a notable contrast in the results of \textit{Just} subsets. On the cleaner, expert-curated \textit{Just-Manual} subset, GPT-5-mini can largely distinguish the senses ``Adjective'', ``Emphatic'', ``Exclusionary'', and ``Temporal'', but fails almost entirely on ``Unelaboratory'' and ``Unexplanatory'' -- two closely related readings that both deny further elaboration or explanation. 
On the more challenging \textit{Just-Subtitle} subset, the model primarily identifies only ``Exclusionary'' and ``Temporal'', with little success on the remaining senses.

For the \textit{Otherwise} dataset, the model distinguishes ``Argumentation'' and ``Exception'' reasonably well, but tends to over-predict ``Consequence'' and struggles to recognize ``Enumeration'' which provides another option to achieve some goal -- a less-used function of Otherwise.

\FloatBarrier
\subsection{Task (2) Temporal Reasoning}
\label{append-error-tr}

\begin{figure*}[htbp]
    \centering
    \begin{subfigure}[b]{0.49\textwidth}
        \centering
        \includegraphics[width=\textwidth]{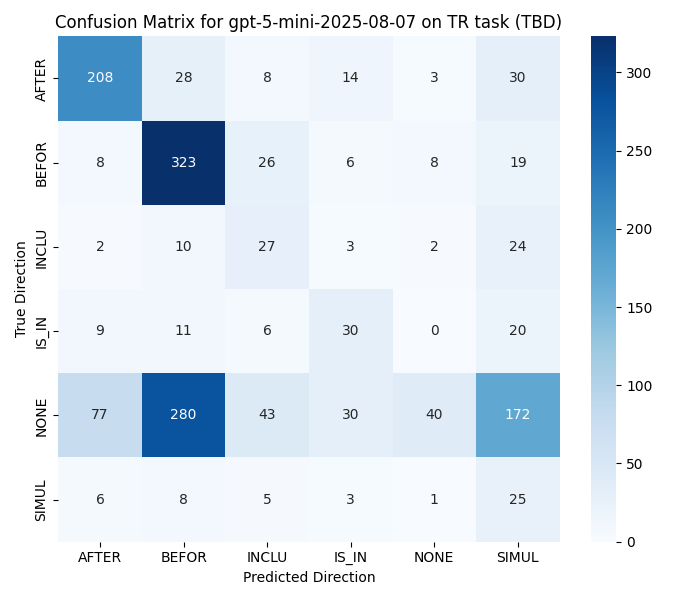}
    \end{subfigure}
    \begin{subfigure}[b]{0.49\textwidth}
        \centering
        \includegraphics[width=\textwidth]{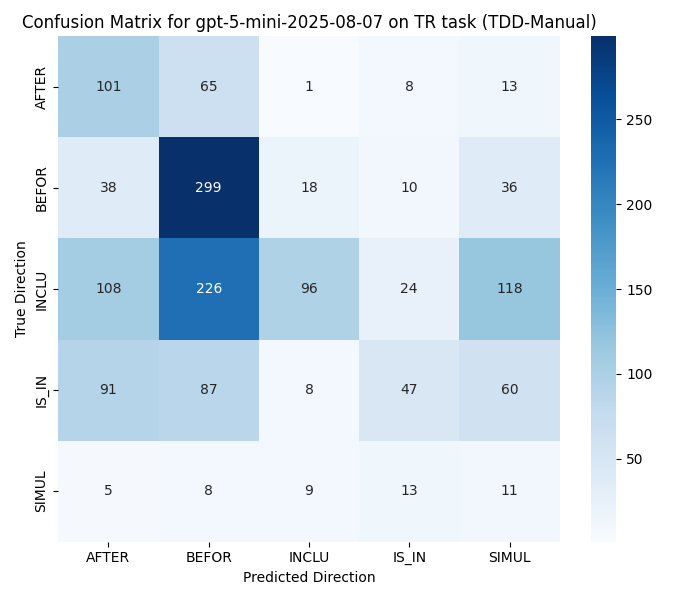}
    \end{subfigure}
    \caption{Confusion matrices on \textbf{Task (2) Temporal Reasoning}, \textbf{TBD} and \textbf{TDD-Man} datasets.
    Temporal labels are: AFTER, BEFORE, INCLUDES, IS\_INCLUDED, NONE, SIMULTANEOUS.
    }
    \label{fig:app-error-tr}
\end{figure*}

\Cref{fig:app-error-tr}.
GPT-5-mini can generally identify when one event occurs before or after another. However, it performs poorly when events overlap or include one another: either predicting such cases only rarely (as in TBD) or over-predicting them (as in TDD-Manual). 
The model also struggles to correctly identify simultaneous events. These results are extremely interesting, suggesting that while the model may possess some commonsense knowledge on straightforward temporal order (before/after), it has difficulty making finer-grained distinctions among more nuanced temporal relations.

\FloatBarrier
\subsection{Task (3) Discourse Relation Recognition}
\label{append-error-dr}

\begin{figure*}[htbp]
    \centering
    \includegraphics[width=\linewidth]{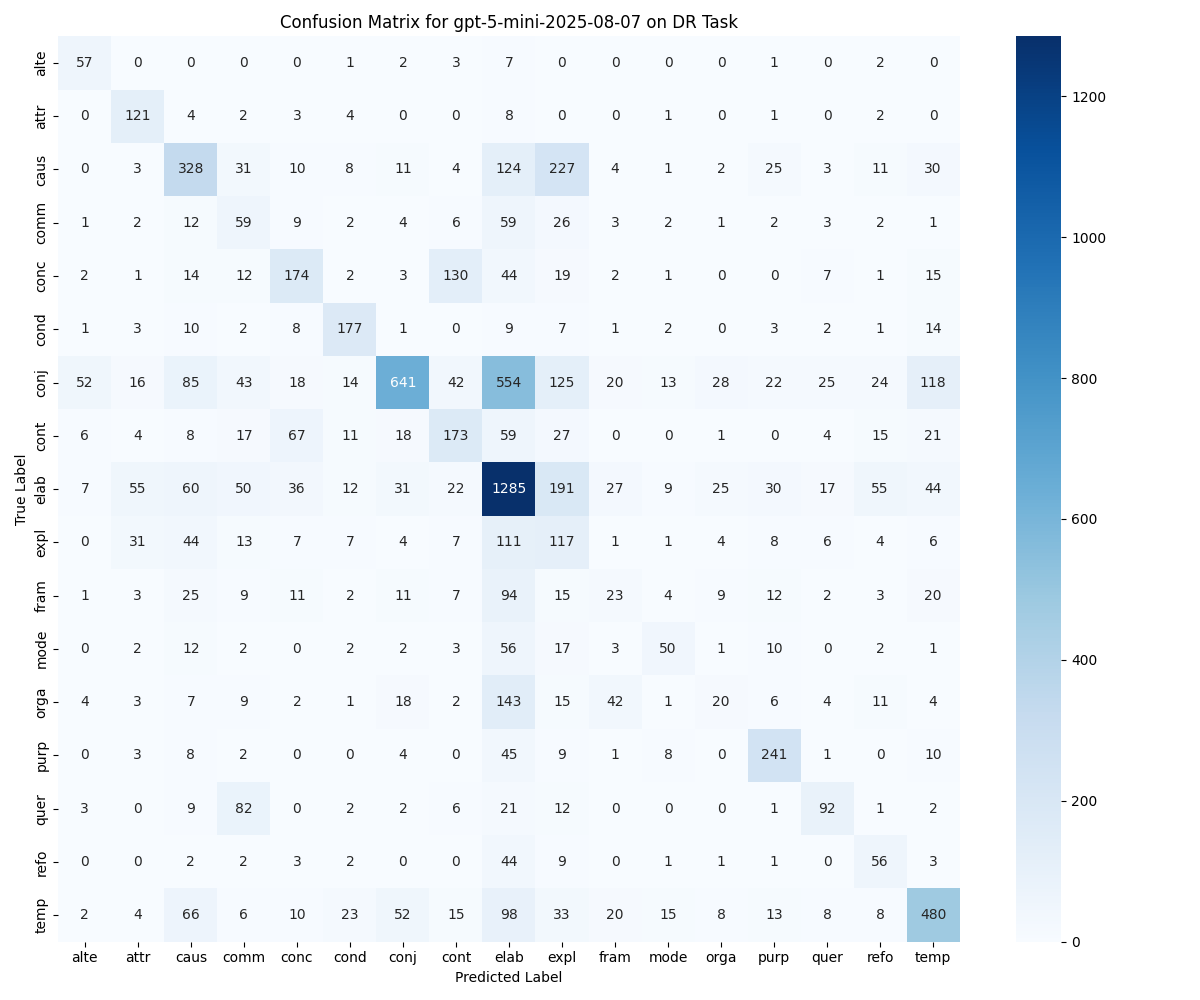}
    \caption{Confusion matrix of \textbf{Task (3) Discourse Relation Classification}, with \textbf{DISRPT 2025 Shared Task}.
    Relation labels are: alternation, attribution, causal, comment, concession, condition, conjunction, contrast, elaboration, explanation, frame, mode, organization, purpose, query, reformulation, temporal.
    }
    \label{fig:app-error-dr}
\end{figure*}

\Cref{fig:app-error-dr}.
This confusion matrix shows strong performance on several major discourse relations including ``elaboration'', ``conjunction'', and ``temporal''. At the same time, it also reveals notable confusion between ``elaboration'' and ``conjunction'': a pattern consistent with the findings reported in DeDisCo \citep{ju2025dedisco}. 
This may stem from the semantic and structural similarity between these two relations. 
Also, splitting apart ``cause'' and ``explanation'' is challenging as their differences can be subtle. Finally, the model exhibits a tendency to over-predict ``elaboration'', the most frequent label in the dataset.

\FloatBarrier
\subsection{Task (4) Sentence Ordering}
\label{append-error-so}

\begin{figure*}[htbp]
    \centering
    \includegraphics[width=.8\linewidth]{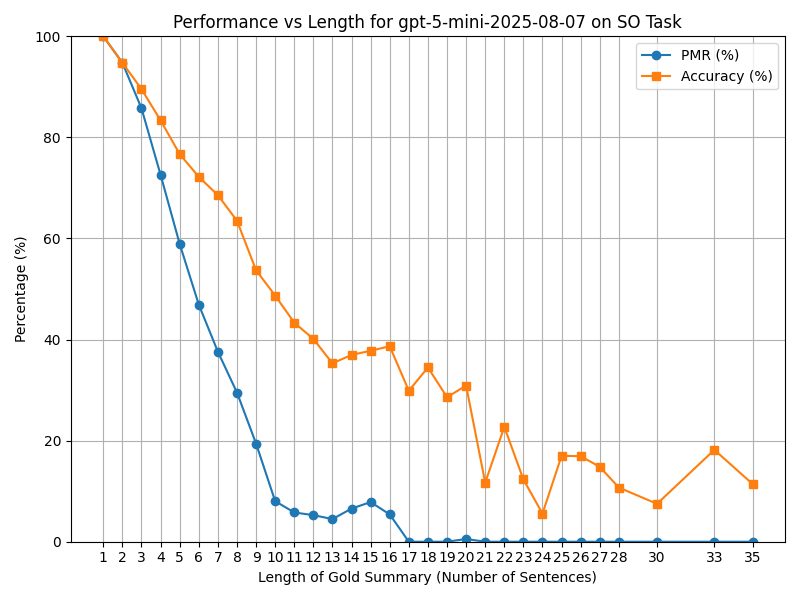}
    \caption{Performance vs. document length on \textbf{Task (4) Sentence Ordering}, all seven datasets' results aggregated.
    PMR: perfect match.
    }
    \label{fig:app-error-so}
\end{figure*}

\Cref{fig:app-error-so}.
We report results split by document-length bins and observe a consistent decline in performance as input length increases across datasets. This trend is expected, as longer documents pose greater reordering difficulty. Once a shuffled document exceeds 17 sentences, the model is unable to produce a perfect reconstruction and achieves only partial matches (around 15\% accuracy). In comparison, supervised models such as reBART \citep{chowdhury2021everything} maintain roughly 30–40\% accuracy on documents of similar length.

\FloatBarrier
\subsection{Task (5) Dialogue Discourse Parsing}
\label{append-error-ddp}

\begin{figure*}[htbp]
    \centering
    \includegraphics[width=.8\linewidth]{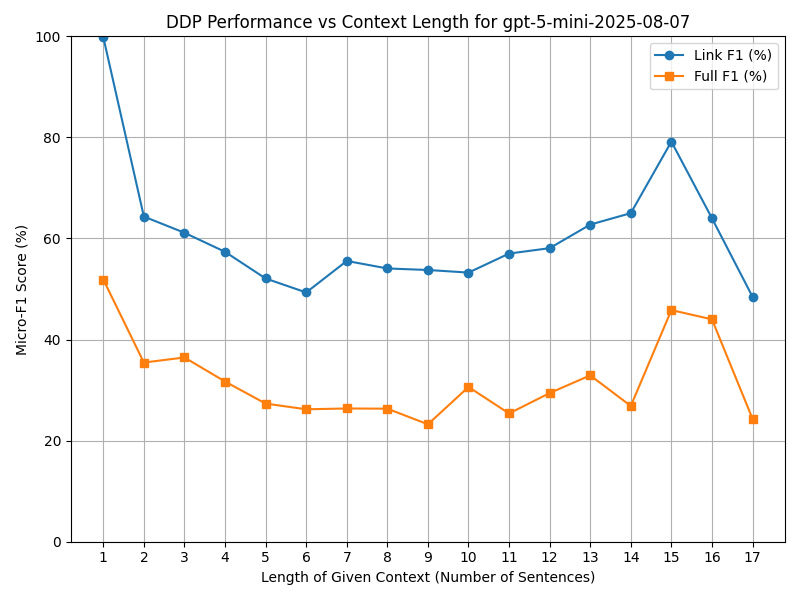}
    \caption{Performance vs. context length on \textbf{Task (5) Dialogue Discourse Parsing}, all three datasets' results aggregated.
    }
    \label{fig:app-error-ddp}
\end{figure*}

\Cref{fig:app-error-ddp}.
We adopt an incremental generation approach and analyze how performance varies with the amount of preceding context (including both the text and the predicted discourse structure). With a small context window of 1–4 sentences, the model achieves reasonably good performance on both link prediction (70) and full-structure prediction (38). As more context is added, we do not observe a consistent decline. From 5–16 sentences, GPT-5-mini remains relatively stable, suggesting that the model largely focuses on nearby sentences rather than the entire context. However, once the context exceeds 17 sentences, performance drops sharply.

\FloatBarrier
\section{Appendix D. Prompt Templates}
\label{append:prompt}


\FloatBarrier
\subsection{Task (1) Discourse Marker Understanding}
\label{append-prompt-dm}

\Cref{tab:app-prompt-dm}.

\begin{table*}[htbp]
    \centering
    \resizebox{\textwidth}{!}{
    \begin{tabularx}{\linewidth}{X}
        \toprule
        \textit{System prompt}: \\
        A conversation between User and Assistant. The User provides one sentence that contains a discourse marker `just'. The assistant identifies the **discourse function** of the marker. Choose one of the following six labels: [Exclusionary, Unelaboratory, Unexplanatory, Emphatic, Temporal, Adjective]. Choose ONLY ONE label. Keep the answer short.\\
        \noindent\hfill (for \textbf{\textit{Just}} dataset) \\
        \midrule
        \textit{System prompt}: \\
        A conversation between User and Assistant. The User provides two arguments linked with a discourse marker `otherwise': [Arg1]. Otherwise, [Arg2]. The assistant identifies the **discourse function** of the marker. Choose one of the following four labels: [Argumentation, Consequence, Exception, Enumeration]. Choose ONLY ONE label. Keep the answer short.\\
        \noindent\hfill (for \textbf{\textit{Otherwise}} dataset) \\
        \midrule
        \textit{User prompt}: \\
        <TEXT>\\
        \\
        Question: What is the function of the discourse marker `just' in the sentence above?\\
        \noindent\hfill (for \textbf{\textit{Just}} dataset) \\
        \midrule
        \textit{User prompt}: \\
        Here are the main functions of `just' along with their definition:\\
        \\
        Exclusionary: `Just' is used to exclude other possibilities or options.\\
        \\
        Unelaboratory: `Just' is used to deny further elaboration on an event or concept.\\
        \\
        Unexplanatory: `Just' is used to deny that there is an explanation or to offer a weak explanation with no stronger one available.\\
        \\
        Emphatic: `Just' is used to add emphasis to an already strong word or phrase.\\
        \\
        Temporal: `Just' is used to indicate that something happened very recently, or close to another event.\\
        \\
        Adjective: `Just' is used used as an adjective to describe a person or idea, especially a law or policy, as fair, appropriate, or lawful.\\
        \\
        Given these, identify the function of `just' in the following sentence. Respond with the function label.\\
        \\
        <TEXT>\\
        \noindent\hfill (for \textbf{\textit{Just}} dataset) \\
        \midrule
        \textit{User prompt}: \\
        Here are the main functions of `just' along with their definition and examples:\\
        \\
        Exclusionary: `Just' is used to exclude other possibilities or options.\\
        Example: ``I'm just looking for a job, not a career.''\\
        \\
        Unelaboratory: `Just' is used to deny further elaboration on an event or concept.\\
        Example: ``I just don't like it.''\\
        \\
        Unexplanatory: `Just' is used to deny that there is an explanation or to offer a weak explanation with no stronger one available.\\
        Example: ``The lights just turn on and off.''\\
        \bottomrule
    \end{tabularx}}
\end{table*}

\begin{table*}[htbp]
    \centering
    \resizebox{\textwidth}{!}{
    \begin{tabularx}{\linewidth}{X}
        \toprule
        Emphatic: `Just' is used to add emphasis to an already strong word or phrase.\\
        Example: ``It's just amazing!''\\
        \\
        Temporal: `Just' is used to indicate that something happened very recently, or close to another event.\\
        Example: ``I just saw him a minute ago.''\\
        \\
        Adjective: `Just' is used used as an adjective to describe a person or idea, especially a law or policy, as fair, appropriate, or lawful.\\
        Example: ``She is a just ruler.''\\
        \\
        Given these, identify the function of `just' in the following sentence. Respond with the function label.\\
        \\
        <TEXT>\\ 
        \noindent\hfill (for \textbf{\textit{Just}} dataset) \\
        \midrule
        \textit{User prompt}: \\
        <TEXT>\\
        \\
        Question: What is the function of the discourse marker `Otherwise' between [Arg1] and [Arg2]?\\
        \noindent\hfill (for \textbf{\textit{Otherwise}} dataset) \\
        \midrule
        Here are the main functions of `Otherwise' along with their definition:\\
        \\
        Consequence: If the situation in Arg1 doesn't occur, the situation in Arg2 would arise.\\
        \\
        Argumentation: Arg2 is undesirable and can be possibly avoided by following Arg1.\\
        \\
        Enumeration: It doesn't take the failure of Arg1 to consider Arg2 as another option.\\
        \\
        Exception: Arg1 is an exception to Arg2.\\
        \\
        Given these, identify the function of `Otherwise' between [Arg1] and [Arg2] in the following sentence. Respond with the function label.\\
        \\
        <TEXT>\\
        \noindent\hfill (for \textbf{\textit{Otherwise}} dataset) \\
        \midrule
        \textit{User prompt}: \\
        Here are the main functions of `Otherwise' along with their definition and examples. Arg1 and Arg2 are placed within square brackets:\\
        \\
        Consequence: If the situation in Arg1 doesn't occur, the situation in Arg2 would arise.\\
        Example: [I like you too.] Otherwise, [we wouldn't be friends.]\\
        \\
        Argumentation: Arg2 is undesirable and can be possibly avoided by following Arg1.\\
        Example: [We have to operate immediately.] Otherwise, [she will die.]\\
        \\
        Enumeration: It doesn't take the failure of Arg1 to consider Arg2 as another option.\\
        Example: [You can choose to go to the beach.] Otherwise, [you can stay at home and read a book.]\\
        \\
        Exception: Arg1 is an exception to Arg2.\\
        Example: [He is usually very punctual.] Otherwise, [he would have arrived on time.]\\
        \bottomrule
    \end{tabularx}}
\end{table*}

\begin{table*}[htbp]
    \centering
    \resizebox{\textwidth}{!}{
    \begin{tabularx}{\linewidth}{X}
        \toprule
        Given these, identify the function of `Otherwise' between [Arg1] and [Arg2] in the following sentence. Respond with the function label.\\
        \\
        <TEXT>\\
        \noindent\hfill (for \textbf{\textit{Otherwise}} dataset) \\
        \bottomrule
    \end{tabularx}}
    \caption{Prompt Templates used for \textbf{Task (1) Discourse Marker Understanding}.
    }
    \label{tab:app-prompt-dm}
\end{table*}

\FloatBarrier
\subsection{Task (2) Temporal Reasoning}
\label{append-prompt-tr}

Table~\ref{tab:app-prompt-tr}.

\begin{table*}[htbp]
    \centering
    \resizebox{\textwidth}{!}{
    \begin{tabularx}{\linewidth}{X}
        \toprule
        \textit{System prompt}: \\ 
        A conversation between User and Assistant. The User provides a document with two target EVENT, each marked with <EVENT e>...</EVENT>. The assistant identifies the **temporal relation** between the two events.\\
        Choose one of the following six labels: [IS\_INCLUDED, INCLUDES, SIMULTANEOUS, BEFORE, AFTER, NONE]. Answer NONE if unsure. Keep the answer short and concise.\\
        \noindent\hfill (for \textbf{TBD} and \textbf{TDD-Man} datasets) \\
        \midrule
        \textit{System prompt}: \\ 
        A conversation between User and Assistant. The User asks a question that requires temporal arithmetic reasoning. The Assistant must reason using the logic and semantics of time to answer. Strictly follow the instructions when answering. Keep the answer short and concise.\\
        \noindent\hfill (for \textbf{ToT-Arithmetic} dataset) \\
        \midrule
        \textit{User prompt}: \\
        Given the document: \\
        \\
        <DOCUMENT> \\
        \\
        Quesiton: What is the temporal relation of <EVENT> <EVENT ID> with respect to <EVENT> <EVENT ID>? Choose one of the following labels: [IS\_INCLUDED, INCLUDES, SIMULTANEOUS, BEFORE, AFTER, NONE].\\
        \noindent\hfill (for \textbf{TBD} and \textbf{TDD-Man} datasets) \\
        \midrule
        \textit{User prompt}: \\
        <QUESTION> \\
        \noindent\hfill (for \textbf{ToT-Arithmetic} dataset) \\
        \bottomrule
    \end{tabularx}}
    \caption{Prompt Templates used for \textbf{Task (2) Temporal Reasoning}.
    }
    \label{tab:app-prompt-tr}
\end{table*}

\FloatBarrier
\subsection{Task (3) Discourse Relation Recognition}
\label{append-prompt-dr}

Table~\ref{tab:app-prompt-dr}.

\begin{table*}[htbp]
    \centering
    \resizebox{\textwidth}{!}{
    \begin{tabularx}{\linewidth}{X}
        \toprule
        \textit{System prompt}: \\
        A conversation between User and Assistant. You are a helpful assistant.\\
        \#\# Role and Goal:\\
        You are an expert in discourse analysis, tasked with identifying the discourse relation between two sentence units based on the provided label. Your goal is to accurately determine the relationship between these two units.\\
        \\
        \#\# Guidelines:\\
        1. You will receive Unit1 and Unit2. Unit1 appears before Unit2 in the original text.\\
        2. You will be informed about the language of these units.\\
        3. You will be informed of the corpus from which the data is drawn, which may help guide your analysis.\\
        4. The framework for analysis will be provided, outlining the structure used for discourse analysis.\\
        5. You will be given the context in which these two units appear.\\
        6. The direction of the relationship between these two units will be given.\\
        7. You will be provided with a set of labels representing possible discourse relations. Do not modify any label. Choose one label that best fits the relationship between Unit1 and Unit2, and output only the chosen label.\\
        \\
        \#\# Labels:\\
        contrast, condition, mode, organization, frame, temporal, concession, reformulation, comment, query, attribution, alternation, purpose, explanation, elaboration, causal, conjunction.\\
        \midrule
        \textit{User prompt}: \\
        \#\# Language:\\
        <LANG>\\
        \\
        \#\# Corpus:\\
        <CORPUS>\\
        \\
        \#\# Framework:\\
        <FRAMEWORK>\\
        \\
        \#\# Context:\\
        <CONTEXT>\\
        \\
        \#\# Direction:\\
        <DIRECTION>\\
        \\
        \#\# Unit1:\\
        <UNIT1>\\
        \\
        \#\# Unit2:\\
        <UNIT2>\\
        \\
        Question: What is the discourse relation between Unit1 and Unit2?"\\
        \bottomrule
    \end{tabularx}}
    \caption{Prompt Templates used for \textbf{Task (3) Discourse Relation Recognition}.
    }
    \label{tab:app-prompt-dr}
\end{table*}

\FloatBarrier
\subsection{Task (4) Sentence Ordering}
\label{append-prompt-so}

Table~\ref{tab:app-prompt-so}.

\begin{table*}[htbp]
    \centering
    \resizebox{\textwidth}{!}{
    \begin{tabularx}{\linewidth}{X}
        \toprule
        \textit{System prompt}: \\
        A conversation between User and Assistant. The User provides a list of shuffled sentences. Each sentence is enclosed within sentence tags, e.g., <s 1> Sentence one. </s 1>. The Assistant reorders the sentences to form a coherent paragraph. Return only the sequence of sentence tags in the correct order, for example: <s 2> <s 1> <s 3>. Do not include any sentence content. Do not repeat or modify any sentence tags.\\
        \midrule
        \textit{User prompt}: \\
        <s 1> <SENTENCE 1> </s {1}>\\
        <s 1> <SENTENCE 2> </s {1}>\\
        ...\\
        <s n> <SENTENCE n> </s {n}>\\
        \bottomrule
    \end{tabularx}}
    \caption{Prompt Templates used for \textbf{Task (4) Sentence Ordering}.
    }
    \label{tab:app-prompt-so}
\end{table*}

\FloatBarrier
\subsection{Task (5) Dialogue Discourse Parsing}
\label{append-prompt-ddp}

Table~\ref{tab:app-prompt-ddp}.

\begin{table*}[htbp]
    \centering
    \resizebox{\textwidth}{!}{
    \begin{tabularx}{\linewidth}{X}
        \toprule
        \textit{System prompt}: \\
        A conversation between User and Assistant. The User provides a boardgame dialogue history / Ubuntu chat log (context and structure), along with a new turn. The assistant identifies the **discourse structure** of the new turn in relation to the context. The structure should be represented in the format: RELATION(S\_ID1, S\_ID2), where RELATION is the discourse relation, S\_ID2 is the ID of the new turn, and S\_ID1 is the ID of the sentence in the context that it relates to. S\_ID2 is bigger than S\_ID1 since dialogue flow is unidirectional and there is no backward relation. Use the following discourse relations: [ACKNOWLEDGEMENT, ALTERNATION, BACKGROUND, CLARIF\_Q, COMMENT, CONDITIONAL, CONTINUATION, CONTRAST, CORRECTION, ELABORATION, EXPLANATION, NARRATION, PARALLEL, QA\_PAIR, Q\_ELABORATION, RESULT]. If multiple relations exist, separate them with spaces. Provide only the structure without any additional text.\\
        \noindent\hfill (for \textbf{STAC} and \textbf{Molweni} datasets) \\
        \midrule
        \textit{System prompt}: \\
        A conversation between User and Assistant. The User provides a Minecraft game history (context and structure), along with a new turn. The assistant identifies the **discourse structure** of the new turn in relation to the context. The structure should be represented in the format: RELATION(S\_ID1, S\_ID2), where RELATION is the discourse relation, S\_ID2 is the ID of the new turn, and S\_ID1 is the ID of the sentence in the context that it relates to. S\_ID2 is bigger than S\_ID1 since dialogue flow is unidirectional and there is no backward relation. Use the following discourse relations: [ACKNOWLEDGEMENT, ALTERNATION, CLARIF\_Q, COMMENT, CONDITIONAL, CONFIRM\_Q, CONTINUATION, CONTRAST, CORRECTION, ELABORATION, EXPLANATION, NARRATION, QA\_PAIR, Q\_ELABORATION, RESULT, SEQUENCE]. If multiple relations exist, separate them with spaces. Provide only the structure without any additional text.\\
        \noindent\hfill (for \textbf{MSDC} dataset) \\
        \midrule
        \textit{User prompt}:\\
        \#\# Context:\\
        <CONTEXT>\\
        \\
        \#\# Structure:\\
        <STRUCTURE>\\
        \\
        \#\# New turn:\\
        <NEW TURN>\\
        \\
        Question: What is the discourse structure of the new turn in relation to the context? Provide the answer in the format: RELATION(S\_ID1, S\_ID2).\\
        \bottomrule
    \end{tabularx}}
    \caption{Prompt Templates used for \textbf{Task (5) Dialogue Discourse Parsing}.
    }
    \label{tab:app-prompt-ddp}
\end{table*}

\FloatBarrier



\section{Appendix E. Detailed Results}
\label{append-results-tables}

\subsection{Task (1) Discourse Marker Understanding}
\label{append-dm}

\Cref{tab:app-1-1-1,tab:app-1-1-2,tab:app-1-1-3,tab:app-1-1-4,tab:app-1-1-5,tab:app-1-1-6,tab:app-1-2-1,tab:app-1-2-2,tab:app-1-2-3}.

\begin{table*}[htbp]
    \centering
    \resizebox{\textwidth}{!}{
\begin{tabular}{llcccccccccc}
        \toprule
        & & \multicolumn{5}{c}{{w/o reasoning}} & \multicolumn{5}{c}{w/ reasoning} \\
        \cmidrule(lr){3-7} \cmidrule(lr){8-12}
        Model & Size & Acc & weighted-Pre & weighted-Rec & weighted-F1 & tok. & Acc & weighted-Pre & weighted-Rec & weighted-F1 & tok. \\
        \midrule
        Qwen3-0.6B & 0.6B & $17.3_{1.4}$ & $4.0_{2.3}$ & $17.3_{1.4}$ & $5.7_{1.8}$ & 4 & $32.0_{1.6}$ & $18.4_{0.5}$ & $32.0_{1.6}$ & $22.9_{0.9}$ & 279 \\
        Qwen3-1.7B & 1.7B & $14.0_{0.4}$ & $2.2_{0.2}$ & $10.5_{0.3}$ & $3.7_{0.2}$ & 3 & $38.9_{0.7}$ & $41.9_{3.4}$ & $38.9_{0.7}$ & $31.9_{1.3}$ & 538 \\
        Qwen3-4B & 4B & $34.7_{0.4}$ & $30.3_{1.7}$ & $34.7_{0.4}$ & $27.0_{0.4}$ & 3 & $55.1_{1.8}$ & $62.5_{3.2}$ & $55.1_{1.8}$ & $50.7_{1.8}$ & 543 \\
        Qwen3-8B & 8B & $44.7_{0.6}$ & $44.0_{3.6}$ & $44.7_{0.6}$ & $34.5_{0.7}$ & 3 & $59.1_{2.9}$ & $67.5_{7.5}$ & $59.1_{2.9}$ & $54.0_{3.8}$ & 583 \\
        Qwen3-14B & 14B & $57.8_{2.0}$ & $55.6_{3.4}$ & $57.8_{2.0}$ & $53.1_{2.0}$ & 2 & $60.2_{0.7}$ & $62.3_{5.7}$ & $60.2_{0.7}$ & $52.6_{1.4}$ & 404 \\
        Qwen3-32B & 32B & $55.1_{2.0}$ & $60.6_{2.6}$ & $55.1_{2.0}$ & $51.4_{2.3}$ & 3 & $59.8_{1.8}$ & $60.0_{7.2}$ & $59.8_{1.8}$ & $53.5_{2.3}$ & 425 \\
        \midrule
        DS-r1-distill-Qwen & 32B &  - & - & - & - & - & $56.7_{0.9}$ & $60.4_{9.9}$ & $51.3_{4.6}$ & $50.3_{5.5}$ & 304 \\
        DeepSeek-r1-0528 & 37/671B &  - & - & - & - & - & $53.7_{0.5}$ & $20.9_{1.8}$ & $19.5_{2.0}$ & $18.5_{1.5}$ & 466 \\
        GPT-5-mini (low) & Unk. &  - & - & - & - & - & $63.3_{2.4}$ & $52.5_{5.0}$ & $60.4_{6.0}$ & $53.3_{5.5}$ & 94 \\
        GPT-5-mini (high) & Unk. &  - & - & - & - & - & $61.7_{0.6}$ & $56.6_{8.7}$ & $61.7_{0.6}$ & $53.5_{1.3}$ & 857 \\
        \bottomrule
    \end{tabular}}
    \caption{Details on \textbf{Task (1) Discourse Marker Understanding} performance on \textbf{\textit{Just-Manual}} dataset, using \textbf{basic} prompting strategy.
    }
    \label{tab:app-1-1-1}
\end{table*}

\begin{table*}[htbp]
    \centering
    \resizebox{\textwidth}{!}{
    \begin{tabular}{llcccccccccc}
        \toprule
        & & \multicolumn{5}{c}{{w/o reasoning}} & \multicolumn{5}{c}{w/ reasoning} \\
        \cmidrule(lr){3-7} \cmidrule(lr){8-12}
        Model & Size & Acc & weighted-Pre & weighted-Rec & weighted-F1 & tok. & Acc & weighted-Pre & weighted-Rec & weighted-F1 & tok. \\
        \midrule
        Qwen3-0.6B & 0.6B & $16.9_{1.1}$ & $5.2_{2.1}$ & $16.9_{1.1}$ & $7.7_{2.3}$ & 3 & $28.4_{1.6}$ & $8.3_{4.2}$ & $9.1_{2.2}$ & $7.0_{1.8}$ & 1427 \\
        Qwen3-1.7B & 1.7B & $34.4_{0.0}$ & $20.6_{0.7}$ & $34.4_{0.0}$ & $22.8_{0.2}$ & 3 & $40.7_{4.2}$ & $15.9_{6.2}$ & $14.1_{3.3}$ & $12.9_{3.3}$ & 1624 \\
        Qwen3-4B & 4B & $28.7_{0.2}$ & $39.4_{3.3}$ & $28.7_{0.2}$ & $25.0_{0.4}$ & 2 & $43.3_{2.0}$ & $18.5_{5.8}$ & $12.7_{4.1}$ & $12.8_{3.9}$ & 1823 \\
        Qwen3-8B & 8B & $49.6_{0.4}$ & $28.7_{0.9}$ & $49.6_{0.4}$ & $35.3_{0.6}$ & 2 & $50.9_{3.1}$ & $37.6_{3.3}$ & $26.1_{3.8}$ & $27.1_{4.0}$ & 1443 \\
        Qwen3-14B & 14B & $47.8_{0.0}$ & $52.4_{0.1}$ & $47.8_{0.0}$ & $40.8_{0.1}$ & 3 & $54.9_{2.0}$ & $23.5_{1.3}$ & $20.3_{0.3}$ & $19.5_{0.3}$ & 1219 \\
        Qwen3-32B & 32B & $68.2_{0.7}$ & $71.5_{0.9}$ & $68.2_{0.7}$ & $66.3_{0.7}$ & 3 & $65.6_{3.5}$ & $46.6_{6.4}$ & $47.5_{5.0}$ & $44.9_{4.7}$ & 608 \\
        \midrule
        DS-r1-distill-Qwen & 32B &  - & - & - & - & - & $63.7_{1.4}$ & $62.7_{6.6}$ & $57.6_{4.2}$ & $53.3_{2.8}$ & 326 \\
        DeepSeek-r1-0528 & 37/671B &  - & - & - & - & - & $66.3_{0.5}$ & $61.6_{7.8}$ & $60.0_{4.7}$ & $53.1_{3.6}$ & 291 \\
        GPT-5-mini (low) & Unk. &  - & - & - & - & - & $65.9_{0.5}$ & $69.4_{3.5}$ & $65.9_{0.5}$ & $59.8_{1.4}$ & 70 \\
        GPT-5-mini (high) & Unk. &  - & - & - & - & - & $66.1_{1.7}$ & $72.7_{7.5}$ & $66.1_{1.7}$ & $60.2_{3.2}$ & 531 \\
        \bottomrule
    \end{tabular}}
    \caption{Details on \textbf{Task (1) Discourse Marker Understanding} performance on \textbf{\textit{Just-Manual}} dataset, using \textbf{``definition''} prompting strategy.
    }
    \label{tab:app-1-1-2}
\end{table*}

\begin{table*}[htbp]
    \centering
    \resizebox{\textwidth}{!}{
    \begin{tabular}{llcccccccccc}
        \toprule
        & & \multicolumn{5}{c}{{w/o reasoning}} & \multicolumn{5}{c}{w/ reasoning} \\
        \cmidrule(lr){3-7} \cmidrule(lr){8-12}
        Model & Size & Acc & weighted-Pre & weighted-Rec & weighted-F1 & tok. & Acc & weighted-Pre & weighted-Rec & weighted-F1 & tok. \\
        \midrule
        Qwen3-0.6B & 0.6B & $20.4_{1.2}$ & $7.2_{0.4}$ & $20.4_{1.2}$ & $10.4_{0.7}$ & 3 & $26.2_{1.8}$ & $6.0_{1.9}$ & $7.9_{1.5}$ & $6.0_{1.4}$ & 1436 \\
        Qwen3-1.7B & 1.7B & $40.4_{0.5}$ & $25.4_{3.5}$ & $40.4_{0.5}$ & $28.1_{0.9}$ & 3 & $40.2_{2.2}$ & $16.9_{3.3}$ & $16.3_{1.8}$ & $14.5_{1.7}$ & 1587 \\
        Qwen3-4B & 4B & $34.7_{0.4}$ & $30.6_{0.2}$ & $34.7_{0.4}$ & $28.9_{0.4}$ & 3 & $44.2_{2.7}$ & $29.5_{10.8}$ & $21.9_{8.8}$ & $22.1_{8.9}$ & 1942 \\
        Qwen3-8B & 8B & $42.2_{0.0}$ & $61.6_{0.0}$ & $42.2_{0.0}$ & $33.5_{0.0}$ & 2 & $46.4_{2.9}$ & $34.1_{3.7}$ & $22.8_{2.9}$ & $24.3_{2.7}$ & 1505 \\
        Qwen3-14B & 14B & $56.9_{0.4}$ & $56.2_{1.5}$ & $56.9_{0.4}$ & $49.3_{0.7}$ & 3 & $55.6_{1.3}$ & $22.5_{5.7}$ & $19.9_{5.4}$ & $19.6_{5.2}$ & 1574 \\
        Qwen3-32B & 32B & $69.8_{1.5}$ & $69.1_{1.9}$ & $69.8_{1.5}$ & $67.8_{1.9}$ & 3 & $64.4_{2.0}$ & $42.1_{8.8}$ & $41.0_{8.1}$ & $39.2_{7.4}$ & 613 \\
        \midrule
        DS-r1-distill-Qwen & 32B &  - & - & - & - & - & $59.6_{2.9}$ & $33.5_{3.1}$ & $32.8_{3.7}$ & $31.2_{3.3}$ & 313 \\
        DeepSeek-r1-0528 & 37/671B &  - & - & - & - & - & $67.0_{1.0}$ & $52.0_{3.0}$ & $52.7_{3.9}$ & $48.0_{3.4}$ & 421 \\
        GPT-5-mini (low) & Unk. &  - & - & - & - & - & $65.6_{0.0}$ & $64.1_{0.4}$ & $65.6_{0.0}$ & $60.7_{0.1}$ & 74 \\
        GPT-5-mini (high) & Unk. &  - & - & - & - & - & $65.6_{1.1}$ & $64.0_{1.8}$ & $65.6_{1.1}$ & $60.3_{1.7}$ & 528 \\
        \bottomrule
    \end{tabular}}
    \caption{Details on \textbf{Task (1) Discourse Marker Understanding} performance on \textbf{\textit{Just-Manual}} dataset, using \textbf{``definition+example''} prompting strategy.
    }
    \label{tab:app-1-1-3}
\end{table*}

\begin{table*}[htbp]
    \centering
    \resizebox{\textwidth}{!}{
    \begin{tabular}{llcccccccccc}
        \toprule
        & & \multicolumn{5}{c}{{w/o reasoning}} & \multicolumn{5}{c}{w/ reasoning} \\
        \cmidrule(lr){3-7} \cmidrule(lr){8-12}
        Model & Size & Acc & weighted-Pre & weighted-Rec & weighted-F1 & tok. & Acc & weighted-Pre & weighted-Rec & weighted-F1 & tok. \\
        \midrule
        Qwen3-0.6B & 0.6B & $14.6_{0.8}$ & $10.3_{5.0}$ & $16.7_{0.4}$ & $5.6_{1.3}$ & 4 & $25.5_{1.3}$ & $18.0_{2.6}$ & $30.4_{7.3}$ & $17.1_{1.0}$ & 260 \\
        Qwen3-1.7B & 1.7B & $13.4_{0.0}$ & $16.3_{5.3}$ & $11.1_{0.4}$ & $4.3_{0.2}$ & 3 & $25.9_{0.8}$ & $14.7_{1.6}$ & $24.9_{1.0}$ & $15.7_{1.1}$ & 491 \\
        Qwen3-4B & 4B & $24.8_{1.0}$ & $22.0_{1.1}$ & $21.4_{0.9}$ & $14.4_{0.7}$ & 3 & $41.5_{0.9}$ & $32.2_{2.9}$ & $33.8_{0.8}$ & $28.0_{0.5}$ & 537 \\
        Qwen3-8B & 8B & $34.5_{0.5}$ & $27.1_{1.3}$ & $29.4_{0.4}$ & $21.4_{0.4}$ & 2 & $46.0_{1.1}$ & $33.9_{5.7}$ & $36.0_{0.5}$ & $30.5_{0.5}$ & 610 \\
        Qwen3-14B & 14B & $33.8_{1.1}$ & $37.2_{1.6}$ & $32.2_{0.8}$ & $24.7_{0.9}$ & 2 & $43.2_{1.9}$ & $36.4_{2.2}$ & $33.8_{2.5}$ & $30.8_{2.3}$ & 463 \\
        Qwen3-32B & 32B & $29.4_{0.8}$ & $29.1_{1.7}$ & $25.9_{0.3}$ & $19.3_{0.7}$ & 3 & $48.5_{1.7}$ & $35.3_{4.0}$ & $36.9_{1.3}$ & $32.7_{1.5}$ & 415 \\
        \midrule
        DS-r1-distill-Qwen & 32B &  - & - & - & - & - & $31.8_{3.3}$ & $33.4_{6.8}$ & $25.7_{4.3}$ & $21.7_{4.2}$ & 291 \\
        DeepSeek-r1-0528 & 37/671B &  - & - & - & - & - & $48.1_{4.0}$ & $10.1_{3.9}$ & $9.6_{4.1}$ & $9.4_{3.9}$ & 419 \\
        GPT-5-mini (low) & Unk. &  - & - & - & - & - & $56.4_{1.1}$ & $36.8_{5.2}$ & $38.2_{2.3}$ & $34.6_{1.5}$ & 103 \\
        GPT-5-mini (high) & Unk. &  - & - & - & - & - & $57.4_{1.7}$ & $37.5_{3.6}$ & $37.4_{0.4}$ & $35.2_{0.1}$ & 818 \\
        \bottomrule
    \end{tabular}}
    \caption{Details on \textbf{Task (1) Discourse Marker Understanding} performance on \textbf{\textit{Just-Subtitle}} dataset, using \textbf{basic} prompting strategy.
    }    
    \label{tab:app-1-1-4}
\end{table*}

\begin{table*}[htbp]
    \centering
    \resizebox{\textwidth}{!}{
    \begin{tabular}{llcccccccccc}
        \toprule
        & & \multicolumn{5}{c}{{w/o reasoning}} & \multicolumn{5}{c}{w/ reasoning} \\
        \cmidrule(lr){3-7} \cmidrule(lr){8-12}
        Model & Size & Acc & weighted-Pre & weighted-Rec & weighted-F1 & tok. & Acc & weighted-Pre & weighted-Rec & weighted-F1 & tok. \\
        \midrule
        Qwen3-0.6B & 0.6B & $13.7_{0.4}$ & $7.3_{0.6}$ & $15.1_{1.0}$ & $6.2_{0.9}$ & 4 & $19.3_{1.2}$ & $3.7_{1.2}$ & $3.4_{1.2}$ & $2.7_{0.2}$ & 1426 \\
        Qwen3-1.7B & 1.7B & $19.7_{0.3}$ & $9.5_{0.1}$ & $19.8_{0.2}$ & $10.8_{0.2}$ & 3 & $26.2_{1.5}$ & $9.5_{2.0}$ & $7.6_{0.8}$ & $6.4_{0.7}$ & 1705 \\
        Qwen3-4B & 4B & $29.8_{0.3}$ & $32.2_{0.2}$ & $27.8_{0.5}$ & $20.5_{0.3}$ & 3 & $30.9_{2.4}$ & $11.6_{4.2}$ & $7.6_{2.6}$ & $7.2_{2.4}$ & 1667 \\
        Qwen3-8B & 8B & $31.3_{0.5}$ & $30.8_{2.9}$ & $46.1_{0.7}$ & $27.4_{1.1}$ & 3 & $32.3_{1.7}$ & $19.5_{2.6}$ & $13.6_{3.1}$ & $12.5_{1.9}$ & 1667 \\
        Qwen3-14B & 14B & $31.3_{0.5}$ & $35.2_{1.6}$ & $46.1_{1.0}$ & $27.3_{0.5}$ & 3 & $32.9_{1.7}$ & $9.6_{4.4}$ & $8.3_{3.5}$ & $7.7_{3.3}$ & 1328 \\
        Qwen3-32B & 32B & $32.8_{0.6}$ & $30.3_{0.4}$ & $30.5_{0.5}$ & $25.0_{0.6}$ & 3 & $48.9_{2.0}$ & $18.3_{3.2}$ & $18.3_{3.4}$ & $17.7_{3.3}$ & 569 \\
        \midrule
        DS-r1-distill-Qwen & 32B &  - & - & - & - & - & $40.5_{2.5}$ & $39.2_{5.8}$ & $33.2_{3.6}$ & $29.3_{3.6}$ & 308 \\
        DeepSeek-r1-0528 & 37/671B &  - & - & - & - & - & $60.2_{1.4}$ & $35.8_{6.1}$ & $34.1_{6.0}$ & $32.5_{6.1}$ & 321 \\
        GPT-5-mini (low) & Unk. &  - & - & - & - & - & $58.4_{1.1}$ & $43.6_{2.4}$ & $41.7_{1.2}$ & $39.3_{1.3}$ & 65 \\
        GPT-5-mini (high) & Unk. &  - & - & - & - & - & $63.1_{0.0}$ & $43.7_{1.7}$ & $44.2_{0.0}$ & $41.9_{1.0}$ & 509 \\
        \bottomrule
    \end{tabular}}
    \caption{Details on \textbf{Task (1) Discourse Marker Understanding} performance on \textbf{\textit{Just-Subtitle}} dataset, using \textbf{``definition''} prompting strategy.
    }    
    \label{tab:app-1-1-5}
\end{table*}

\begin{table*}[htbp]
    \centering
    \resizebox{\textwidth}{!}{
    \begin{tabular}{llcccccccccc}
        \toprule
        & & \multicolumn{5}{c}{{w/o reasoning}} & \multicolumn{5}{c}{w/ reasoning} \\
        \cmidrule(lr){3-7} \cmidrule(lr){8-12}
        Model & Size & Acc & weighted-Pre & weighted-Rec & weighted-F1 & tok. & Acc & weighted-Pre & weighted-Rec & weighted-F1 & tok. \\
        \midrule
        Qwen3-0.6B & 0.6B & $19.1_{0.9}$ & $7.8_{1.1}$ & $18.4_{0.7}$ & $9.6_{0.4}$ & 3 & $16.9_{1.8}$ & $3.2_{0.5}$ & $3.8_{1.1}$ & $2.4_{0.5}$ & 1439 \\
        Qwen3-1.7B & 1.7B & $20.3_{0.4}$ & $9.9_{0.4}$ & $20.1_{0.5}$ & $13.0_{0.3}$ & 2 & $25.5_{1.9}$ & $7.1_{1.5}$ & $5.5_{0.5}$ & $5.0_{0.4}$ & 1743 \\
        Qwen3-4B & 4B & $29.9_{0.3}$ & $24.8_{0.2}$ & $27.7_{0.4}$ & $21.7_{0.2}$ & 3 & $30.3_{1.1}$ & $21.4_{10.0}$ & $13.3_{6.6}$ & $12.8_{6.7}$ & 1944 \\
        Qwen3-8B & 8B & $30.9_{0.6}$ & $28.3_{2.2}$ & $44.9_{0.6}$ & $25.5_{0.7}$ & 2 & $32.2_{1.4}$ & $22.9_{5.1}$ & $16.0_{4.7}$ & $14.0_{2.8}$ & 1593 \\
        Qwen3-14B & 14B & $36.2_{0.4}$ & $38.2_{1.6}$ & $48.6_{0.5}$ & $32.3_{0.7}$ & 3 & $33.2_{1.3}$ & $14.6_{5.1}$ & $11.5_{4.8}$ & $10.8_{4.1}$ & 1427 \\
        Qwen3-32B & 32B & $35.7_{1.1}$ & $34.1_{1.6}$ & $46.1_{0.5}$ & $30.3_{0.8}$ & 2 & $47.1_{1.2}$ & $15.0_{1.1}$ & $14.3_{1.1}$ & $13.9_{1.1}$ & 718 \\
        \midrule
        DS-r1-distill-Qwen & 32B &  - & - & - & - & - & $35.8_{0.3}$ & $17.9_{3.7}$ & $13.4_{3.2}$ & $12.5_{2.7}$ & 375 \\
        DeepSeek-r1-0528 & 37/671B &  - & - & - & - & - & $59.5_{3.6}$ & $27.0_{1.3}$ & $22.6_{2.7}$ & $21.5_{2.6}$ & 380 \\
        GPT-5-mini (low) & Unk. &  - & - & - & - & - & $62.9_{0.6}$ & $48.0_{1.7}$ & $44.0_{0.5}$ & $42.1_{0.3}$ & 87 \\
        GPT-5-mini (high) & Unk. &  - & - & - & - & - & $63.4_{0.3}$ & $46.5_{1.3}$ & $40.7_{0.0}$ & $40.1_{0.2}$ & 534 \\
        \bottomrule
    \end{tabular}}
    \caption{Details on \textbf{Task (1) Discourse Marker Understanding} performance on \textbf{\textit{Just-Subtitle}} dataset, using \textbf{``definition+example''} prompting strategy.
    }    
    \label{tab:app-1-1-6}
\end{table*}

\begin{table*}[htbp]
    \centering
    \resizebox{\textwidth}{!}{
    \begin{tabular}{llcccccccccc}
        \toprule
        & & \multicolumn{5}{c}{{w/o reasoning}} & \multicolumn{5}{c}{w/ reasoning} \\
        \cmidrule(lr){3-7} \cmidrule(lr){8-12}
        Model & Size & Acc & weighted-Pre & weighted-Rec & weighted-F1 & tok. & Acc & weighted-Pre & weighted-Rec & weighted-F1 & tok. \\
        \midrule
        Qwen3-0.6B & 0.6B & $44.6_{0.0}$ & $19.9_{0.0}$ & $44.6_{0.0}$ & $27.5_{0.0}$ & 3 & $22.2_{1.3}$ & $24.9_{2.8}$ & $22.2_{1.3}$ & $19.0_{1.5}$ & 277 \\
        Qwen3-1.7B & 1.7B & $30.3_{0.7}$ & $20.1_{0.2}$ & $30.3_{0.7}$ & $24.1_{0.4}$ & 2 & $23.9_{0.6}$ & $26.2_{4.1}$ & $23.9_{0.6}$ & $17.5_{0.9}$ & 492 \\
        Qwen3-4B & 4B & $25.7_{0.2}$ & $8.3_{0.2}$ & $25.7_{0.2}$ & $11.9_{0.2}$ & 1 & $32.3_{1.2}$ & $24.5_{17.9}$ & $32.3_{1.2}$ & $21.0_{0.7}$ & 569 \\
        Qwen3-8B & 8B & $18.6_{0.1}$ & $38.2_{2.5}$ & $18.6_{0.1}$ & $6.7_{0.2}$ & 2 & $32.3_{0.7}$ & $34.1_{0.5}$ & $32.3_{0.7}$ & $24.0_{0.7}$ & 512 \\
        Qwen3-14B & 14B & $34.3_{0.1}$ & $42.6_{0.6}$ & $34.3_{0.1}$ & $25.6_{0.1}$ & 2 & $32.1_{1.0}$ & $30.0_{3.4}$ & $32.1_{1.0}$ & $23.6_{0.9}$ & 509 \\
        Qwen3-32B & 32B & $36.5_{1.3}$ & $50.7_{5.3}$ & $36.5_{1.3}$ & $38.1_{1.3}$ & 2 & $34.3_{0.4}$ & $19.7_{0.4}$ & $34.3_{0.4}$ & $23.8_{0.4}$ & 373 \\
        \midrule
        DS-r1-distill-Qwen & 32B &  - & - & - & - & - & $33.1_{1.1}$ & $25.9_{1.1}$ & $33.1_{1.1}$ & $24.9_{0.9}$ & 276 \\
        DeepSeek-r1-0528 & 37/671B &  - & - & - & - & - & $31.2_{0.8}$ & $39.1_{1.2}$ & $31.2_{0.8}$ & $25.8_{0.4}$ & 373 \\
        GPT-5-mini (low) & Unk. &  - & - & - & - & - & $34.6_{0.4}$ & $25.4_{0.6}$ & $34.6_{0.4}$ & $25.5_{0.4}$ & 74 \\
        GPT-5-mini (high) & Unk. &  - & - & - & - & - & $35.5_{0.5}$ & $31.1_{5.6}$ & $35.5_{0.5}$ & $26.5_{0.2}$ & 375 \\
        \bottomrule
    \end{tabular}}
    \caption{Details on \textbf{Task (1) Discourse Marker Understanding} performance on \textbf{\textit{Otherwise}} dataset, using \textbf{basic} prompting strategy.
    }    
    \label{tab:app-1-2-1}
\end{table*}

\begin{table*}[htbp]
    \centering
    \resizebox{\textwidth}{!}{
    \begin{tabular}{llcccccccccc}
        \toprule
        & & \multicolumn{5}{c}{{w/o reasoning}} & \multicolumn{5}{c}{w/ reasoning} \\
        \cmidrule(lr){3-7} \cmidrule(lr){8-12}
        Model & Size & Acc & weighted-Pre & weighted-Rec & weighted-F1 & tok. & Acc & weighted-Pre & weighted-Rec & weighted-F1 & tok. \\
        \midrule
        Qwen3-0.6B & 0.6B & $44.6_{0.0}$ & $19.9_{0.0}$ & $44.6_{0.0}$ & $27.5_{0.0}$ & 2 & $22.7_{0.9}$ & $25.0_{3.6}$ & $22.7_{0.9}$ & $18.1_{1.4}$ & 402 \\
        Qwen3-1.7B & 1.7B & $24.2_{1.6}$ & $28.3_{1.7}$ & $24.2_{1.7}$ & $20.5_{1.6}$ & 1 & $21.2_{1.0}$ & $35.9_{6.0}$ & $21.2_{1.0}$ & $11.9_{1.1}$ & 515 \\
        Qwen3-4B & 4B & $30.3_{0.6}$ & $56.1_{0.5}$ & $30.3_{0.6}$ & $32.3_{0.4}$ & 2 & $21.8_{0.9}$ & $61.8_{4.9}$ & $21.8_{0.9}$ & $15.1_{1.7}$ & 547 \\
        Qwen3-8B & 8B & $33.6_{0.9}$ & $65.2_{1.1}$ & $33.6_{0.9}$ & $33.5_{0.9}$ & 2 & $39.0_{1.7}$ & $70.1_{1.8}$ & $39.0_{1.7}$ & $38.6_{1.9}$ & 1033 \\
        Qwen3-14B & 14B & $43.5_{0.4}$ & $57.8_{0.4}$ & $43.5_{0.4}$ & $44.3_{0.5}$ & 2 & $45.2_{2.6}$ & $70.9_{1.8}$ & $45.2_{2.6}$ & $44.6_{2.1}$ & 766 \\
        Qwen3-32B & 32B & $55.0_{0.2}$ & $64.0_{1.2}$ & $55.0_{0.2}$ & $54.9_{0.4}$ & 2 & $49.3_{2.3}$ & $71.3_{1.5}$ & $49.3_{2.3}$ & $50.3_{1.8}$ & 737 \\
        \midrule
        DS-r1-distill-Qwen & 32B &  - & - & - & - & - & $32.4_{2.3}$ & $60.0_{1.5}$ & $32.4_{2.3}$ & $31.4_{3.3}$ & 369 \\
        DeepSeek-r1-0528 & 37/671B &  - & - & - & - & - & $38.9_{1.9}$ & $70.3_{1.6}$ & $38.9_{1.9}$ & $45.2_{1.7}$ & 682 \\
        GPT-5-mini (low) & Unk. &  - & - & - & - & - & $36.8_{0.8}$ & $64.0_{2.8}$ & $36.8_{0.8}$ & $34.0_{1.4}$ & 138 \\
        GPT-5-mini (high) & Unk. &  - & - & - & - & - & $43.9_{1.0}$ & $74.5_{4.1}$ & $43.9_{1.0}$ & $43.6_{1.3}$ & 751 \\
        \bottomrule
    \end{tabular}}
    \caption{Details on \textbf{Task (1) Discourse Marker Understanding} performance on \textbf{\textit{Otherwise}} dataset, using \textbf{``definition''} prompting strategy.
    }    
    \label{tab:app-1-2-2}
\end{table*}

\begin{table*}[htbp]
    \centering
    \resizebox{\textwidth}{!}{
    \begin{tabular}{llcccccccccc}
        \toprule
        & & \multicolumn{5}{c}{{w/o reasoning}} & \multicolumn{5}{c}{w/ reasoning} \\
        \cmidrule(lr){3-7} \cmidrule(lr){8-12}
        Model & Size & Acc & weighted-Pre & weighted-Rec & weighted-F1 & tok. & Acc & weighted-Pre & weighted-Rec & weighted-F1 & tok. \\
        \midrule
        Qwen3-0.6B & 0.6B & $44.6_{0.0}$ & $19.9_{0.1}$ & $44.6_{0.0}$ & $27.5_{0.1}$ & 2 & $15.7_{1.4}$ & $32.0_{14.3}$ & $15.7_{1.4}$ & $11.8_{2.1}$ & 1399 \\
        Qwen3-1.7B & 1.7B & $21.0_{0.8}$ & $32.2_{2.0}$ & $21.0_{0.8}$ & $11.7_{0.8}$ & 2 & $16.2_{1.8}$ & $48.6_{9.0}$ & $16.2_{1.8}$ & $13.1_{1.4}$ & 1442 \\
        Qwen3-4B & 4B & $41.8_{0.8}$ & $56.5_{0.5}$ & $41.8_{0.8}$ & $44.5_{0.7}$ & 2 & $30.7_{1.9}$ & $61.7_{1.5}$ & $30.7_{1.9}$ & $33.8_{1.9}$ & 1585 \\
        Qwen3-8B & 8B & $46.7_{0.3}$ & $60.2_{1.0}$ & $46.7_{0.3}$ & $49.2_{0.4}$ & 2 & $45.6_{1.4}$ & $71.8_{2.2}$ & $45.6_{1.4}$ & $46.8_{1.3}$ & 1574 \\
        Qwen3-14B & 14B & $51.8_{0.7}$ & $62.9_{0.5}$ & $51.8_{0.7}$ & $53.8_{0.6}$ & 2 & $51.4_{1.6}$ & $73.7_{0.6}$ & $51.4_{1.6}$ & $53.9_{1.1}$ & 1582 \\
        Qwen3-32B & 32B & $63.8_{0.8}$ & $67.5_{0.4}$ & $63.8_{0.8}$ & $62.9_{0.6}$ & 2 & $66.1_{0.6}$ & $74.5_{0.5}$ & $66.1_{0.6}$ & $65.3_{0.6}$ & 873 \\
        \midrule
        DS-r1-distill-Qwen & 32B &  - & - & - & - & - & $44.3_{2.7}$ & $65.2_{2.7}$ & $44.3_{2.7}$ & $47.0_{2.8}$ & 453 \\
        DeepSeek-r1-0528 & 37/671B &  - & - & - & - & - & $58.8_{2.2}$ & $74.3_{0.7}$ & $58.8_{2.2}$ & $63.9_{1.4}$ & 656 \\
        GPT-5-mini (low) & Unk. &  - & - & - & - & - & $51.9_{2.4}$ & $73.9_{1.5}$ & $51.9_{2.4}$ & $54.8_{2.5}$ & 152 \\
        GPT-5-mini (high) & Unk. &  - & - & - & - & - & $71.8_{1.7}$ & $79.5_{0.6}$ & $71.8_{1.7}$ & $71.7_{1.8}$ & 937 \\
        \bottomrule
    \end{tabular}}
    \caption{Details on \textbf{Task (1) Discourse Marker Understanding} performance on \textbf{\textit{Otherwise}} dataset, using \textbf{``definition+example''} prompting strategy.
    }    
    \label{tab:app-1-2-3}
\end{table*}

\FloatBarrier

\subsection{Task (2) Temporal Reasoning}
\label{append-tr}

\Cref{tab:app-2-1-1,tab:app-2-1-2,tab:app-2-2-1,tab:app-2-2-2,tab:app-2-3-1}.

\begin{table*}[htbp]
    \centering
    \resizebox{\textwidth}{!}{
    \begin{tabular}{llcccccccccc}
        \toprule
        & & \multicolumn{5}{c}{{w/o reasoning}} & \multicolumn{5}{c}{w/ reasoning} \\
        \cmidrule(lr){3-7} \cmidrule(lr){8-12}
        Model & Size & micro-F1 & weighted-Pre & weighted-Rec & weighted-F1 & tok. & micro-F1 & weighted-Pre & weighted-Rec & weighted-F1 & tok. \\
        \midrule
        Qwen3-0.6B & 0.6B & $27.3_{9.3}$ & $21.9_{2.0}$ & $27.3_{9.3}$ & $16.6_{9.4}$ & 3 &$24.0_{0.8}$ & $32.5_{1.7}$ & $24.0_{0.8}$ & $21.0_{1.0}$ & 285 \\
        Qwen3-1.7B & 1.7B & $28.0_{0.5}$ & $30.9_{12.1}$ & $28.0_{0.5}$ & $23.2_{0.5}$ & 2 &$31.3_{0.3}$ & $37.4_{0.9}$ & $31.3_{0.3}$ & $30.5_{0.5}$ & 568 \\
        Qwen3-4B & 4B & $30.5_{0.3}$ & $41.0_{1.0}$ & $30.5_{0.3}$ & $25.8_{0.5}$ & 2 &$33.8_{0.5}$ & $44.2_{1.0}$ & $33.8_{0.5}$ & $29.6_{0.3}$ & 718 \\
        Qwen3-8B & 8B & $34.5_{0.4}$ & $40.6_{2.0}$ & $34.5_{0.4}$ & $29.1_{0.7}$ & 2 &$37.6_{1.0}$ & $46.7_{1.1}$ & $37.6_{1.0}$ & $33.6_{0.8}$ & 754 \\
        Qwen3-14B & 14B & $34.0_{0.2}$ & $47.9_{0.4}$ & $34.0_{0.2}$ & $30.1_{0.4}$ & 2 &$42.5_{0.1}$ & $49.7_{0.7}$ & $42.5_{0.1}$ & $42.1_{0.5}$ & 472 \\
        Qwen3-32B & 32B & $30.6_{0.5}$ & $42.9_{0.5}$ & $30.6_{0.5}$ & $22.8_{0.2}$ & 2 &$42.4_{0.4}$ & $53.8_{0.9}$ & $42.4_{0.4}$ & $36.6_{0.3}$ & 464 \\
        \midrule
        DS-r1-distill-Qwen & 32B &  - & - & - & - & - & $27.7_{2.7}$ & $49.2_{1.7}$ & $27.7_{2.7}$ & $26.5_{1.3}$ & 317 \\
        DeepSeek-r1-0528 & 37/671B &  - & - & - & - & - & $39.9_{0.0}$ & $50.9_{0.0}$ & $39.9_{0.0}$ & $33.0_{0.0}$ & 476 \\
        GPT-5-mini (low) & Unk. &  - & - & - & - & - & $41.9_{0.1}$ & $55.1_{0.7}$ & $41.9_{0.1}$ & $38.8_{0.1}$ & 144 \\
        GPT-5-mini (high) & Unk. &  - & - & - & - & - & $45.2_{0.3}$ & $56.3_{0.0}$ & $45.2_{0.3}$ & $41.4_{0.6}$ & 731 \\
        \bottomrule
    \end{tabular}}
    \caption{Details on \textbf{Task (2) Temporal Reasoning} performance on \textbf{TBD} dataset, using \textbf{``without context''} prompting strategy.}  
    \label{tab:app-2-1-1}
\end{table*}

\begin{table*}[htbp]
    \centering
    \resizebox{\textwidth}{!}{
    \begin{tabular}{llcccccccccc}
        \toprule
        & & \multicolumn{5}{c}{{w/o reasoning}} & \multicolumn{5}{c}{w/ reasoning} \\
        \cmidrule(lr){3-7} \cmidrule(lr){8-12}
        Model & Size & micro-F1 & Weighted-Pre & Weighted-Rec & Weighted-F1 & tok. & micro-F1 & Weighted-Pre & Weighted-Rec & Weighted-F1 & tok. \\
        \midrule
        Qwen3-0.6B & 0.6B & $32.0_{5.5}$ & $23.2_{0.5}$ & $32.0_{5.5}$ & $24.0_{4.7}$ & 3 &$21.0_{0.3}$ & $30.2_{0.7}$ & $21.0_{0.3}$ & $17.9_{0.2}$ & 346 \\
        Qwen3-1.7B & 1.7B & $23.4_{0.4}$ & $34.7_{1.2}$ & $23.4_{0.4}$ & $15.6_{0.8}$ & 2 &$26.2_{0.8}$ & $37.6_{2.1}$ & $26.2_{0.8}$ & $22.3_{0.9}$ & 559 \\
        Qwen3-4B & 4B & $28.3_{0.2}$ & $42.7_{1.5}$ & $28.3_{0.2}$ & $18.8_{0.3}$ & 2 &$31.7_{0.7}$ & $45.6_{1.6}$ & $31.7_{0.7}$ & $26.1_{0.6}$ & 778 \\
        Qwen3-8B & 8B & $25.4_{0.3}$ & $46.8_{2.9}$ & $25.4_{0.3}$ & $16.8_{0.4}$ & 2 &$35.3_{0.4}$ & $50.6_{1.6}$ & $35.3_{0.4}$ & $29.6_{0.6}$ & 760 \\
        Qwen3-14B & 14B & $31.1_{0.4}$ & $51.0_{0.5}$ & $31.1_{0.4}$ & $27.7_{0.7}$ & 3 &$39.8_{0.5}$ & $53.7_{0.9}$ & $39.8_{0.5}$ & $39.6_{0.5}$ & 501 \\
        Qwen3-32B & 32B & $27.8_{0.7}$ & $54.4_{4.9}$ & $27.8_{0.7}$ & $21.1_{1.2}$ & 3 &$40.8_{0.2}$ & $56.6_{2.0}$ & $40.8_{0.2}$ & $34.5_{0.5}$ & 475 \\
        \midrule
        DS-r1-distill-Qwen & 32B &  - & - & - & - & - & $23.8_{2.5}$ & $52.8_{3.0}$ & $23.8_{2.5}$ & $22.5_{0.7}$ & 283 \\
        DeepSeek-r1-0528 & 37/671B &  - & - & - & - & - & $33.1_{0.0}$ & $57.3_{0.0}$ & $33.1_{0.0}$ & $28.6_{0.0}$ & 457 \\
        GPT-5-mini (low) & Unk. &  - & - & - & - & - & $38.7_{0.2}$ & $59.6_{0.2}$ & $38.7_{0.2}$ & $34.4_{0.6}$ & 166 \\
        GPT-5-mini (high) & Unk. &  - & - & - & - & - & $43.4_{0.3}$ & $61.9_{1.9}$ & $43.4_{0.3}$ & $38.0_{0.4}$ & 870 \\
        \bottomrule
    \end{tabular}}
    \caption{Details on \textbf{Task (2) Temporal Reasoning} performance on \textbf{TBD} dataset, using \textbf{``with context''} prompting strategy.}  
    \label{tab:app-2-1-2}
\end{table*}

\begin{table*}[htbp]
    \centering
    \resizebox{\textwidth}{!}{
    \begin{tabular}{llcccccccccc}
        \toprule
        & & \multicolumn{5}{c}{{w/o reasoning}} & \multicolumn{5}{c}{w/ reasoning} \\
        \cmidrule(lr){3-7} \cmidrule(lr){8-12}
        Model & Size & micro-F1 & weighted-Pre & weighted-Rec & weighted-F1 & tok. & micro-F1 & weighted-Pre & weighted-Rec & weighted-F1 & tok. \\
        \midrule
        Qwen3-0.6B & 0.6B & $12.6_{0.1}$ & $11.7_{7.8}$ & $12.6_{0.1}$ & $2.9_{0.2}$ & 3 &$25.8_{1.1}$ & $29.1_{1.1}$ & $25.8_{1.1}$ & $25.6_{1.0}$ & 392 \\
        Qwen3-1.7B & 1.7B & $12.5_{0.0}$ & $1.6_{0.0}$ & $12.5_{0.0}$ & $2.8_{0.0}$ & 2 &$21.9_{0.6}$ & $29.8_{2.5}$ & $21.9_{0.6}$ & $17.9_{0.6}$ & 594 \\
        Qwen3-4B & 4B & $19.4_{0.5}$ & $11.6_{0.3}$ & $19.4_{0.5}$ & $12.5_{0.3}$ & 2 &$25.4_{0.4}$ & $37.1_{0.9}$ & $25.4_{0.4}$ & $20.1_{0.5}$ & 879 \\
        Qwen3-8B & 8B & $18.2_{0.2}$ & $22.3_{0.4}$ & $18.2_{0.2}$ & $11.3_{0.1}$ & 2 &$28.8_{0.3}$ & $42.0_{1.6}$ & $28.8_{0.3}$ & $23.1_{0.4}$ & 846 \\
        Qwen3-14B & 14B & $26.1_{0.2}$ & $57.2_{2.6}$ & $26.1_{0.2}$ & $13.8_{0.1}$ & 2 &$29.5_{0.5}$ & $39.2_{1.5}$ & $29.5_{0.5}$ & $22.0_{0.5}$ & 582 \\
        Qwen3-32B & 32B & $27.5_{0.1}$ & $37.7_{1.5}$ & $27.5_{0.1}$ & $15.3_{0.7}$ & 2 &$31.6_{0.4}$ & $43.3_{2.5}$ & $31.6_{0.4}$ & $23.9_{0.5}$ & 509 \\
        \midrule
        DS-r1-distill-Qwen & 32B &  - & - & - & - & - & $18.6_{2.6}$ & $42.4_{1.6}$ & $18.6_{2.6}$ & $15.8_{1.4}$ & 298 \\
        DeepSeek-r1-0528 & 37/671B &  - & - & - & - & - & $22.2_{0.0}$ & $48.8_{0.0}$ & $22.2_{0.0}$ & $21.0_{0.0}$ & 478 \\
        GPT-5-mini (low) & Unk. &  - & - & - & - & - & $34.4_{0.2}$ & $49.6_{0.5}$ & $34.4_{0.2}$ & $30.5_{0.3}$ & 204 \\
        GPT-5-mini (high) & Unk. &  - & - & - & - & - & $36.7_{0.3}$ & $52.7_{0.5}$ & $36.7_{0.3}$ & $34.6_{0.2}$ & 1203 \\
        \bottomrule
    \end{tabular}}
    \caption{Details on \textbf{Task (2) Temporal Reasoning} performance on \textbf{TDD-Man} dataset, using \textbf{``without context''} prompting strategy.}  
    \label{tab:app-2-2-1}
\end{table*}

\begin{table*}[htbp]
    \centering
    \resizebox{\textwidth}{!}{
    \begin{tabular}{llcccccccccc}
        \toprule
        & & \multicolumn{5}{c}{{w/o reasoning}} & \multicolumn{5}{c}{w/ reasoning} \\
        \cmidrule(lr){3-7} \cmidrule(lr){8-12}
        Model & Size & micro-F1 & Weighted-Pre & Weighted-Rec & Weighted-F1 & tok. & micro-F1 & Weighted-Pre & Weighted-Rec & Weighted-F1 & tok. \\
        \midrule
        Qwen3-0.6B & 0.6B & $17.4_{3.4}$ & $10.0_{3.3}$ & $17.4_{3.4}$ & $8.0_{3.5}$ & 3 &$23.1_{0.7}$ & $31.2_{1.0}$ & $23.1_{0.7}$ & $19.2_{0.5}$ & 297 \\
        Qwen3-1.7B & 1.7B & $13.0_{0.3}$ & $29.5_{1.7}$ & $13.0_{0.3}$ & $5.8_{0.1}$ & 2 &$21.4_{0.1}$ & $36.7_{3.6}$ & $21.4_{0.1}$ & $18.4_{0.5}$ & 624 \\
        Qwen3-4B & 4B & $19.4_{0.2}$ & $11.0_{0.4}$ & $19.4_{0.2}$ & $12.6_{0.2}$ & 2 &$23.8_{0.1}$ & $40.6_{0.2}$ & $23.8_{0.1}$ & $19.9_{0.1}$ & 833 \\
        Qwen3-8B & 8B & $23.4_{0.3}$ & $20.8_{1.3}$ & $23.4_{0.3}$ & $14.3_{0.1}$ & 2 &$27.6_{0.8}$ & $44.8_{2.3}$ & $27.6_{0.8}$ & $24.8_{0.8}$ & 904 \\
        Qwen3-14B & 14B & $22.8_{0.1}$ & $46.7_{8.6}$ & $22.8_{0.1}$ & $13.7_{0.1}$ & 2 &$24.0_{0.6}$ & $39.6_{2.4}$ & $24.0_{0.6}$ & $20.4_{0.7}$ & 546 \\
        Qwen3-32B & 32B & $24.9_{0.5}$ & $29.6_{0.4}$ & $24.9_{0.5}$ & $15.0_{0.5}$ & 2 &$28.9_{1.4}$ & $44.1_{3.3}$ & $28.9_{1.4}$ & $24.4_{1.3}$ & 460 \\
        \midrule
        DS-r1-distill-Qwen & 32B &  - & - & - & - & - & $16.9_{2.8}$ & $42.8_{2.9}$ & $16.9_{2.8}$ & $15.4_{1.6}$ & 298 \\
        DeepSeek-r1-0528 & 37/671B &  - & - & - & - & - & $25.9_{0.0}$ & $47.1_{0.0}$ & $25.9_{0.0}$ & $22.7_{0.0}$ & 433 \\
        GPT-5-mini (low) & Unk. &  - & - & - & - & - & $27.2_{0.5}$ & $53.3_{0.8}$ & $27.2_{0.5}$ & $28.6_{0.6}$ & 171 \\
        GPT-5-mini (high) & Unk. &  - & - & - & - & - & $27.9_{0.2}$ & $53.5_{0.0}$ & $27.9_{0.2}$ & $30.1_{0.1}$ & 935 \\
        \bottomrule
    \end{tabular}}
    \caption{Details on \textbf{Task (2) Temporal Reasoning} performance on \textbf{TDD-Man} dataset, using \textbf{``with context''} prompting strategy.}  
    \label{tab:app-2-2-2}
\end{table*}

\begin{table*}[htbp]
    \centering
    \resizebox{\textwidth}{!}{
    \begin{tabular}{llccccccccc}
        \toprule    
        Model & Size & Acc (all) & Acc (tz) & Acc (as) & Acc (scd) & Acc (mop) & Acc (dur) & Acc (cmp) & Acc (trk) & tok. \\
        \midrule
        & \multicolumn{10}{c}{baselines} \\ 
        \midrule
        Claude-3-Sonnet & Unk. & - & $74.00$ & $58.57$ & $29.60$ & $26.57$ & $15.00$ & $39.14$ & $40.40$ & - \\
        GPT-4 & Unk. & - & $88.00$ & $76.28$ & $43.60$ & $46.86$ & $16.00$ & $63.14$ & $45.60$ & - \\
        Gemini 1.5 Pro & Unk. & - & $90.00$ & $71.14$ & $40.00$ & $62.57$ & $13.50$ & $55.43$ & $41.20$ & - \\
        \midrule
        & \multicolumn{10}{c}{{w/o reasoning}} \\  
        \midrule
        Qwen3-0.6B & 0.6B & $7.0_{0.1}$ & $17.0_{1.0}$ & $8.9_{0.2}$ & $9.7_{3.1}$ & $0.0_{0.0}$ & $0.0_{0.0}$ & $8.4_{1.3}$ & $10.8_{1.6}$ & 15 \\
        Qwen3-1.7B & 1.7B & $8.9_{0.2}$ & $22.7_{1.3}$ & $14.4_{0.5}$ & $7.6_{1.2}$ & $0.3_{0.0}$ & $2.0_{0.0}$ & $11.7_{0.6}$ & $10.8_{0.4}$ & 46 \\
\bottomrule
    \end{tabular}}
    \caption{Details on \textbf{Task (2) Temporal Reasoning} performance on \textbf{ToT-Arithmetic} dataset.
    Question type abbreviations: all -- average score, tz -- Timezone, as -- AddSubtract, scd -- Schedule, mop -- MultiOperations, dur -- Duration, cmp -- Compare, trk -- Trick; see explanation on question types in \citet{fatemi2024testoftime}.
    }  
    \label{tab:app-2-3-1}
\end{table*}

\FloatBarrier

\subsection{Task (3) Discourse Relation Recognition}
\label{append-dr}

\Cref{tab:app-3-0,tab:app-3-1-1,tab:app-3-1-2,tab:app-3-1-3,tab:app-3-1-4}.

\begin{table*}[htbp]
    \centering
    \resizebox{\textwidth}{!}{
    \begin{tabular}{llcccccccccccccccc}
        \toprule
        & Model
        & Ces & Deu & Eng & Eus & Fas & Fra & Ita & Nld & Pcm & Pol & Por & Rus & Spa & Tha & Tur & Zho\\
        \midrule
        \multirow{10}{*}{\rotatebox[origin=c]{90}{\small BeDiscovER}} 
        & Qwen3-1.7B
        & $9.3$ & $10.7$ & $22.1$ & $16.3$ & $18.7$ & $25.0$ & $19.5$ & $16.7$ & $14.6$ & $25$ & $23.5$ & $22.5$ & $19.4$ & $38.7$ & $18.3$ & $18.0$ \\
        & Qwen3-14B 
        & $25.5$ & $25.0$ & $36.1$ & $30.4$ & $29.9$ & $41.8$ & $30.8$ & $30.4$ & $26.0$ & $41.4$ & $38.3$ & $38.1$ & $36.0$ & $55.8$ & $34.6$ & $32.0$ \\
        & Qwen3-32B 
        & $28.4$ & $26.5$ & $37.5$ & $30.7$ & $\underline{32.7}$ & $\underline{43.6}$ & $32.7$ & $33.0$ & $24.2$ & $40.3$ & $41.7$ & $\underline{40.7}$ & $37.0$ & $53.7$ & $34.1$ & $33.3$ \\
        & DS-r1-distill-Qwen 
        & $26.0$ & $24.8$ & $32.9$ & $28.1$ & $26.9$ & $37.7$ & $25.3$ & $31.8$ & $22.4$ & $38.7$ & $36.6$ & $27.2$ & $36.1$ & $38.4$ & $27.6$ & $27.8$ \\
        & DeepSeek-r1-0528
        & $\underline{39.8}$ & $38.1$ & $\underline{42.1}$ & $38.4$ & $31.7$ & $43.5$ & $\underline{40.4}$ & $36.9$ & $28.0$ & $\underline{51.3}$ & $47.6$ & $38.4$ & $\underline{48.1}$ & $\underline{61.1}$ & $42.2$ & $\underline{42.1}$ \\
        & GPT-5-mini (low)
        & $38.2$ & $\underline{39.2}$ & $43.7$ & $\underline{40.9}$ & $29.9$ & $40.5$ & $36.0$ & $\underline{35.1}$ & $\underline{31.5}$ & $47.3$ & $\underline{48.0}$ & $40.6$ & $43.9$ & $59.1$ & $\underline{42.8}$ & $39.4$ \\
        & GPT-5-mini (high) 
        & $\mathbf{43.2}$ & $\mathbf{44.3}$ & $\mathbf{47.1}$ & $\mathbf{42.5}$ & $\mathbf{33.0}$ & $\mathbf{44.3}$ & $\mathbf{40.8}$ & $\mathbf{38.4}$ & $\mathbf{36.9}$ & $\mathbf{56.7}$ & $\mathbf{51.4}$ & $\mathbf{45.1}$ & $\mathbf{50.0}$ & $\mathbf{64.9}$ & $\mathbf{45.6}$ & $\mathbf{45.8}$ \\

        \cdashline{2-18}
        & GPT-4o-mini 
        & $30.1$ & $22.4$ & $29.9$ & $23.6$ & $17.4$ & $28.8$ & $27.1$ & $22.6$ & $29.6$ & $28.6$ & $33.6$ & $18.3$ & $9.8$ & $43.7$ & $29.9$ & $25.3$ \\
        & Llama-4-Scout 
        & $34.1$ & $32.5$ & $36.6$ & $31.0$ & $33.5$ & $36.1$ & $22.9$ & $32.6$ & $26.1$ & $44.8$ & $41.3$ & $32.9$ & $37.1$ & $60.4$ & $33.2$ & $35.1$ \\
        & Qwen2.5-72B 
        & $34.8$ & $26.9$ & $34.8$ & $26.9$ & $31.4$ & $36.5$ & $23.1$ & $32.6$ & $27.3$ & $37.6$ & $40.5$ & $25.8$ & $39.6$ & $47.2$ & $28.9$ & $32.7$ \\
        
        \midrule        
        \multirow{7}{*}{\rotatebox[origin=c]{90}{\small Supervised}} 
        & \textit{\#Datasets} & \textit{1} & \textit{2} & \textit{14} & \textit{1} & \textit{1} & \textit{1} & \textit{1} & \textit{1} & \textit{1} & \textit{1} & \textit{3} & \textit{1} & \textit{2} & \textit{1} & \textit{2} & \textit{5} \\ 
        & \textit{\#Train examples} & \textit{978} & \textit{4,072} & \textit{133,731} & \textit{2,533} & \textit{4,100} & \textit{2,177} & \textit{944} & \textit{1,608} & \textit{7,834} & \textit{7,040} & \textit{12,942} & \textit{20,014} & \textit{2,679} & \textit{8,274} & \textit{2,444} & \textit{22,000} \\
        & DeDisCo \shortcite{ju2025dedisco} & $\mathbf{56.1}$ & $\mathbf{65.5}$ & $\mathbf{78.4}$ & $50.1$ & $\underline{59.3}$ & $\mathbf{60.1}$ & $\mathbf{72.0}$ & $\mathbf{67.4}$ & $59.9$ & $\mathbf{72.0}$ & $\mathbf{75.9}$ & $\mathbf{73.9}$ & $\mathbf{72.3}$ & $\mathbf{97.1}$ & $\mathbf{64.0}$ & $\mathbf{78.6}$ \\
        & HITS \shortcite{banerjee2025hits} & $\underline{53.4}$ & $\underline{61.4}$ & $\underline{74.7}$ & $\underline{54.0}$ & $\mathbf{59.8}$ & $57.0$ & $\underline{68.5}$ & $\underline{64.9}$ & $\mathbf{60.4}$ & $\underline{72.0}$ & $73.1$ & $\underline{72.6}$ & $\underline{67.8}$ & $95.7$ & $63.0$ & $\underline{72.7}$ \\
        & DisCreT \shortcite{pujol2025discut} & $48.0$ & $57.3$ & $70.6$ & $\mathbf{54.4}$ & $57.6$ & $\underline{58.0}$ & $66.7$ & $59.7$ & $\underline{57.7}$ & $60.0$ & $\underline{74.8}$ & $66.7$ & $63.2$ & $\underline{97.0}$ & $\underline{63.5}$ & $67.8$ \\
        & CLAC \shortcite{turk2025clac} & $48.0$ & $53.7$ & $70.8$ & $50.9$ & $55.4$ & $53.8$ & $60.3$ & $56.3$ & $56.3$ & $54.8$ & $72.6$ & $66.4$ & $60.3$ & $97.0$ & $60.9$ & $68.1$ \\
        & SeCoRel \shortcite{lalitha2025secorel} & $43.9$ & $49.8$ & $67.5$ & $52.6$ & $52.2$ & $55.2$ & $60.0$ & $54.2$ & $56.1$ & $52.8$ & $70.2$ & $62.5$ & $58.8$ & $96.3$ & $56.6$ & $62.2$ \\
        \bottomrule
    \end{tabular}}
    \caption{\textbf{Task (3) Discourse Relation Recognition [language-split]} scores across 16 languages: BeDiscovER (top) vs. supervised systems (bottom). Language abbreviations (left to right): Czech, German, English, Basque, Persian, French, Italian, Dutch, Nigerian Pidgin, Polish, Portuguese, Russian, Spanish, Thai, Turkish, and Chinese.
    The numbers of datasets and training examples in each language are given in supervised settings.
    }
    \label{tab:app-3-0}
\end{table*}

\begin{table*}[htbp]
    \centering
    \resizebox{0.95\textwidth}{!}{
\begin{tabular}{llccccccccccccc}
        \toprule
        & & \multicolumn{4}{c}{ces.rst.crdt} & \multicolumn{4}{c}{deu.pdtb.pcc} & \multicolumn{4}{c}{deu.rst.pcc}\\
        \cmidrule(lr){3-6} \cmidrule(lr){7-10} \cmidrule(lr){11-14}
        Model & Size & w/o r & tok. & w/ r & tok. & w/o r & tok. & w/ r & tok. & w/o r & tok. & w/ r & tok. \\
        \midrule
        Qwen3-0.6B & 0.6B & $5.1_{2.3}$ & 3 & $3.2_{1.1}$ & 2505 & $8.4_{5.1}$ & 2 & $9.1_{0.8}$ & 1979 & $5.5_{2.2}$ & 2 & $6.7_{1.2}$ & 1515 \\
        Qwen3-1.7B & 1.7B & $13.1_{1.8}$ & 1107 & $9.3_{1.8}$ & 2706 & $13.4_{0.5}$ & 1504 & $10.3_{2.1}$ & 2280 & $13.9_{1.1}$ & 904 & $11.1_{0.3}$ & 1833 \\
        Qwen3-4B & 4B & $16.9_{0.7}$ & 3 & $11.5_{2.0}$ & 2543 & $21.5_{0.7}$ & 2 & $12.5_{3.0}$ & 2120 & $14.3_{0.7}$ & 3 & $15.8_{0.7}$ & 1653 \\
        Qwen3-8B & 8B & $22.7_{0.3}$ & 3 & $11.3_{0.9}$ & 1544 & $20.6_{0.5}$ & 5 & $16.7_{2.4}$ & 1547 & $12.9_{0.7}$ & 3 & $17.7_{4.6}$ & 1245 \\
        Qwen3-14B & 14B & $28.4_{0.7}$ & 3 & $25.5_{1.5}$ & 1199 & $28.0_{0.4}$ & 3 & $23.0_{1.7}$ & 1121 & $23.3_{0.5}$ & 3 & $26.4_{1.4}$ & 950 \\
        Qwen3-32B & 32B & $34.5_{0.6}$ & 2 & $28.4_{1.3}$ & 689 & $25.6_{1.2}$ & 2 & $24.2_{3.6}$ & 660 & $23.2_{1.7}$ & 2 & $28.3_{1.0}$ & 602 \\
\bottomrule
    \end{tabular}}
    \caption{Details on \textbf{Task (3) Discourse Relation Recognition [language.framework.dataset split]} scores on 38 separate datasets in DISRPT shared task (1/4).
    }
    \label{tab:app-3-1-1}
\end{table*}
        
\begin{table*}[htbp]
    \centering
    \resizebox{0.95\textwidth}{!}{
    \begin{tabular}{llccccccccccccc}
        \toprule
        & & \multicolumn{4}{c}{eng.rst.rstdt} & \multicolumn{4}{c}{eng.rst.sts} & \multicolumn{4}{c}{eng.rst.umuc}\\
        \cmidrule(lr){3-6} \cmidrule(lr){7-10} \cmidrule(lr){11-14}
        Model & Size & w/o r & tok. & w/ r & tok. & w/o r & tok. & w/ r & tok. & w/o r & tok. & w/ r & tok. \\
        \midrule
        Qwen3-0.6B & 0.6B &$3.9_{0.9}$ & 2 & $15.8_{0.5}$ & 563 & $7.9_{0.9}$ & 2 & $13.2_{0.8}$ & 1290 & $3.7_{1.6}$ & 2 & $16.5_{1.1}$ & 988 \\
        Qwen3-1.7B & 1.7B &$9.9_{1.4}$ & 186 & $26.3_{0.7}$ & 1190 & $13.7_{0.9}$ & 875 & $16.6_{2.3}$ & 1692 & $12.2_{2.1}$ & 450 & $22.8_{1.4}$ & 1472 \\
        Qwen3-4B & 4B &$31.0_{0.6}$ & 2 & $33.7_{0.3}$ & 921 & $26.3_{1.1}$ & 11 & $21.7_{1.5}$ & 1475 & $18.6_{0.2}$ & 5 & $27.1_{0.8}$ & 1281 \\
        Qwen3-8B & 8B &$25.6_{0.6}$ & 2 & $36.1_{2.0}$ & 1312 & $28.3_{1.0}$ & 3 & $21.7_{1.2}$ & 1306 & $18.5_{0.5}$ & 3 & $34.0_{3.8}$ & 1291 \\
        Qwen3-14B & 14B &$38.8_{0.6}$ & 3 & $43.1_{0.6}$ & 814 & $36.0_{0.9}$ & 3 & $30.4_{1.6}$ & 924 & $30.0_{0.8}$ & 3 & $41.9_{1.5}$ & 878 \\
        Qwen3-32B & 32B &$39.9_{1.1}$ & 2 & $44.5_{1.1}$ & 556 & $33.3_{1.2}$ & 2 & $29.2_{1.0}$ & 614 & $36.0_{0.4}$ & 2 & $42.9_{0.3}$ & 593 \\
\bottomrule
    \end{tabular}}
    \caption{Details on \textbf{Task (3) Discourse Relation Recognition [language.framework.dataset split]} scores on 38 separate datasets in DISRPT shared task (2/4).
    }
    \label{tab:app-3-1-2}
\end{table*}

\begin{table*}[htbp]
    \centering
    \resizebox{0.95\textwidth}{!}{
    \begin{tabular}{llccccccccccccc}
        \toprule
        & & \multicolumn{4}{c}{por.pdtb.crpc} & \multicolumn{4}{c}{por.pdtb.tedm} & \multicolumn{4}{c}{por.rst.cstn}\\
        \cmidrule(lr){3-6} \cmidrule(lr){7-10} \cmidrule(lr){11-14}
        Model & Size & w/o r & tok. & w/ r & tok. & w/o r & tok. & w/ r & tok. & w/o r & tok. & w/ r & tok. \\
        \midrule
        Qwen3-0.6B & 0.6B &$12.2_{5.3}$ & 3 & $17.5_{0.6}$ & 652 & $13.6_{4.2}$ & 2 & $14.3_{1.9}$ & 1207 & $6.0_{3.3}$ & 3 & $10.3_{1.2}$ & 1476 \\
        Qwen3-1.7B & 1.7B &$14.7_{2.9}$ & 247 & $25.6_{0.6}$ & 1098 & $17.3_{1.1}$ & 873 & $23.5_{1.2}$ & 1530 & $10.3_{0.4}$ & 618 & $21.7_{1.8}$ & 1828 \\
        Qwen3-4B & 4B &$23.7_{0.5}$ & 3 & $26.4_{1.0}$ & 963 & $25.7_{0.4}$ & 3 & $24.5_{0.5}$ & 1377 & $31.5_{0.9}$ & 18 & $27.6_{2.2}$ & 1652 \\
        Qwen3-8B & 8B &$28.5_{1.1}$ & 3 & $31.6_{0.9}$ & 1353 & $27.0_{2.4}$ & 3 & $29.5_{1.3}$ & 1388 & $30.4_{2.0}$ & 3 & $34.9_{1.1}$ & 1322 \\
        Qwen3-14B & 14B &$39.5_{0.9}$ & 3 & $39.7_{0.3}$ & 794 & $32.5_{0.2}$ & 3 & $35.6_{2.0}$ & 873 & $46.0_{0.3}$ & 3 & $39.6_{3.0}$ & 971 \\
        Qwen3-32B & 32B &$39.9_{1.6}$ & 2 & $41.1_{0.6}$ & 545 & $33.7_{1.5}$ & 2 & $38.8_{1.9}$ & 569 & $47.1_{0.3}$ & 2 & $45.0_{1.0}$ & 624 \\
\bottomrule
    \end{tabular}}
    \caption{Details on \textbf{Task (3) Discourse Relation Recognition [language.framework.dataset split]} scores on 38 separate datasets in DISRPT shared task (3/4).
    }
    \label{tab:app-3-1-3}
\end{table*}

\begin{table*}[htbp]
    \centering
    \resizebox{0.72\textwidth}{!}{
    \begin{tabular}{llccccccccc}
        \toprule
        & & \multicolumn{4}{c}{zho.rst.gcdt} & \multicolumn{4}{c}{zho.rst.sctb} \\
        \cmidrule(lr){3-6} \cmidrule(lr){7-10}
        Model & Size & w/o r & tok. & w/ r & tok. & w/o r & tok. & w/ r & tok. \\
        \midrule
        Qwen3-0.6B & 0.6B &$6.3_{2.2}$ & 4 & $14.9_{0.8}$ & 703 & $4.3_{3.0}$ & 3 & $6.2_{1.1}$ & 2321 \\
        Qwen3-1.7B & 1.7B &$11.8_{0.9}$ & 188 & $18.9_{0.2}$ & 1113 & $4.6_{3.6}$ & 1310 & $13.6_{1.5}$ & 2498 \\
        Qwen3-4B & 4B &$19.3_{0.2}$ & 4 & $24.2_{1.7}$ & 946 & $33.5_{0.5}$ & 3 & $15.7_{1.9}$ & 2348 \\
        Qwen3-8B & 8B &$20.7_{0.3}$ & 3 & $26.4_{1.9}$ & 1204 & $32.7_{1.9}$ & 4 & $17.4_{0.8}$ & 1521 \\
        Qwen3-14B & 14B &$27.2_{0.4}$ & 3 & $31.1_{0.3}$ & 785 & $43.0_{0.4}$ & 4 & $33.3_{2.5}$ & 1039 \\
        Qwen3-32B & 32B &$31.4_{0.4}$ & 2 & $33.2_{0.6}$ & 499 & $43.4_{1.3}$ & 2 & $37.1_{1.3}$ & 556 \\
        \midrule
        DS-r1-distill-Qwen & 32B & - & - &$25.0_{0.8}$ & 278 & - & - & $35.2_{0.6}$ & 293 \\
        DeepSeek-r1-0528 & 37/671B & - & - &$36.8_{0.0}$ & 1120 & - & - & $54.1_{0.0}$ & 1036 \\
        GPT-5-mini (low) & Unk. & - & - &$31.8_{0.3}$ & 124 & - & - & $44.7_{0.6}$ & 115 \\
        GPT-5-mini (high) & Unk. & - & - &$38.4_{0.0}$ & 944 & - & - & $50.3_{0.0}$ & 882 \\
        \midrule
        DeDisCO & - & \multicolumn{4}{c}{$75.35$} & \multicolumn{4}{c}{$75.47$} \\
        HITS & - & \multicolumn{4}{c}{$71.46$} & \multicolumn{4}{c}{$57.86$} \\
        DiscReT & - & \multicolumn{4}{c}{$61.91$} & \multicolumn{4}{c}{$55.35$} \\
        \bottomrule
    \end{tabular}}
    \caption{Details on \textbf{Task (3) Discourse Relation Recognition [language.framework.dataset split]} scores on 38 separate datasets in DISRPT shared task (4/4).
    }
    \label{tab:app-3-1-4}
\end{table*}

\FloatBarrier

\subsection{Task (4) Sentence Ordering}
\label{append-so}

\Cref{tab:app-4-1-1,tab:app-4-1-2,tab:app-4-1-3,tab:app-4-1-4,tab:app-4-1-5,tab:app-4-1-6,tab:app-4-1-7}

\begin{table*}[htbp]
    \centering
    \resizebox{\textwidth}{!}{
    \begin{tabular}{llcccccccccccccc}
        \toprule
        & & \multicolumn{6}{c}{{w/o reasoning}} & \multicolumn{6}{c}{w/ reasoning} \\
        \cmidrule(lr){3-8} \cmidrule(lr){9-14}
        Model & Size & Acc & Lumr & Pmr & Skip & Lcs & tok. & Acc & Lumr & Pmr & Skip & Lcs & tok. \\
        \midrule
        Qwen3-0.6B & 0.6B & $21.7_{0.3}$ & $32.4_{4.7}$ & $5.4_{0.9}$ & $50.0_{1.9}$ & $55.3_{0.5}$ & 24 & $22.3_{0.3}$ & $8.4_{0.0}$ & $6.3_{0.8}$ & $53.0_{0.4}$ & $56.3_{0.1}$ & 559 \\
        Qwen3-1.7B & 1.7B & $31.2_{0.2}$ & $13.4_{0.6}$ & $13.6_{0.2}$ & $65.2_{0.3}$ & $64.3_{0.1}$ & 25 & $38.0_{0.2}$ & $8.9_{0.7}$ & $21.1_{0.4}$ & $71.5_{0.6}$ & $68.6_{0.5}$ & 720 \\
        Qwen3-4B & 4B & $54.3_{0.1}$ & $1.1_{0.0}$ & $35.7_{0.2}$ & $84.8_{0.0}$ & $78.3_{0.1}$ & 25 & $58.7_{0.3}$ & $2.1_{0.5}$ & $41.0_{0.4}$ & $87.0_{0.0}$ & $81.0_{0.1}$ & 449 \\
        Qwen3-8B & 8B & $60.6_{0.1}$ & $0.3_{0.1}$ & $42.1_{0.1}$ & $88.1_{0.0}$ & $82.0_{0.1}$ & 25 & $64.5_{0.2}$ & $1.0_{0.1}$ & $48.7_{0.4}$ & $89.5_{0.1}$ & $84.0_{0.2}$ & 413 \\
        Qwen3-14B & 14B & $69.8_{0.0}$ & $0.9_{0.1}$ & $54.6_{0.1}$ & $91.5_{0.0}$ & $87.1_{0.0}$ & 25 & $70.6_{0.4}$ & $0.7_{0.0}$ & $55.5_{0.7}$ & $91.7_{0.1}$ & $87.2_{0.2}$ & 451 \\
        Qwen3-32B & 32B & $74.6_{0.1}$ & $1.4_{0.1}$ & $60.4_{0.1}$ & $92.9_{0.2}$ & $89.3_{0.1}$ & 25 & $74.0_{0.3}$ & $1.5_{0.1}$ & $59.6_{0.4}$ & $92.8_{0.1}$ & $88.8_{0.2}$ & 445 \\
\bottomrule
    \end{tabular}}
    \caption{Details on \textbf{Task (4) Sentence Ordering} on \textbf{AAN abstract} dataset.
    Metric abbreviations: Acc - accuracy; Lumr – length unmatch rate (proportion of outputs whose generated length differs from the original sequence); Pmr – perfect match rate; Skip – skip bigram match rate; Lcs – longest common subsequence.
    }
    \label{tab:app-4-1-1}
\end{table*}

\begin{table*}[htbp]
    \centering
    \resizebox{\textwidth}{!}{
    \begin{tabular}{llcccccccccccc}
        \toprule
        & & \multicolumn{6}{c}{{w/o reasoning}} & \multicolumn{6}{c}{w/ reasoning} \\
        \cmidrule(lr){3-8} \cmidrule(lr){9-14}
        Model & Size & Acc & Lumr & Pmr & Skip & Lcs & tok. & Acc & Lumr & Pmr & Skip & Lcs & tok. \\
        \midrule
        Qwen3-0.6B & 0.6B & $17.6_{0.6}$ & $29.5_{4.8}$ & $5.3_{0.8}$ & $46.4_{1.4}$ & $51.4_{0.6}$ & 27 & $19.2_{0.3}$ & $9.5_{1.5}$ & $8.3_{0.7}$ & $51.7_{0.4}$ & $52.9_{0.0}$ & 629 \\
        Qwen3-1.7B & 1.7B & $26.4_{0.2}$ & $17.8_{0.1}$ & $12.0_{0.2}$ & $58.4_{0.2}$ & $59.5_{0.2}$ & 30 & $31.4_{0.6}$ & $10.4_{0.5}$ & $19.1_{0.5}$ & $67.6_{0.1}$ & $62.5_{0.5}$ & 829 \\
        Qwen3-4B & 4B & $47.2_{0.1}$ & $0.4_{0.2}$ & $31.6_{0.1}$ & $82.2_{0.0}$ & $73.5_{0.1}$ & 27 & $47.7_{0.1}$ & $2.3_{0.3}$ & $32.4_{1.5}$ & $82.6_{0.2}$ & $74.5_{0.3}$ & 477 \\
        Qwen3-8B & 8B & $52.3_{0.2}$ & $0.6_{0.0}$ & $35.9_{0.3}$ & $85.3_{0.0}$ & $77.2_{0.1}$ & 27 & $53.5_{0.3}$ & $1.2_{0.1}$ & $37.6_{0.5}$ & $85.5_{0.3}$ & $78.2_{0.1}$ & 458 \\
        Qwen3-14B & 14B & $58.6_{0.2}$ & $0.8_{0.1}$ & $42.9_{0.2}$ & $88.2_{0.0}$ & $81.4_{0.1}$ & 27 & $59.1_{0.4}$ & $0.4_{0.4}$ & $43.4_{0.4}$ & $88.2_{0.1}$ & $81.5_{0.1}$ & 475 \\
        Qwen3-32B & 32B & $63.8_{0.2}$ & $0.3_{0.0}$ & $48.2_{0.2}$ & $89.5_{0.0}$ & $84.1_{0.1}$ & 27 & $61.9_{0.2}$ & $0.2_{0.0}$ & $46.1_{0.3}$ & $89.1_{0.3}$ & $83.1_{0.2}$ & 450 \\
\bottomrule
    \end{tabular}}
    \caption{Details on \textbf{Task (4) Sentence Ordering} on \textbf{ArXiv abstract} dataset.
    Metric abbreviations: Acc - accuracy; Lumr – length unmatch rate (proportion of outputs whose generated length differs from the original sequence); Pmr – perfect match rate; Skip – skip bigram match rate; Lcs – longest common subsequence.
    }
    \label{tab:app-4-1-2}
\end{table*}

\begin{table*}[htbp]
    \centering
    \resizebox{\textwidth}{!}{
    \begin{tabular}{llcccccccccccccc}
        \toprule
        & & \multicolumn{6}{c}{{w/o reasoning}} & \multicolumn{6}{c}{w/ reasoning} \\
        \cmidrule(lr){3-8} \cmidrule(lr){9-14}
        Model & Size & Acc & Lumr & Pmr & Skip & Lcs & tok. & Acc & Lumr & Pmr & Skip & Lcs & tok. \\
        \midrule
        Qwen3-0.6B & 0.6B & $15.5_{1.7}$ & $32.7_{4.7}$ & $1.7_{0.4}$ & $48.1_{2.2}$ & $51.0_{1.3}$ & 30 & $17.1_{0.4}$ & $10.4_{0.7}$ & $1.7_{0.4}$ & $52.5_{0.7}$ & $52.2_{0.2}$ & 677 \\
        Qwen3-1.7B & 1.7B & $23.8_{0.2}$ & $21.0_{0.8}$ & $3.7_{0.0}$ & $61.6_{0.2}$ & $58.5_{0.3}$ & 32 & $30.8_{1.1}$ & $14.2_{0.7}$ & $11.0_{1.5}$ & $69.2_{1.3}$ & $64.6_{1.3}$ & 915 \\
        Qwen3-4B & 4B & $48.5_{0.1}$ & $1.5_{0.1}$ & $21.4_{0.4}$ & $84.6_{0.0}$ & $76.2_{0.0}$ & 31 & $54.1_{0.9}$ & $2.8_{0.4}$ & $26.2_{0.9}$ & $86.5_{0.1}$ & $78.1_{0.4}$ & 508 \\
        Qwen3-8B & 8B & $54.3_{0.1}$ & $0.8_{0.0}$ & $25.6_{0.1}$ & $87.1_{0.1}$ & $78.9_{0.0}$ & 31 & $60.0_{1.6}$ & $1.7_{1.2}$ & $33.5_{1.8}$ & $89.4_{0.4}$ & $81.8_{0.3}$ & 475 \\
        Qwen3-14B & 14B & $66.3_{0.1}$ & $0.8_{0.0}$ & $42.2_{0.2}$ & $91.5_{0.1}$ & $85.6_{0.1}$ & 31 & $66.1_{0.1}$ & $0.3_{0.2}$ & $43.0_{1.3}$ & $91.8_{0.0}$ & $85.7_{0.4}$ & 548 \\
        Qwen3-32B & 32B & $71.2_{0.0}$ & $0.7_{0.1}$ & $47.7_{0.3}$ & $92.5_{0.2}$ & $87.8_{0.1}$ & 31 & $68.6_{0.4}$ & $0.6_{0.2}$ & $44.2_{0.9}$ & $92.3_{0.1}$ & $86.9_{0.3}$ & 499 \\
\bottomrule
    \end{tabular}}
    \caption{Details on \textbf{Task (4) Sentence Ordering} on \textbf{Neurips abstract} dataset.
    Metric abbreviations: Acc - accuracy; Lumr – length unmatch rate (proportion of outputs whose generated length differs from the original sequence); Pmr – perfect match rate; Skip – skip bigram match rate; Lcs – longest common subsequence.
    }
    \label{tab:app-4-1-3}
\end{table*}

\begin{table*}[htbp]
    \centering
    \resizebox{\textwidth}{!}{
    \begin{tabular}{llcccccccccccccc}
        \toprule
        & & \multicolumn{6}{c}{{w/o reasoning}} & \multicolumn{6}{c}{w/ reasoning} \\
        \cmidrule(lr){3-8} \cmidrule(lr){9-14}
        Model & Size & Acc & Lumr & Pmr & Skip & Lcs & tok. & Acc & Lumr & Pmr & Skip & Lcs & tok. \\
        \midrule
        Qwen3-0.6B & 0.6B & $9.6_{0.6}$ & $41.6_{8.6}$ & $1.1_{0.1}$ & $43.5_{2.2}$ & $38.8_{0.9}$ & 62 & $10.7_{0.4}$ & $28.2_{0.2}$ & $4.3_{0.4}$ & $49.7_{0.1}$ & $41.0_{0.1}$ & 770 \\
        Qwen3-1.7B & 1.7B & $12.7_{0.1}$ & $42.3_{0.9}$ & $1.9_{0.1}$ & $52.1_{0.0}$ & $44.7_{0.1}$ & 59 & $14.7_{0.2}$ & $33.2_{0.3}$ & $6.6_{0.5}$ & $57.3_{0.3}$ & $46.2_{0.7}$ & 1242 \\
        Qwen3-4B & 4B & $23.3_{0.1}$ & $18.7_{0.1}$ & $11.1_{0.0}$ & $71.5_{0.1}$ & $57.5_{0.1}$ & 54 & $23.6_{0.0}$ & $19.7_{2.0}$ & $12.0_{0.4}$ & $72.0_{0.2}$ & $58.4_{0.3}$ & 610 \\
        Qwen3-8B & 8B & $26.0_{0.1}$ & $11.7_{0.3}$ & $11.2_{0.1}$ & $74.8_{0.1}$ & $60.2_{0.2}$ & 65 & $26.4_{0.3}$ & $16.5_{2.4}$ & $12.0_{0.3}$ & $74.6_{0.2}$ & $61.1_{0.3}$ & 591 \\
        Qwen3-14B & 14B & $30.9_{0.0}$ & $10.1_{0.6}$ & $15.6_{0.0}$ & $77.8_{0.1}$ & $65.7_{0.2}$ & 52 & $29.2_{0.5}$ & $12.1_{1.7}$ & $14.3_{0.1}$ & $76.7_{0.3}$ & $64.3_{0.1}$ & 616 \\
        Qwen3-32B & 32B & $32.8_{0.3}$ & $9.3_{0.5}$ & $15.5_{0.1}$ & $78.5_{0.1}$ & $67.7_{0.1}$ & 52 & $30.2_{0.3}$ & $11.4_{0.1}$ & $14.3_{0.2}$ & $77.5_{0.2}$ & $65.6_{0.2}$ & 546 \\
\bottomrule
    \end{tabular}}
    \caption{Details on \textbf{Task (4) Sentence Ordering} on \textbf{NSF abstract} dataset.
    Metric abbreviations: Acc - accuracy; Lumr – length unmatch rate (proportion of outputs whose generated length differs from the original sequence); Pmr – perfect match rate; Skip – skip bigram match rate; Lcs – longest common subsequence.
    }
    \label{tab:app-4-1-4}
\end{table*}

\begin{table*}[htbp]
    \centering
    \resizebox{\textwidth}{!}{
    \begin{tabular}{llcccccccccccccc}
        \toprule
        & & \multicolumn{6}{c}{{w/o reasoning}} & \multicolumn{6}{c}{w/ reasoning} \\
        \cmidrule(lr){3-8} \cmidrule(lr){9-14}
        Model & Size & Acc & Lumr & Pmr & Skip & Lcs & tok. & Acc & Lumr & Pmr & Skip & Lcs & tok. \\
        \midrule
        Qwen3-0.6B & 0.6B & $22.8_{0.5}$ & $21.4_{8.3}$ & $1.2_{0.2}$ & $52.5_{3.4}$ & $57.9_{1.4}$ & 23 & $25.6_{0.1}$ & $7.7_{3.4}$ & $2.3_{0.4}$ & $56.6_{0.3}$ & $60.5_{0.2}$ & 487 \\
        Qwen3-1.7B & 1.7B & $37.5_{0.2}$ & $0.0_{0.0}$ & $8.0_{0.3}$ & $68.8_{0.2}$ & $68.1_{0.1}$ & 24 & $44.7_{0.6}$ & $1.9_{0.1}$ & $17.1_{0.5}$ & $73.8_{0.3}$ & $74.0_{0.4}$ & 647 \\
        Qwen3-4B & 4B & $63.6_{0.1}$ & $0.0_{0.0}$ & $35.1_{0.2}$ & $86.5_{0.1}$ & $83.5_{0.1}$ & 24 & $72.3_{0.6}$ & $0.1_{0.0}$ & $50.2_{1.0}$ & $90.0_{0.2}$ & $88.0_{0.4}$ & 481 \\
        Qwen3-8B & 8B & $69.1_{0.0}$ & $0.0_{0.0}$ & $43.6_{0.1}$ & $89.3_{0.0}$ & $86.3_{0.0}$ & 24 & $78.4_{0.2}$ & $0.0_{0.1}$ & $59.9_{1.3}$ & $92.5_{0.1}$ & $90.6_{0.2}$ & 525 \\
        Qwen3-14B & 14B & $79.5_{0.1}$ & $0.0_{0.1}$ & $61.5_{0.1}$ & $92.8_{0.1}$ & $91.1_{0.0}$ & 24 & $82.5_{0.5}$ & $0.0_{0.1}$ & $67.3_{0.7}$ & $94.0_{0.2}$ & $92.5_{0.1}$ & 367 \\
        Qwen3-32B & 32B & $82.7_{0.1}$ & $0.1_{0.0}$ & $67.0_{0.3}$ & $94.2_{0.1}$ & $92.6_{0.2}$ & 24 & $84.4_{0.7}$ & $0.0_{0.1}$ & $70.6_{0.9}$ & $94.7_{0.3}$ & $93.3_{0.2}$ & 452 \\
\bottomrule
    \end{tabular}}
    \caption{Details on \textbf{Task (4) Sentence Ordering} on \textbf{ROC stories} dataset.
    Metric abbreviations: Acc - accuracy; Lumr – length unmatch rate (proportion of outputs whose generated length differs from the original sequence); Pmr – perfect match rate; Skip – skip bigram match rate; Lcs – longest common subsequence.
    }
    \label{tab:app-4-1-5}
\end{table*}

\begin{table*}[htbp]
    \centering
    \resizebox{\textwidth}{!}{
    \begin{tabular}{llcccccccccccccc}
        \toprule
        & & \multicolumn{6}{c}{{w/o reasoning}} & \multicolumn{6}{c}{w/ reasoning} \\
        \cmidrule(lr){3-8} \cmidrule(lr){9-14}
        Model & Size & Acc & Lumr & Pmr & Skip & Lcs & tok. & Acc & Lumr & Pmr & Skip & Lcs & tok. \\
        \midrule
        Qwen3-0.6B & 0.6B & $21.9_{0.8}$ & $24.6_{6.8}$ & $2.0_{0.5}$ & $47.9_{2.2}$ & $55.8_{1.1}$ & 24 & $24.6_{0.9}$ & $10.9_{1.6}$ & $5.1_{0.6}$ & $54.3_{1.4}$ & $57.8_{0.7}$ & 515 \\
        Qwen3-1.7B & 1.7B & $28.3_{0.4}$ & $6.0_{0.1}$ & $3.9_{0.2}$ & $57.3_{0.2}$ & $61.5_{0.1}$ & 24 & $35.6_{1.0}$ & $2.8_{0.6}$ & $14.6_{0.7}$ & $66.4_{0.4}$ & $65.7_{0.7}$ & 699 \\
        Qwen3-4B & 4B & $46.9_{0.2}$ & $0.8_{0.0}$ & $22.2_{0.2}$ & $76.4_{0.0}$ & $73.3_{0.1}$ & 24 & $49.6_{0.3}$ & $0.9_{0.2}$ & $24.8_{0.7}$ & $78.5_{0.2}$ & $75.2_{0.3}$ & 561 \\
        Qwen3-8B & 8B & $50.0_{0.1}$ & $0.2_{0.1}$ & $25.9_{0.2}$ & $78.9_{0.1}$ & $75.4_{0.1}$ & 23 & $52.0_{0.7}$ & $0.3_{0.1}$ & $27.4_{1.1}$ & $80.2_{0.3}$ & $76.7_{0.3}$ & 630 \\
        Qwen3-14B & 14B & $53.9_{0.1}$ & $0.0_{0.0}$ & $29.3_{0.3}$ & $81.2_{0.1}$ & $77.5_{0.1}$ & 23 & $55.1_{0.4}$ & $0.1_{0.2}$ & $31.2_{0.9}$ & $81.4_{0.6}$ & $78.1_{0.8}$ & 449 \\
        Qwen3-32B & 32B & $54.0_{0.2}$ & $0.2_{0.2}$ & $29.7_{0.2}$ & $80.8_{0.0}$ & $77.6_{0.2}$ & 23 & $55.3_{0.6}$ & $0.5_{0.3}$ & $31.2_{0.9}$ & $81.8_{0.4}$ & $78.2_{0.2}$ & 491 \\
\bottomrule
    \end{tabular}}
    \caption{Details on \textbf{Task (4) Sentence Ordering} on \textbf{SIND} dataset.
    Metric abbreviations: Acc - accuracy; Lumr – length unmatch rate (proportion of outputs whose generated length differs from the original sequence); Pmr – perfect match rate; Skip – skip bigram match rate; Lcs – longest common subsequence.
    }
    \label{tab:app-4-1-6}
\end{table*}

\begin{table*}[htbp]
    \centering
    \resizebox{\textwidth}{!}{
    \begin{tabular}{llcccccccccccccc}
        \toprule
        & & \multicolumn{6}{c}{{w/o reasoning}} & \multicolumn{6}{c}{w/ reasoning} \\
        \cmidrule(lr){3-8} \cmidrule(lr){9-14}
        Model & Size & Acc & Lumr & Pmr & Skip & Lcs & tok. & Acc & Lumr & Pmr & Skip & Lcs & tok. \\
        \midrule
        Qwen3-0.6B & 0.6B & $6.8_{0.3}$ & $53.7_{10.2}$ & $1.8_{0.6}$ & $38.2_{1.7}$ & $31.9_{1.5}$ & 114 & $8.5_{0.1}$ & $45.2_{1.2}$ & $4.2_{0.8}$ & $43.9_{0.8}$ & $34.4_{0.5}$ & 941 \\
        Qwen3-1.7B & 1.7B & $12.1_{0.1}$ & $36.6_{1.0}$ & $5.2_{0.1}$ & $55.4_{0.1}$ & $41.8_{0.1}$ & 73 & $12.7_{0.2}$ & $47.7_{1.8}$ & $8.9_{0.3}$ & $52.9_{0.3}$ & $40.4_{0.3}$ & 1313 \\
        Qwen3-4B & 4B & $20.2_{0.1}$ & $26.3_{0.7}$ & $12.8_{0.1}$ & $72.7_{0.1}$ & $53.3_{0.1}$ & 74 & $21.2_{0.3}$ & $33.7_{1.7}$ & $14.1_{0.9}$ & $71.9_{0.1}$ & $54.5_{0.4}$ & 976 \\
        Qwen3-8B & 8B & $20.8_{0.1}$ & $21.1_{0.9}$ & $14.8_{0.2}$ & $72.8_{0.3}$ & $53.7_{0.3}$ & 139 & $23.2_{0.4}$ & $31.5_{0.7}$ & $15.9_{0.4}$ & $75.1_{0.1}$ & $58.2_{0.2}$ & 902 \\
        Qwen3-14B & 14B & $27.2_{0.1}$ & $23.5_{1.0}$ & $17.4_{0.1}$ & $80.5_{0.1}$ & $63.1_{0.1}$ & 72 & $27.1_{0.2}$ & $29.5_{0.6}$ & $18.2_{0.2}$ & $79.1_{0.2}$ & $63.7_{0.1}$ & 608 \\
        Qwen3-32B & 32B & $30.4_{0.3}$ & $15.2_{0.6}$ & $19.3_{0.3}$ & $82.7_{0.0}$ & $65.9_{0.1}$ & 73 & $29.1_{0.6}$ & $21.5_{2.4}$ & $19.1_{0.5}$ & $80.9_{0.4}$ & $65.2_{0.3}$ & 545 \\
\bottomrule
    \end{tabular}}
    \caption{Details on \textbf{Task (4) Sentence Ordering} on \textbf{Wikipedia movie plots} dataset.
    Metric abbreviations: Acc - accuracy; Lumr – length unmatch rate (proportion of outputs whose generated length differs from the original sequence); Pmr – perfect match rate; Skip – skip bigram match rate; Lcs – longest common subsequence.
    }
    \label{tab:app-4-1-7}
\end{table*}

\FloatBarrier

\subsection{Task (5) Dialogue Discourse Parsing}
\label{append-ddp}

\Cref{tab:app-5-1-1,tab:app-5-1-2,tab:app-5-1-3}.

\begin{table*}[htbp]
    \centering
    \resizebox{\textwidth}{!}{
    \begin{tabular}{llcccccccccccc}
        \toprule
        & & \multicolumn{6}{c}{single turn} & \multicolumn{6}{c}{auto-regressive} \\
        \cmidrule(lr){3-8} \cmidrule(lr){9-14}
        & & \multicolumn{3}{c}{w/o reasoning} & \multicolumn{3}{c}{w/ reasoning} & \multicolumn{3}{c}{w/o reasoning} & \multicolumn{3}{c}{w/ reasoning} \\
        \cmidrule(lr){3-5} \cmidrule(lr){6-8} \cmidrule(lr){9-11} \cmidrule(lr){12-14}
        Model & Size & Link & Full & tok. & Link & Full & tok. & Link & Full & tok. & Link & Full & tok.\\
        \midrule
        Qwen3-1.7B & 1.7B & $17.5_{0.5}$ & $2.5_{0.1}$ & 12 & $39.7_{0.6}$ & $8.4_{0.9}$ & 944 & $16.0_{1.0}$ & $0.4_{0.0}$ & 9 & $35.9_{2.5}$ & $5.2_{0.2}$ & 762 \\
        Qwen3-4B & 4B & $42.8_{0.2}$ & $8.0_{0.2}$ & 9 & $56.9_{1.0}$ & $22.0_{0.4}$ & 1020 & $34.8_{0.7}$ & $4.6_{0.3}$ & 9 & $58.8_{0.7}$ & $19.6_{0.8}$ & 849 \\
        Qwen3-8B & 8B & $52.4_{0.2}$ & $12.4_{0.3}$ & 9 & $59.2_{0.3}$ & $21.9_{0.5}$ & 942 & $51.6_{0.9}$ & $8.9_{1.1}$ & 9 & $58.8_{0.1}$ & $17.2_{0.9}$ & 754 \\
        Qwen3-14B & 14B & $55.7_{0.2}$ & $16.2_{0.2}$ & 9 & $62.5_{1.3}$ & $33.1_{0.3}$ & 750 & $55.8_{0.3}$ & $13.2_{0.9}$ & 9 & $61.3_{0.5}$ & $29.8_{1.3}$ & 703 \\
        Qwen3-32B & 32B & $58.1_{0.4}$ & $21.5_{1.4}$ & 9 & $62.1_{0.2}$ & $33.3_{0.4}$ & 611 & $57.8_{0.6}$ & $16.5_{1.3}$ & 9 & $62.4_{0.5}$ & $31.3_{0.3}$ & 583 \\
        \midrule
        DS-r1-distill-Qwen & 32B & - & - & - & $62.9_{0.4}$ & $25.5_{0.6}$ & 430 & - & - & - & $61.5_{0.7}$ & $22.3_{1.5}$ & 429 \\
        DeepSeek-r1-0528 & 37/671B & - & - & - & $68.0_{0.0}$ & $42.3_{0.0}$ & 1494 & - & - & - & $66.3_{0.0}$ & $38.7_{0.0}$ & 1526 \\
        GPT-5-mini (low) & Unk. & - & - & - & $62.1_{0.4}$ & $33.6_{0.8}$ & 146 & - & - & - & $61.1_{0.3}$ & $32.1_{0.8}$ & 142 \\
        GPT-5-mini (high) & Unk. & - & - & - &$67.0_{0.5}$ & $39.9_{0.0}$ & 1235 & - & - & - & $66.0_{0.0}$ & $38.8_{0.0}$ & 1254 \\        
        \bottomrule
    \end{tabular}}
    \caption{Details on \textbf{Task (5) Dialogue Discourse Parsing} on \textbf{STAC} dataset.
    }
    \label{tab:app-5-1-1}
\end{table*}

\begin{table*}[htbp]
    \centering
    \resizebox{\textwidth}{!}{
    \begin{tabular}{llcccccccccccc}
        \toprule
        & & \multicolumn{6}{c}{single turn} & \multicolumn{6}{c}{auto-regressive} \\
        \cmidrule(lr){3-8} \cmidrule(lr){9-14}
        & & \multicolumn{3}{c}{w/o reasoning} & \multicolumn{3}{c}{w/ reasoning} & \multicolumn{3}{c}{w/o reasoning} & \multicolumn{3}{c}{w/ reasoning} \\
        \cmidrule(lr){3-5} \cmidrule(lr){6-8} \cmidrule(lr){9-11} \cmidrule(lr){12-14}
        Model & Size & Link & Full & tok. & Link & Full & tok. & Link & Full & tok. & Link & Full & tok.\\
        \midrule
        Qwen3-1.7B & 1.7B & $14.1_{0.2}$ & $3.0_{0.1}$ & 17 & $39.5_{0.3}$ & $10.6_{0.5}$ & 907 & $20.7_{0.5}$ & $0.2_{0.0}$ & 8 & $43.1_{0.5}$ & $5.3_{0.4}$ & 728 \\
        Qwen3-4B & 4B & $41.1_{0.5}$ & $7.6_{0.2}$ & 9 & $55.0_{0.3}$ & $23.3_{0.7}$ & 994 & $33.7_{0.7}$ & $4.9_{0.4}$ & 9 & $57.9_{0.3}$ & $22.6_{0.2}$ & 943 \\
        Qwen3-8B & 8B & $48.5_{0.2}$ & $10.4_{0.4}$ & 8 & $58.8_{0.6}$ & $20.0_{0.2}$ & 1006 & $46.4_{0.8}$ & $7.4_{0.3}$ & 8 & $60.6_{0.1}$ & $16.8_{0.1}$ & 963 \\
        Qwen3-14B & 14B & $58.0_{0.3}$ & $19.9_{0.4}$ & 9 & $59.6_{0.3}$ & $27.2_{0.4}$ & 743 & $61.4_{0.1}$ & $18.0_{0.3}$ & 9 & $60.1_{0.6}$ & $25.7_{0.8}$ & 743 \\
        Qwen3-32B & 32B & $59.0_{0.3}$ & $18.4_{0.9}$ & 9 & $59.0_{0.1}$ & $24.7_{0.4}$ & 611 & $58.1_{0.6}$ & $13.7_{1.7}$ & 9 & $58.3_{0.5}$ & $22.2_{0.8}$ & 629 \\
        \midrule
        DS-r1-distill-Qwen & 32B & - & - & - & $59.3_{0.4}$ & $17.5_{0.4}$ & 445 & - & - & - & $58.6_{0.6}$ & $16.1_{0.3}$ & 446 \\
        DeepSeek-r1-0528 & 37/671B & - & - & - & $57.7_{0.0}$ & $23.8_{0.0}$ & 1499 & - & - & - & $56.7_{0.0}$ & $22.3_{0.0}$ & 1511 \\
        GPT-5-mini (low) & Unk. & - & - & - & $57.3_{0.6}$ & $29.0_{0.1}$ & 157 & - & - & - & $55.6_{0.1}$ & $27.7_{0.3}$ & 160 \\
        GPT-5-mini (high) & Unk. & - & - & - & $59.2_{0.0}$ & $31.5_{0.0}$ & 1272 & - & - & - & $58.4_{0.0}$ & $30.2_{0.0}$ & 1344 \\     
        \bottomrule
    \end{tabular}}
    \caption{Details on \textbf{Task (5) Dialogue Discourse Parsing} on \textbf{Molweni} dataset.
    }
    \label{tab:app-5-1-2}
\end{table*}

\begin{table*}[htbp]
    \centering
    \resizebox{\textwidth}{!}{
    \begin{tabular}{llcccccccccccc}
        \toprule
        & & \multicolumn{6}{c}{single turn} & \multicolumn{6}{c}{auto-regressive} \\
        \cmidrule(lr){3-8} \cmidrule(lr){9-14}
        & & \multicolumn{3}{c}{w/o reasoning} & \multicolumn{3}{c}{w/ reasoning} & \multicolumn{3}{c}{w/o reasoning} & \multicolumn{3}{c}{w/ reasoning} \\
        \cmidrule(lr){3-5} \cmidrule(lr){6-8} \cmidrule(lr){9-11} \cmidrule(lr){12-14}
        Model & Size & Link & Full & tok. & Link & Full & tok. & Link & Full & tok. & Link & Full & tok.\\
        \midrule
        Qwen3-1.7B & 1.7B & $4.4_{0.0}$ & $1.4_{0.0}$ & 16 & $29.4_{0.1}$ & $6.0_{0.3}$ & 926 & $2.5_{0.2}$ & $0.3_{0.0}$ & 149 & $36.7_{2.2}$ & $3.6_{0.1}$ & 747 \\
        Qwen3-4B & 4B & $35.8_{0.2}$ & $11.7_{0.2}$ & 10 & $67.6_{0.5}$ & $24.0_{0.4}$ & 893 & $19.9_{0.9}$ & $5.7_{0.1}$ & 10 & $72.2_{0.1}$ & $17.2_{0.5}$ & 764 \\
        Qwen3-8B & 8B & $57.9_{0.4}$ & $20.0_{0.2}$ & 11 & $72.1_{0.1}$ & $29.4_{0.5}$ & 832 & $56.1_{0.8}$ & $8.8_{0.2}$ & 11 & $73.7_{0.2}$ & $17.9_{0.3}$ & 657 \\
        Qwen3-14B & 14B & $58.5_{0.2}$ & $24.4_{0.2}$ & 12 & $67.9_{0.2}$ & $34.5_{0.2}$ & 658 & $70.4_{0.2}$ & $15.1_{0.4}$ & 10 & $72.7_{0.2}$ & $22.6_{0.1}$ & 536 \\
        Qwen3-32B & 32B & $65.0_{0.3}$ & $28.1_{0.3}$ & 12 & $67.9_{0.2}$ & $36.2_{0.4}$ & 551 & $68.3_{0.8}$ & $10.4_{0.7}$ & 10 & $70.5_{0.1}$ & $23.0_{0.3}$ & 491 \\
        \midrule
        DS-r1-distill-Qwen & 32B & - & - & - & $68.9_{0.4}$ & $36.1_{0.4}$ & 415 & - & - & - & $69.6_{0.4}$ & $25.7_{0.7}$ & 412 \\
        DeepSeek-r1-0528 & 37/671B & - & - & - &$69.4_{0.0}$ & $48.6_{0.0}$ & 1702 & - & - & - & $69.6_{0.0}$ & $34.2_{0.0}$ & 1893 \\
        GPT-5-mini (low) & Unk. & - & - & - & $65.8_{0.0}$ & $37.8_{0.0}$ & 141 & - & - & - & $66.1_{0.1}$ & $26.8_{0.4}$ & 146 \\
        GPT-5-mini (high) & Unk. & - & - & - &$69.1_{0.0}$ & $46.3_{0.0}$ & 1321 & - & - & - & $66.4_{0.0}$ & $31.9_{0.0}$ & 1472 \\
        \bottomrule
    \end{tabular}}
    \caption{Details on \textbf{Task (5) Dialogue Discourse Parsing} on \textbf{MSDC} dataset.
    }
    \label{tab:app-5-1-3}
\end{table*}

\FloatBarrier

\section{Appendix F. Visualizations}
\label{append-plot}

\subsection{Task (1) Discourse Marker Understanding}
\label{append-dm-v}

\begin{figure*}[htbp]
    \centering
    \begin{subfigure}[b]{0.49\textwidth}
        \centering
        \includegraphics[width=\textwidth]{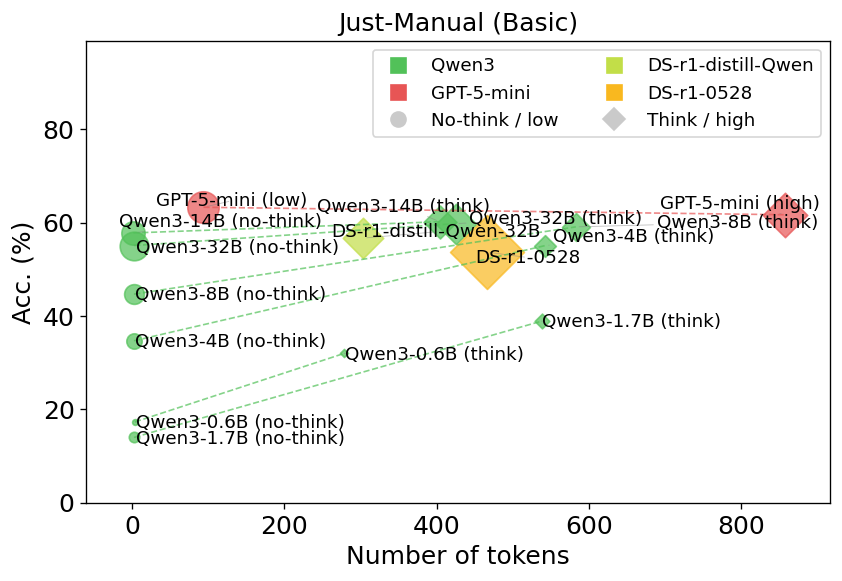}
    \end{subfigure}
    \begin{subfigure}[b]{0.49\textwidth}
        \centering
        \includegraphics[width=\textwidth]{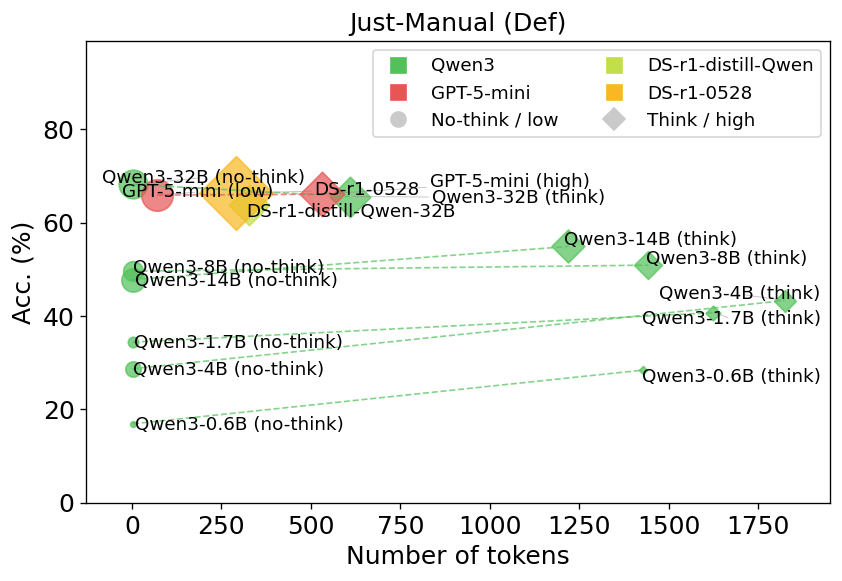}
    \end{subfigure}
    \begin{subfigure}[b]{0.49\textwidth}
        \centering
        \includegraphics[width=\textwidth]{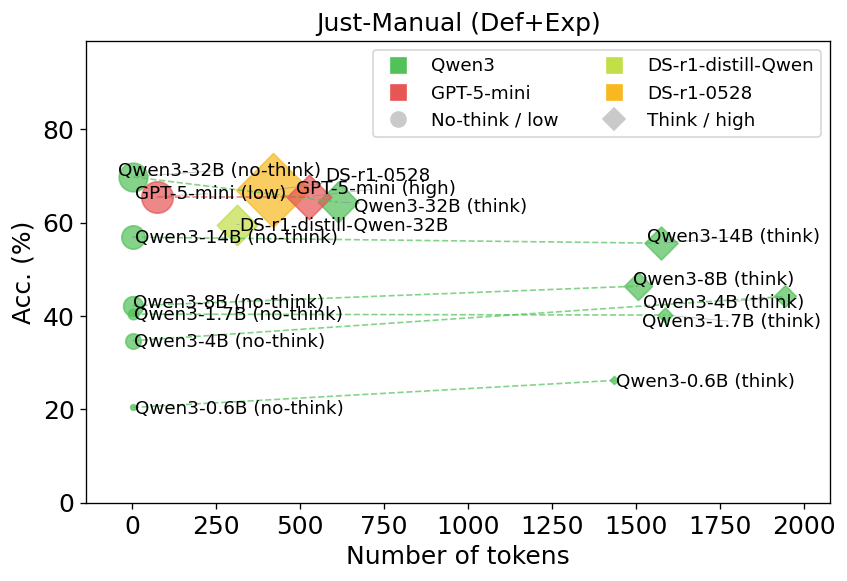}
    \end{subfigure}
    \caption{Performance comparison on \textbf{Task (1) Discourse Marker Understanding} on \textbf{\textit{Just-Manual}} dataset with different \textbf{reasoning} mode or effort.
    }
    \label{fig:app-1-1}
\end{figure*}

\begin{figure*}[htbp]
    \centering
    \begin{subfigure}[b]{0.49\textwidth}
        \centering
        \includegraphics[width=\textwidth]{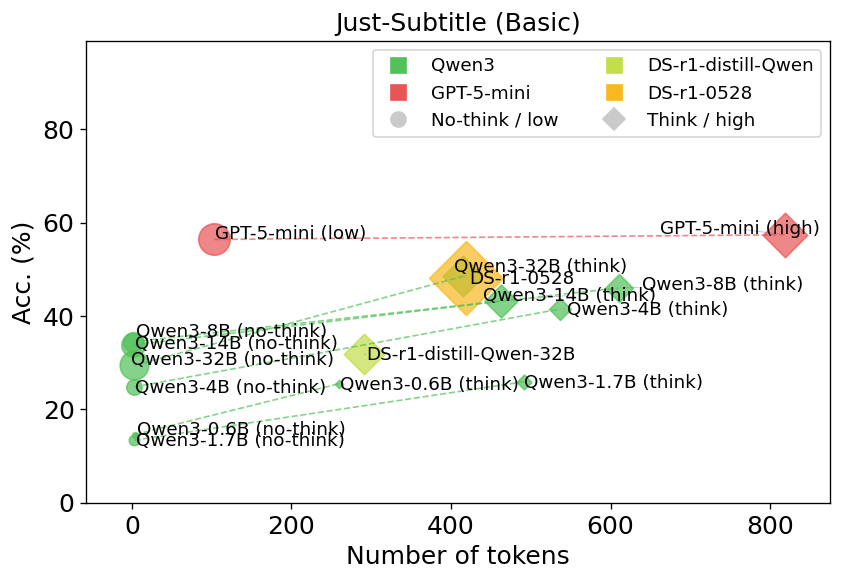}
    \end{subfigure}
    \begin{subfigure}[b]{0.49\textwidth}
        \centering
        \includegraphics[width=\textwidth]{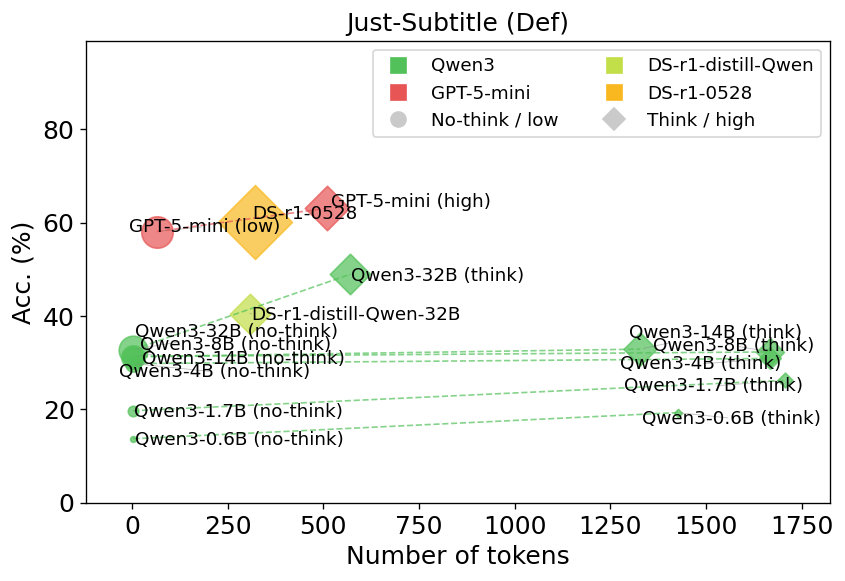}
    \end{subfigure}
    \begin{subfigure}[b]{0.49\textwidth}
        \centering
        \includegraphics[width=\textwidth]{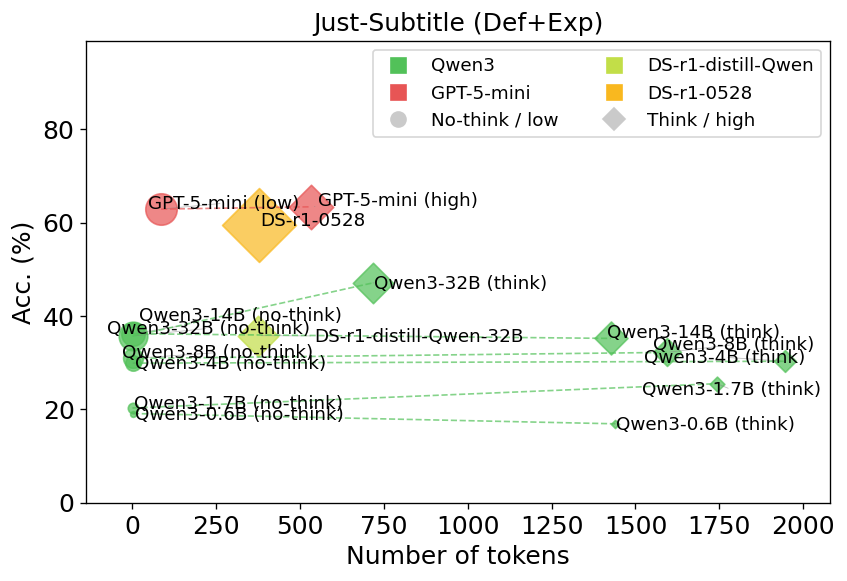}
    \end{subfigure}
    \caption{Performance comparison on \textbf{Task (1) Discourse Marker Understanding} on \textbf{\textit{Just-Subtitle}} dataset with different \textbf{reasoning} mode or effort.
    }
    \label{fig:app-1-2}
\end{figure*}

\begin{figure*}[htbp]
    \centering
    \begin{subfigure}[b]{0.49\textwidth}
        \centering
        \includegraphics[width=\textwidth]{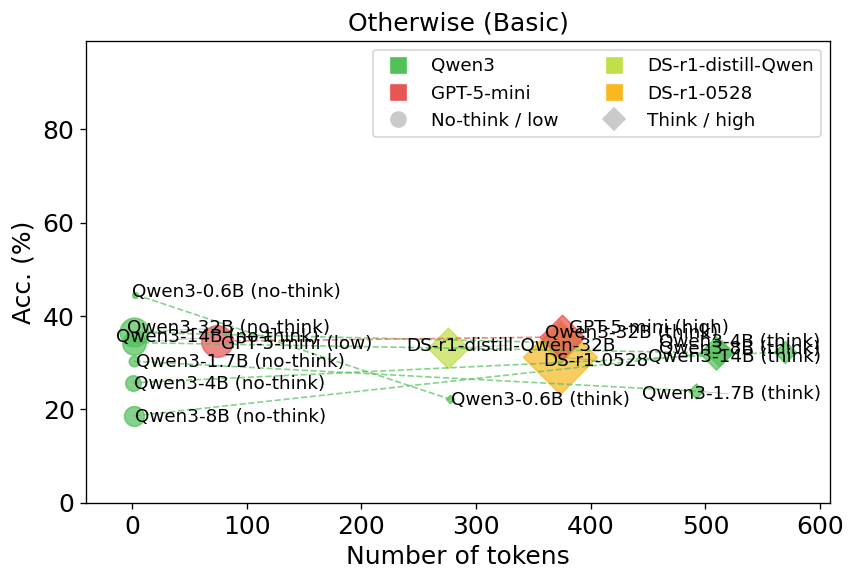}
    \end{subfigure}
    \begin{subfigure}[b]{0.49\textwidth}
        \centering
        \includegraphics[width=\textwidth]{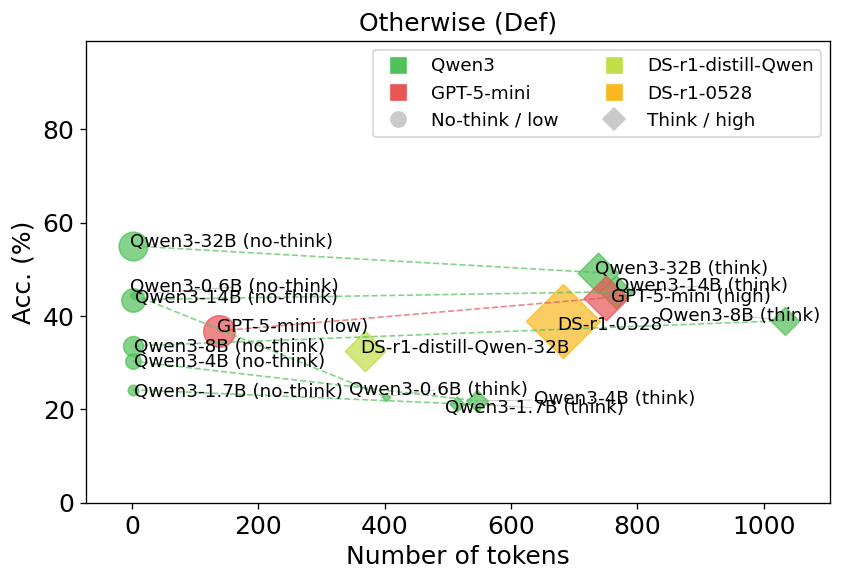}
    \end{subfigure}
    \begin{subfigure}[b]{0.49\textwidth}
        \centering
        \includegraphics[width=\textwidth]{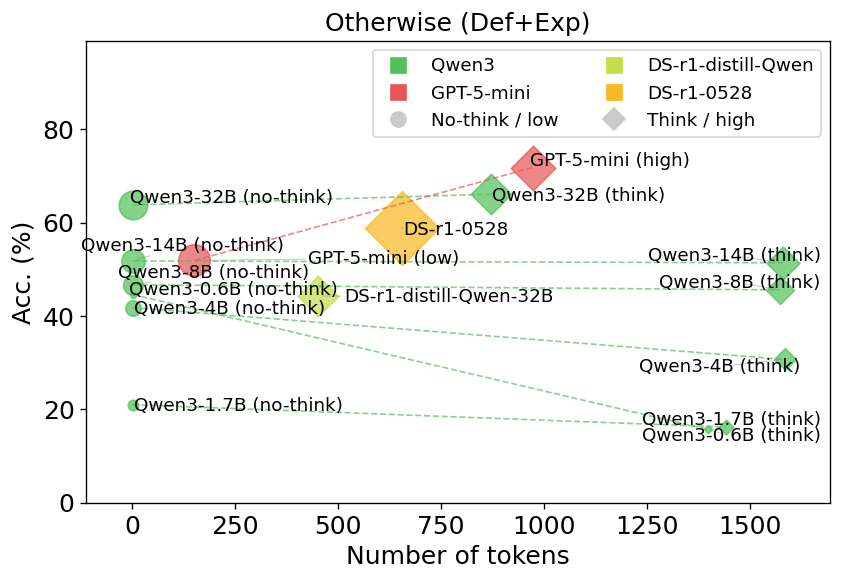}
    \end{subfigure}
    \caption{Performance comparison on \textbf{Task (1) Discourse Marker Understanding} on \textbf{\textit{Otherwise}} dataset with different \textbf{reasoning} mode or effort.
    }
    \label{fig:app-1-3}
\end{figure*}

\Cref{fig:app-1-1,fig:app-1-2,fig:app-1-3}.

\FloatBarrier

\subsection{Task (2) Temporal Reasoning}
\label{append-tr-v}

\begin{figure*}[htbp]
    \centering
    \begin{subfigure}[b]{0.49\textwidth}
        \centering
        \includegraphics[width=\textwidth]{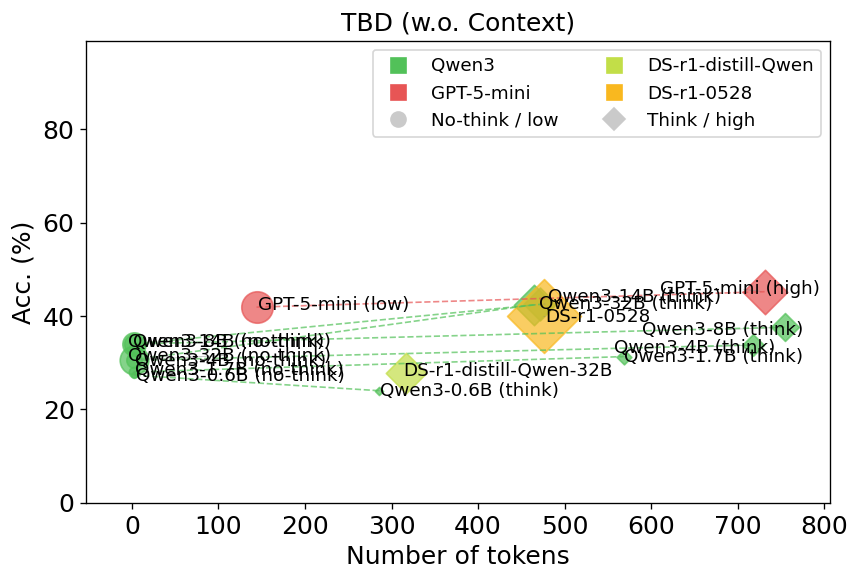}
    \end{subfigure}
    \begin{subfigure}[b]{0.49\textwidth}
        \centering
        \includegraphics[width=\textwidth]{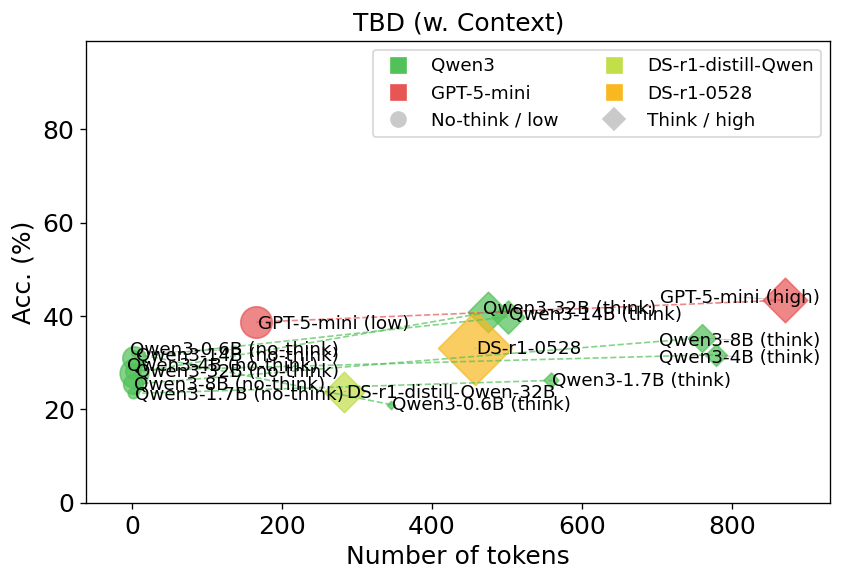}
    \end{subfigure}
    \caption{Performance comparison on \textbf{Task (2) Temporal Reasoning} on \textbf{TBD} dataset with different \textbf{reasoning} mode or effort.
    }
    \label{fig:app-2-1}
\end{figure*}

\begin{figure*}[htbp]
    \centering
    \begin{subfigure}[b]{0.49\textwidth}
        \centering
        \includegraphics[width=\textwidth]{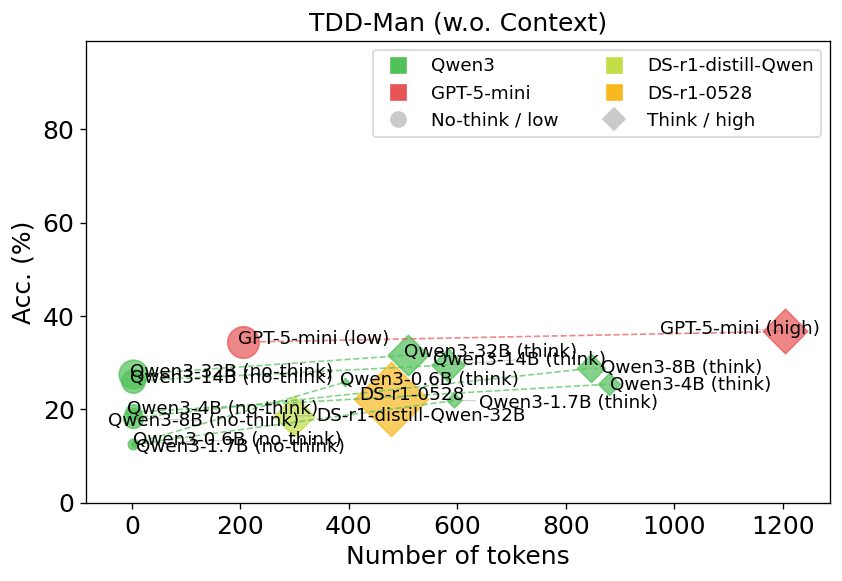}
    \end{subfigure}
    \begin{subfigure}[b]{0.49\textwidth}
        \centering
        \includegraphics[width=\textwidth]{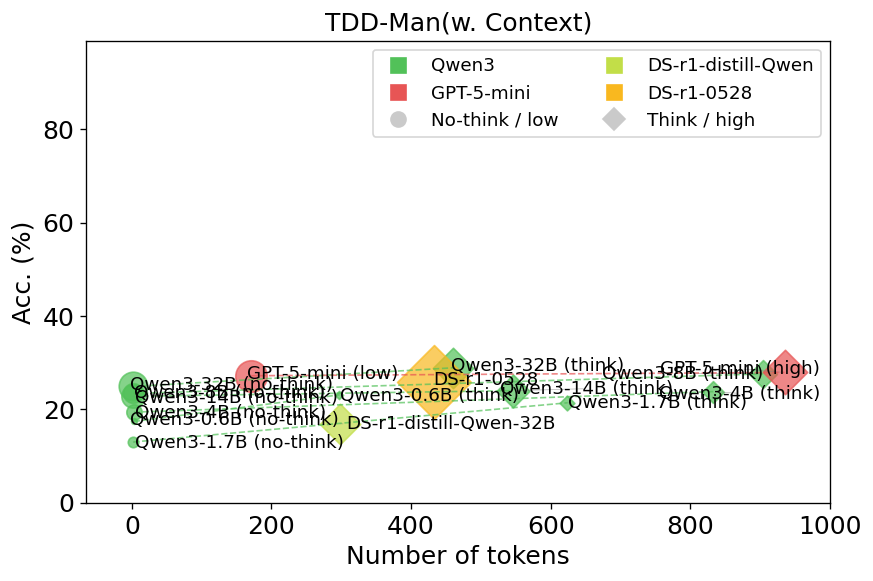}
    \end{subfigure}
    \caption{Performance comparison on \textbf{Task (2) Temporal Reasoning} on \textbf{TDD-Man} dataset with different \textbf{reasoning} mode or effort.
    }
    \label{fig:app-2-2}
\end{figure*}

\begin{figure*}[htbp]
    \centering
    \begin{subfigure}[b]{0.49\textwidth}
        \centering
        \includegraphics[width=\textwidth]{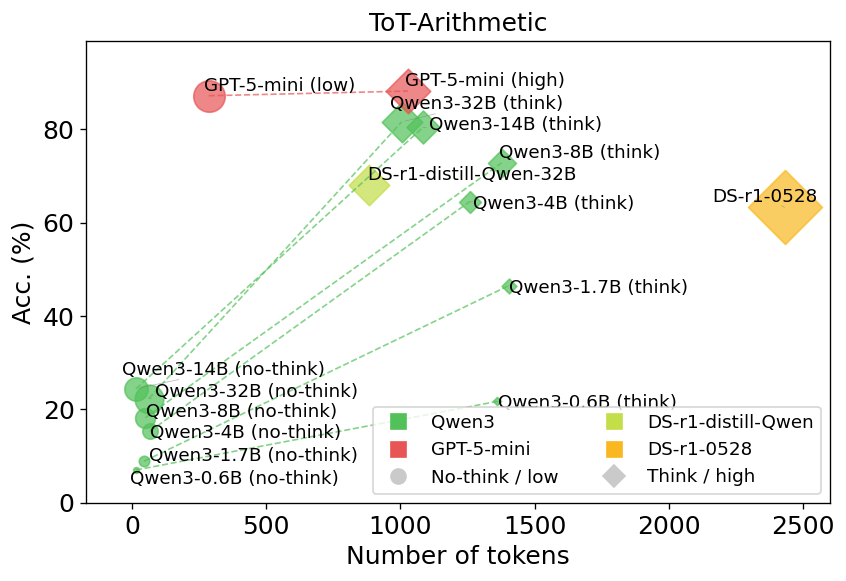}
    \end{subfigure}
    \caption{Performance comparison on \textbf{Task (2) Temporal Reasoning} on \textbf{ToT-Arithmetic } dataset with different \textbf{reasoning} mode or effort.
    }
    \label{fig:app-2-3}
\end{figure*}

\Cref{fig:app-2-1,fig:app-2-2,fig:app-2-3}.

\FloatBarrier



\FloatBarrier

\subsection{Task (4) Sentence Ordering}
\label{append-so-v}

\begin{figure*}[htbp]
    \centering
    \begin{subfigure}[b]{0.49\textwidth}
        \centering
        \includegraphics[width=\textwidth]{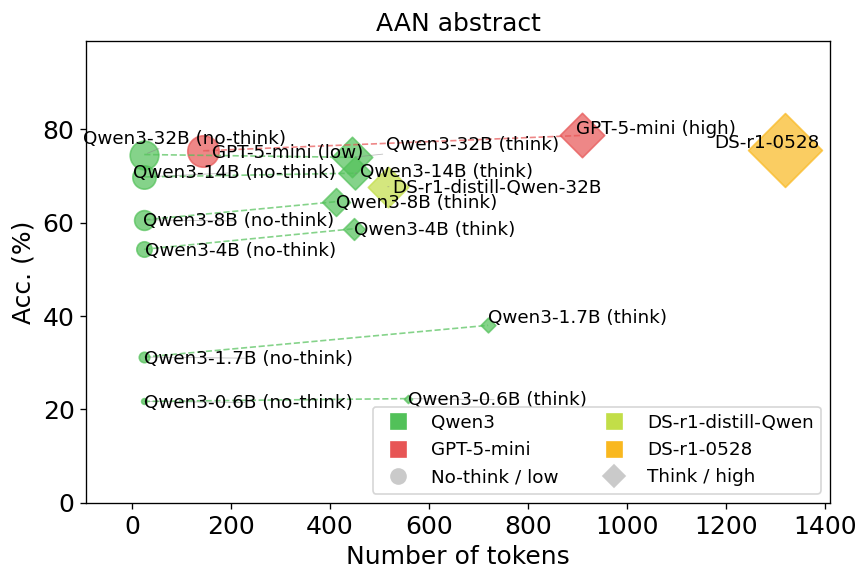}
    \end{subfigure}
    \begin{subfigure}[b]{0.49\textwidth}
        \centering
        \includegraphics[width=\textwidth]{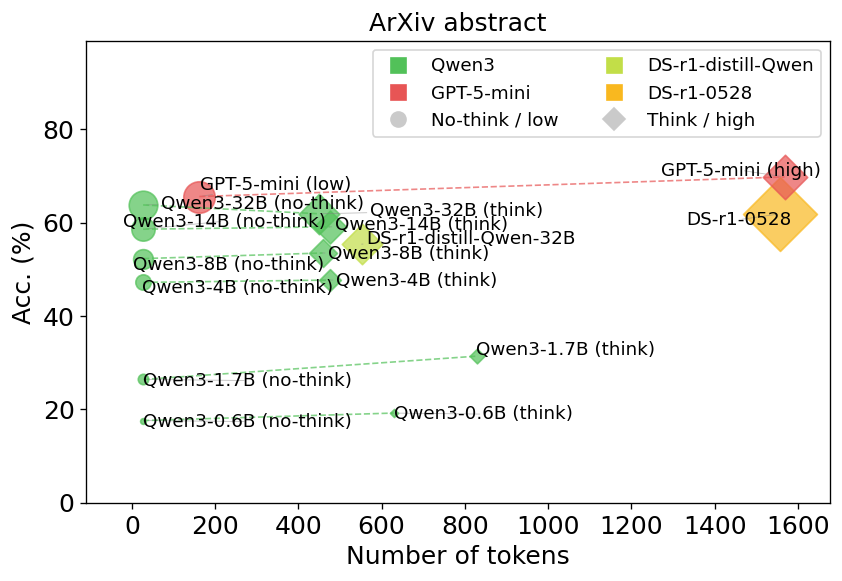}
    \end{subfigure}
    \caption{Performance comparison on \textbf{Task (4) Sentence Ordering} on \textbf{AAN abstract} and \textbf{ArXiv abstract} datasets with different \textbf{reasoning} mode or effort.
    }
    \label{fig:app-4-1}
\end{figure*}

\begin{figure*}[htbp]
    \centering
    \begin{subfigure}[b]{0.49\textwidth}
        \centering
        \includegraphics[width=\textwidth]{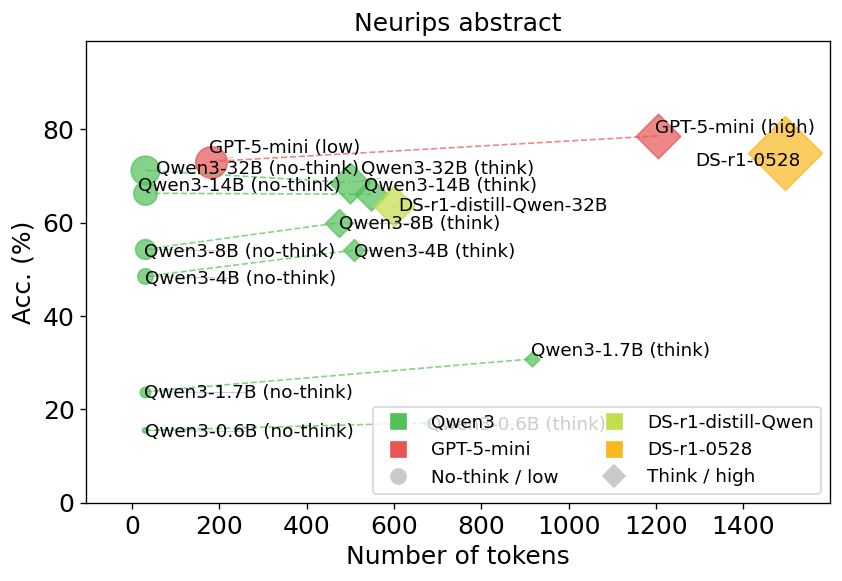}
    \end{subfigure}
    \begin{subfigure}[b]{0.49\textwidth}
        \centering
        \includegraphics[width=\textwidth]{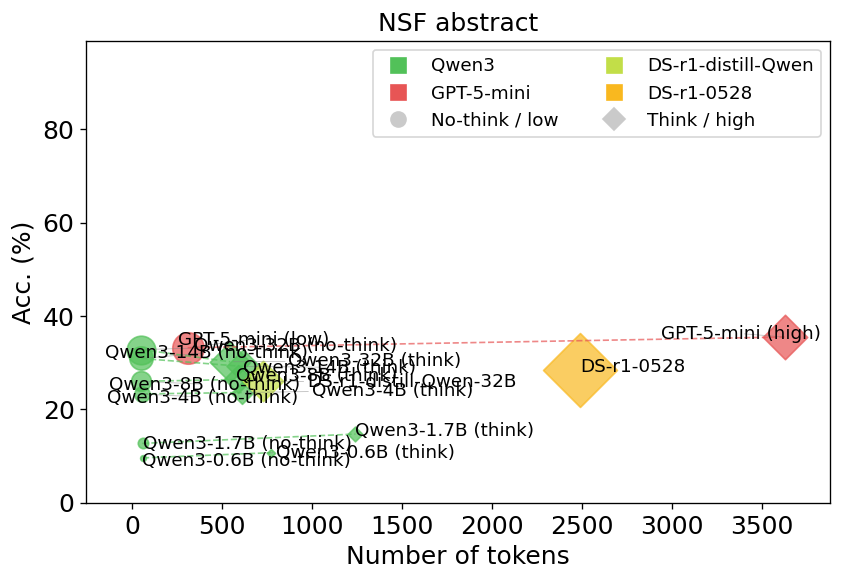}
    \end{subfigure}
    \caption{Performance comparison on \textbf{Task (4) Sentence Ordering} on \textbf{Neurips abstract} and \textbf{NSF abstract} datasets with different \textbf{reasoning} mode or effort.
    }
    \label{fig:app-4-2}
\end{figure*}

\begin{figure*}[htbp]
    \centering
    \begin{subfigure}[b]{0.49\textwidth}
        \centering
        \includegraphics[width=\textwidth]{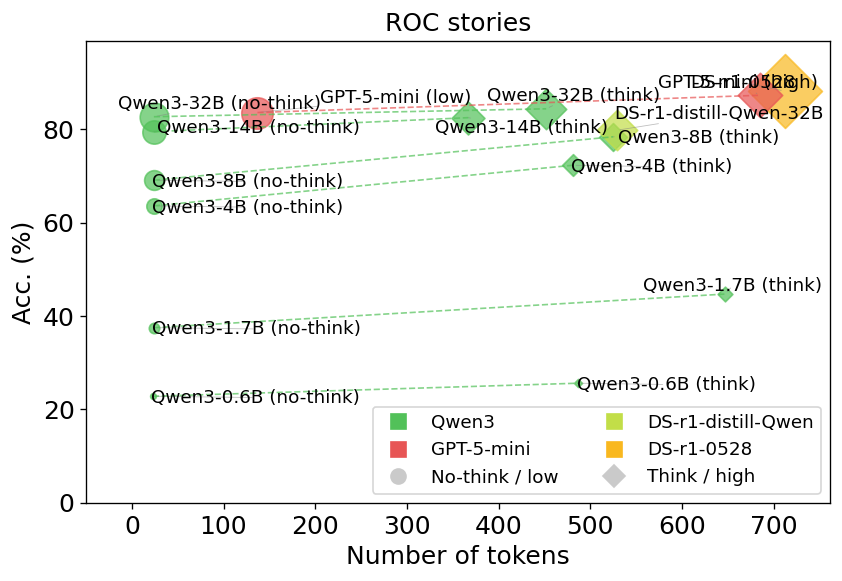}
    \end{subfigure}
    \begin{subfigure}[b]{0.49\textwidth}
        \centering
        \includegraphics[width=\textwidth]{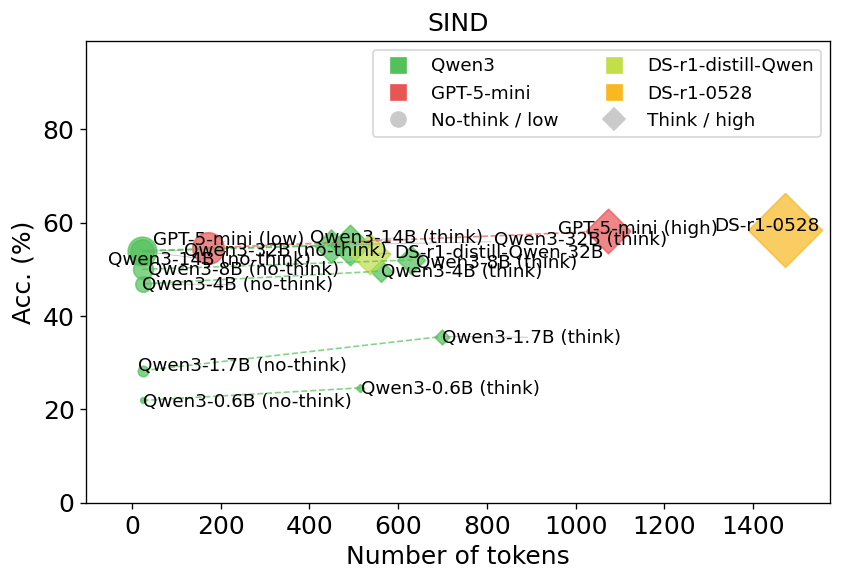}
    \end{subfigure}
    \caption{Performance comparison on \textbf{Task (4) Sentence Ordering} on \textbf{ROC stories} and \textbf{SIND} datasets with different \textbf{reasoning} mode or effort.
    }
    \label{fig:app-4-3}
\end{figure*}

\begin{figure*}[htbp]
    \centering
    \begin{subfigure}[b]{0.49\textwidth}
        \centering
        \includegraphics[width=\textwidth]{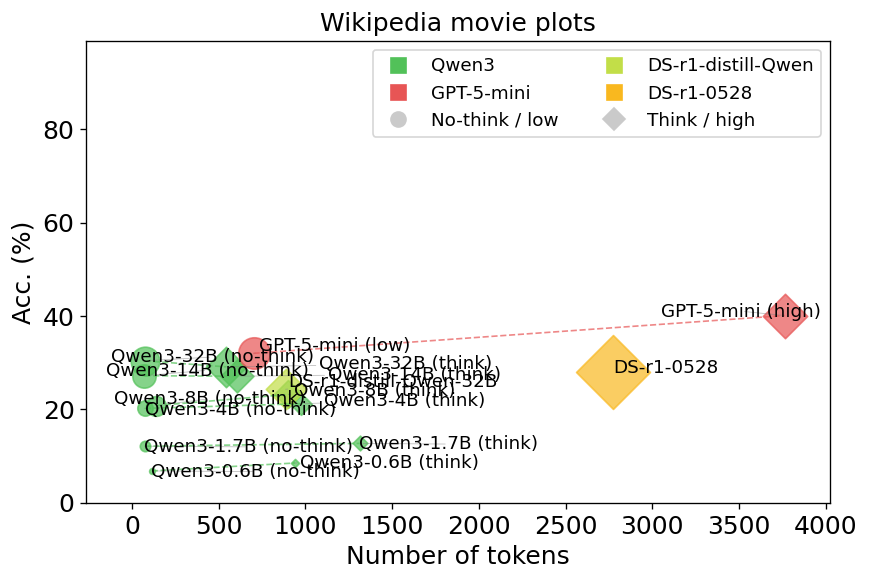}
    \end{subfigure}
    \caption{Performance comparison on \textbf{Task (4) Sentence Ordering} on \textbf{Wikipedia movie plots} dataset with different \textbf{reasoning} mode or effort.
    }
    \label{fig:app-4-4}
\end{figure*}

\Cref{fig:app-4-1,fig:app-4-2,fig:app-4-3,fig:app-4-4}.

\FloatBarrier

\subsection{Task (5) Dialogue Discourse Parsing}
\label{append-ddp-v}

\begin{figure*}[htbp]
    \centering
    \begin{subfigure}[b]{0.49\textwidth}
        \centering
        \includegraphics[width=\textwidth]{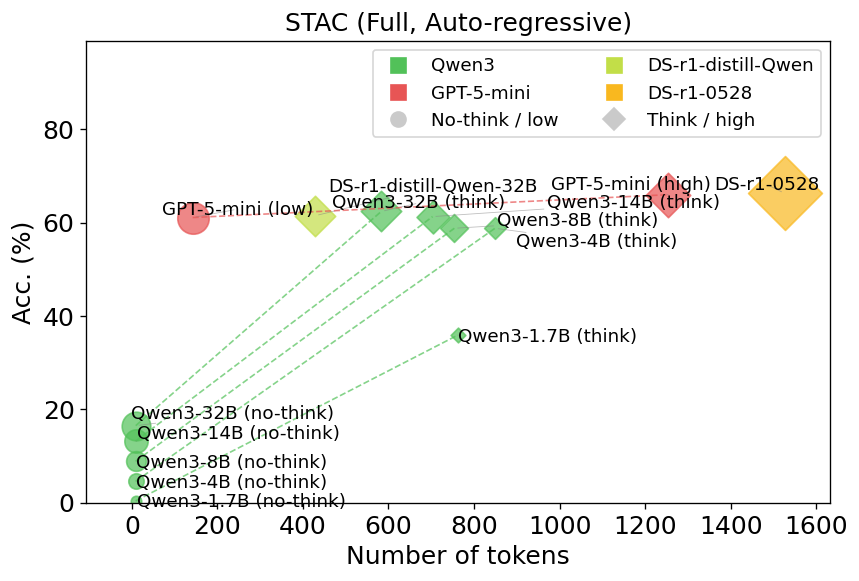}
    \end{subfigure}
    \begin{subfigure}[b]{0.49\textwidth}
        \centering
        \includegraphics[width=\textwidth]{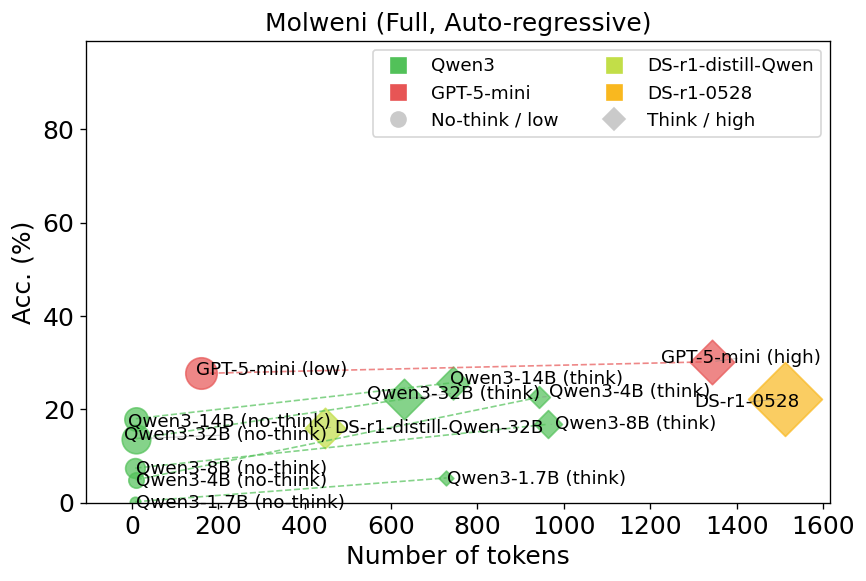}
    \end{subfigure}
    \begin{subfigure}[b]{0.49\textwidth}
        \centering
        \includegraphics[width=\textwidth]{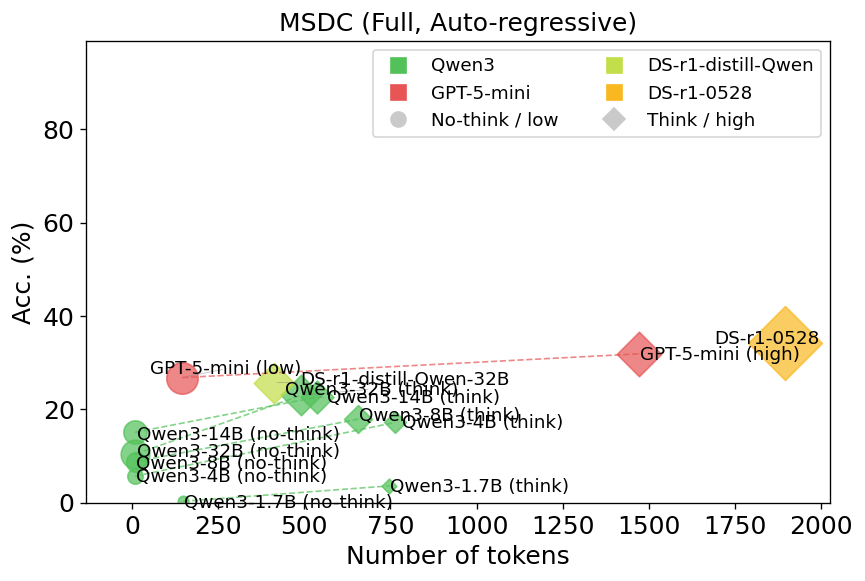}
    \end{subfigure}
    \caption{Performance comparison on \textbf{Task (4) Sentence Ordering} on \textbf{STAC}, \textbf{Molweni} and \textbf{MSDC} dataset with different \textbf{reasoning} mode or effort.
    }
    \label{fig:app-5-1}
\end{figure*}

\Cref{fig:app-5-1}.

\end{document}